\documentclass[journal,twocolumn]{IEEEtran}

\usepackage{dsfont}
\usepackage{amsmath,amsfonts,amssymb,amsthm}
\usepackage{graphicx,caption,wrapfig,subfig}
\usepackage{bm}
\newcommand*{\xhat}[1]{#1\kern-0.35em\hat{\phantom{#1}}}
\usepackage{mathrsfs,multicol}
\usepackage{amstext}
\usepackage{float}
\usepackage{nomencl}

\renewcommand{\vec}[1]{\mathbf{#1}}
\newcommand{\T}{\ensuremath{^\mathsf{T}}}
\DeclareMathOperator{\Tr}{Tr}

\newtheorem{lemma}{Lemma}[section]

\usepackage{scalerel}
\newcommand\reallywidehat[1]{\arraycolsep=0pt\relax%
\begin{array}{c}
\stretchto{
  \scaleto{
    \scalerel*[\widthof{\ensuremath{#1}}]{\kern-.5pt\bigwedge\kern-0.5pt}
    {\rule[-\textheight/2]{1ex}{\textheight}} 
  }{\textheight} %
}{1ex}\\           
#1\\                 
\rule{-1ex}{0ex}
\end{array}
}

\title{Robust Shape Control of Gyroscopic Tensegrity Robotic Arm}

\author{Raman~Goyal$^{1}$,
Manoranjan~Majji$^{2}$,
and~Robert~E.~Skelton$^{3}$

\thanks{$^{1}$Postdoctoral Researcher, Department of Aerospace Engineering, Texas A\&M University, College Station, TX 77843. Email: \textit{ramaniitrgoyal92@tamu.edu}}
\thanks{$^{2}$Assistant Professor, Department of Aerospace Engineering, Texas A\&M University, College Station, TX 77843. Email: \textit{mmajji@tamu.edu}}
\thanks{$^{3}$TEES Distinguished Research Professor, Department of Aerospace Engineering, Texas A\&M University, College Station, TX 77843. Email: \textit{bobskelton@tamu.edu}}}

\begin{document}

\maketitle


\begin{abstract}
This paper proposes a model-based approach to control the shape of a tensegrity system by driving its node position locations. The nonlinear dynamics of the tensegrity system is used to regulate position, velocity, and acceleration to the specified reference trajectory. State feedback control design is used to obtain the solution for the control variable as a \textit{linear programming} problem. Shape control for the gyroscopic tensegrity systems is discussed, and it is observed that these systems increase the reachable space for the structure by providing independent control over certain rotational degrees of freedom. Disturbance rejection of the tensegrity system is further studied in the paper. A methodology to calculate the control gains to bound the errors for five different types of problems is provided. The formulation uses a Linear Matrix Inequality (LMI) approach to stipulate the desired performance bounds on the error for $\mathcal{H}_\infty$, generalized $\mathcal{H}_2$, LQR, \textit{covariance control} and stabilizing control problem. A high degree of freedom tensegrity $T_2D_1$ robotic arm is used as an example to show the efficacy of the formulation.

\def\keywordstitle{Keywords}
\end{abstract}


\section{Introduction}
Tensegrity structures are networks of axially loaded compressive (bars or struts) or tensile (strings or cables) members [\cite{Snelson_1965}]. A ``class-k" tensegrity structure has a maximum of $k$ bars connected with ball joints, restricting any torque/moment transfer between the elements. Thus, a ``class-1" structure has all the compressive members floating with no two compression members touching each other [\cite{Skelton_2009_Tensegrity_Book}]. The unique property of tensegrity structures allowing to change the shape of the structure without changing its stiffness and changing the stiffness without changing the shape makes them suitable for soft robotics [\cite{Skelton_2009_Tensegrity_Book,Rieffel_softRobot}]. Tensegrity is also called as ``The Architecture of Life" due to its presence in tissues, viruses, cells, and even in humans [\cite{Ingber_1998}]. 
There have been a large number of studies on kinematic and static analysis for various topological considerations to support different loading conditions with minimum mass [\cite{Tibert_Pellegrino_2011_Review,Arsenault_Gosselin_2008_IJRR,Goyal_2020_MRC,Skelton_2010_Michell}].
%
%
The minimal mass architecture along with compliance and morphable shape characteristics makes tensegrity systems suitable for applications like planetary landers [\cite{Caluwaerts_2014, Goyal_2019_Buckling}], deployable space structures [\cite{Tibert_Pellegrino_2003_Mast,yang2019deployment,Peng_2018_DynamicDeploy}] and flexible robots [\cite{Sabelhaus_2017_ACC,SoftRobots2_IJRR,Karnan_Goyal_2017_IROS}]. The easy addition of redundant strings enables robust design approaches making the structure well suited for human-robot interaction. A theoretical framework for sensor placements and damage detection in tensegrity structures is a topic of recent research [\cite{Goyal_2019_SEMC}].

Tensegrity structures are most suitable to serve as morphing structures to achieve different shapes like tensegrity plates, domes, self-tunable antennas and wings [\cite{Edwin_plate,tibert2002_ReflectorSatellites,moored_MorphWing}]. However, much of the research in the past was done using static or kinematic analysis. The research in the last two decades shifted from static studies to dynamic analysis where several different approaches were used to develop dynamic models for tensegrity structures [\cite{Murakami_2001a_Dynamics,Goyal_Dynamics_2019,Peng_2018_AIAA}], which further led the exploration of various active control strategies [\cite{Varol_2019_IEEE_Access,Karnan_Goyal_2017_IROS}]. Learning/evolutionary algorithm based approaches [\cite{Caluwaerts_2014,Paul_2006,Zhang_Levine_2017_ICRA,IanSmith_RL,Bekris_2019_IJRR2}] and 
model-based approaches [\cite{Sabelhaus_2017_ACC,Karnan_Goyal_2017_IROS}] have been used in the control the tensegrity structures. A comparison between the model-based and data-based approach is an emerging topic of recent research [\cite{Wang_2020_RAL}]. Both of these active control strategies are useful for the deployment of tensegrity-based civil engineering structures [\cite{IanSmith_CivilControl,LandolfBarbarigos_Thesis}], and tensegrity robots [\cite{Bekris_2019_IJRR1,sabelhaus2015system,Koizumi_2012_ICRA,blissIwasakiCPGexperimental}].
In this paper, a model-based output regulation approach for the nonlinear shape control of a general ``class-k" tensegrity structure using control Lyapunov function is discussed. The advantage of having a very precise dynamic model due to the 1-dimensional (1D) deformation motion of each member is exploited in this model-based approach [\cite{Goyal_Dynamics_2019}]. 
Many other model-based control techniques are used for the control of tensegrity structures.
The Central Pattern Generator (CPG) controller mimics the periodic system in our biological neural circuits [\cite{blissIwasakiCPG,blissIwasakiCPGexperimental}]. Model Predictive Control (MPC) approaches have recently been used to control tensegrity structures [\cite{PENG_2020_MPC,Sabelhaus_2017_ACC}]. 
\cite{Yang_Sultan_2017_IJRNC} designed an adaptive controller for these underactuated systems and further used robust control techniques for the controlled deployment and stabilization using $\mathcal{H}_\infty$ control theory but specifically for the tensegrity membrane system. A shape controller using output feedback linear parameter varying formulation is developed in \cite{Yang_Sultan_2017_LPVControl}. 
A pose estimation strategy was recently developed to control the position of the end effector of a tensegrity manipulator [\cite{Varol_Manpltr}]. A framework for automatic robotic assembly of tensegrity structures by controlling the motion of a specific end-effector is also recently proposed [\cite{Varol_RoboPlan}]. The end-effector for this research was attached to a standard industrial robot [\cite{Varol_RoboPlan}]. 



Most of the dynamic compensators for controlling tensegrity robots are based on the kinematics of mechanical manipulators. These kinematic models typically involve transcendental functions. In some cases, approximate linear models either derived from data or physics are used. 
The control law designed in this paper is based on the recently developed matrix-based exact nonlinear dynamics model which assumes bars to be rigid and strings to be Hookean [\cite{Goyal_2020_GYRO}]. This exact dynamic model is free of trigonometric functions, making the trajectory design more efficient and implementation of the controller very simple. 
The dynamic model also allows for some or all bars in the structure to have small gyroscopic wheels. This is further shown to increase the reachable space of allowable final shapes of the structure.
The control algorithm to change the shape of the structure uses ``force density" as the control variable which is dependent on the axial deformation of string members in the formulated gyroscopic tensegrity systems.
The final equation to solve for the control variable at each step is obtained as a linear programming problem that can be solved very efficiently. 
This paper further develops the methodology to calculate feedback gains to put different bounds on the position errors for five different types of problems. 
The robustness to external disturbances is solved by posing them as feasibility problems by providing different kinds of bounds on the performance using an LMI formulation [\cite{Skelton_LMI_1998}].

\begin{figure}[h!]
    \centering
    \includegraphics[width=1\linewidth]{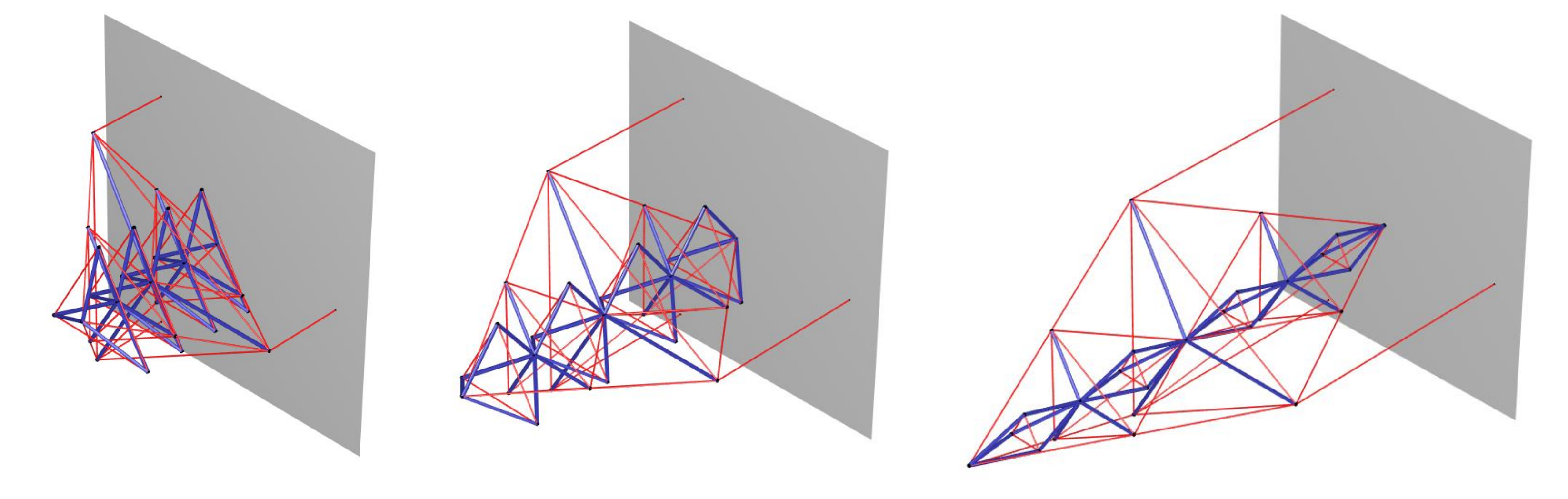}   
    \caption{Different configuration of tensegrity robotic arm for extension motion.}
    \label{f:Intro_fig}
\end{figure}

\begin{figure}[h!]
    \centering
    \includegraphics[width=1\linewidth]{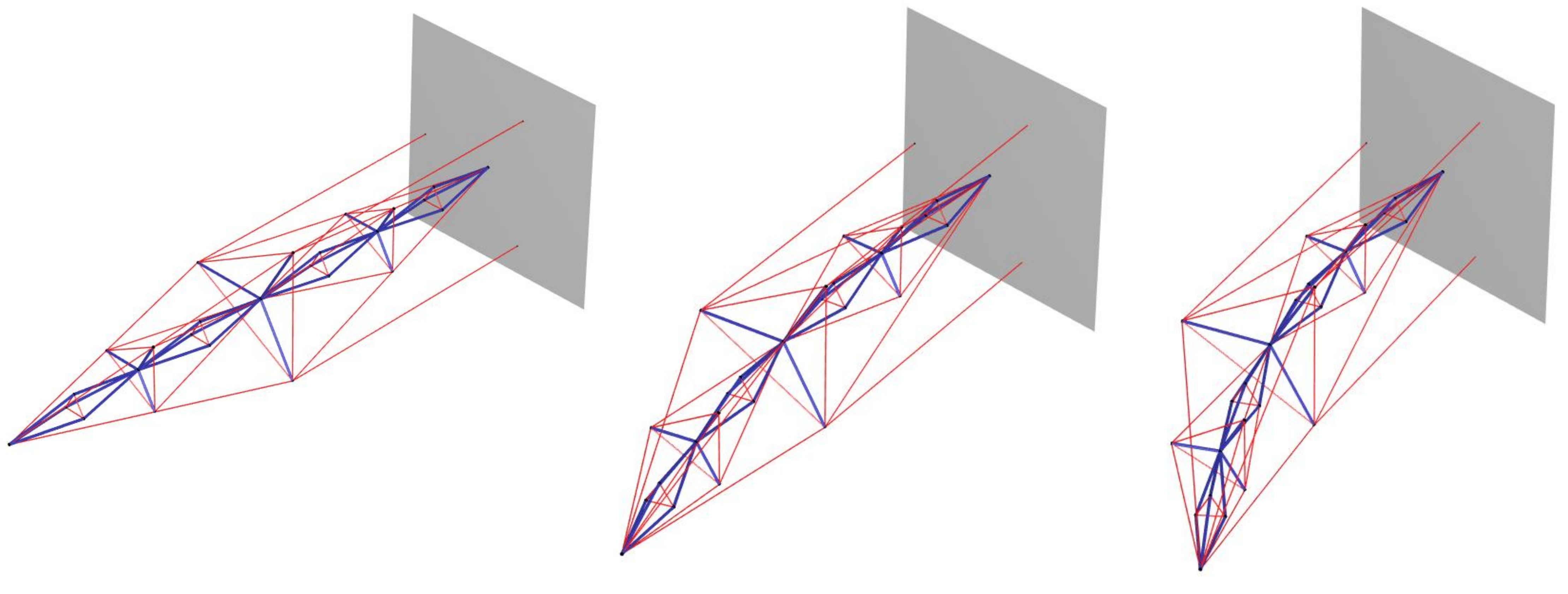}   
    \caption{Different configuration of tensegrity robotic arm for angular rotation motion.}
    \label{f:Intro_fig2}
\end{figure}

The organization of the paper is as follows: 
a brief description of the dynamic model of the gyroscopic tensegrity system is given in section 2. In section 3, a Lyapunov-based nonlinear controller is given to regulate the error in position to zero. The formulation for the shape control of the gyroscopic tensegrity system is derived, which allows to control the location of the nodes. The formulation to regulate the velocity and acceleration of the nodes is also given. Section 4 provides the vectorized formulation for nonlinear tensegrity dynamics in the presence of given disturbances. The vector form allowed us to find the control gains to bound different types of errors for the reduced-order controller. The LMI formulation is used to solve for the control gains to bound the errors for different types of disturbances along with a bound on the error in the length of the bars. A new deployable tensegrity robotic arm is introduced by combining T-bar and D-bar tensegrity structures [\cite{Skelton_2009_Tensegrity_Book}].  
The tensegrity $T_2D_1$ robotic arm example is used to show the extension from a stowed configuration, movement of the end effector of the arm to reach a desired given position in 3-dimensional space and the rotation of the end-effector about its own axis with the use of gyroscopic tensegrity system, which was not possible with the standard tensegrity model. Finally, the results for the disturbance rejection were shown for different kinds of performance requirements.

\section{Dynamics of gyroscopic tensegrity structures}
This section provides a brief overview of the nonlinear dynamic model of a general tensegrity structure. The final form of class-1 tensegrity dynamics is formulated as a second-order matrix differential equation [\cite{Goyal_2020_GYRO}], which can be written in compact matrix form with the following definitions of $M_s$, $K_s$, and $W_T$ as:
\begin{align}
\ddot{N}M_s + NK_s &= W_T, \label{eq:gyro_dyn_full}\\
\tau_B = \Hat{J_a} \hat{l}^2 \dot{\omega}_W,
\end{align}
where
\begin{align}
&M_s = \begin{bmatrix}
C_{nb}\T (C_b\T \hat{J}_t C_b + C_r\T \hat{m}_{t} C_r) & C_{ns}\T \hat{m}_s
\end{bmatrix},\\
&K_s = \begin{bmatrix}
C_s\T \hat{\gamma} C_{sb} - C_{nb}\T C_b\T \hat{\lambda} C_b & C_s\T\hat{\gamma}C_{ss}
\end{bmatrix},\\
&W_T = W + \begin{bmatrix}
(\tilde{B} \hat{\dot{B}}\hat{J_a}\hat{\omega}_w\hat{l}^{-1}- \tilde{B}\hat{\mathcal{T}}\hat{l}^{-2}) C_b & 0
\end{bmatrix},
\end{align}
$\tau_B = [\tau_{b1}~\tau_{b2} \cdots \tau_{b\beta}]\T$, $\dot{\omega}_W = [\dot{\omega}_{w1}~\dot{\omega}_{w2} \cdots \dot{\omega}_{w\beta}]\T$ and $\lambda$ represents the force density (compressive force per unit length) in the bars, given by:
\begin{align}
    \hat{\lambda} = -\hat{J}\hat{l}^{-2} \lfloor\dot{B}\T\dot{B} \rfloor - \frac{1}{2}\hat{l}^{-2}\lfloor B\T (W-N C_s\T\hat{\gamma}C_s)C_{nb}\T C_b\T \rfloor,
    \label{eq:Class_1_lambda}
\end{align}
and $N = \begin{bmatrix}n_1 & n_2 &\cdots& n_{2\beta+\sigma}\end{bmatrix} \in \mathbb{R}^{3 \times (2\beta+\sigma)}$ represents the matrix containing the node position vectors $n_i \in \mathbb{R}^{3 \times 1}$, $\beta$ represents the number of bars, and $\sigma$ represents the number of string-to-string nodes. 
The acceleration vector corresponding to the $i$\textsuperscript{th} node is represented by $\ddot{n}_i$, which forms the matrix $\ddot{N} = \begin{bmatrix} \ddot{n}_1 & \ddot{n}_2 &\cdots& \ddot{n}_{2\beta+\sigma}\end{bmatrix}$. The bar matrix $B = \begin{bmatrix}\vec{b}_1 & \vec{b}_2 & \cdots & \vec{b}_\beta\end{bmatrix} \in \mathbb{R}^{3 \times \beta}$ and string matrix $S= \begin{bmatrix}\vec{s}_1 & \vec{s}_2 & \cdots & \vec{s}_\alpha\end{bmatrix}\in \mathbb{R}^{3 \times \alpha}$ contain bar vectors $\vec{b}_i$ and string vectors $\vec{s}_i$, respectively. 
The diagonal matrices $\hat{m}_b $ and $\hat{m}_s$ are formed by arranging bar masses and string point masses along the diagonal elements, respectively, and $\hat{J}_a$ is a diagonal matrix with $J_{a_{ii}} = \frac{m_{b_i}}{12}+\frac{m_{b_i} r_{b_i}^2}{l_i^2}$ as the diagonal element that allows to accommodate the inertia in the bars. 
The connectivity matrices $C_{nb}, C_b, C_r, C_{ns}, C_{s}, C_{sb}$, and $C_{ss}$ define the connections between different nodes to form bar vector, string vectors and string-to-string node positions.
The external force matrix $W=\begin{bmatrix}w_1 & w_2 &\cdots& w_{2\beta+\sigma}\end{bmatrix}$ represents the matrix containing external force vector $\vec{w}_i$ corresponding to each node position vector $\vec{n}_i$. 
The term `force density in strings' is denoted by $\gamma_i$ to describe tensile force per unit length in the $i$\textsuperscript{th} string member and $\hat{\gamma}$ represents the diagonal matrix formed from $\gamma_i$ as its diagonal elements. 
Equation~(\ref{eq:Class_1_lambda}) provides the analytical formula to calculate the diagonal matrix $\hat{\lambda}$ where $\lfloor \circ \rfloor$ operator sets every off-diagonal element of the square matrix to zero. 
The diagonal matrix $\hat{l}$ is formed by the length of the bar members where $l_i$ denotes the length of the $i$\textsuperscript{th} bar. 
For class-$k$ tensegrity systems, the linear constraints of the form $NP=D$ can be added to the system, where $P \in \mathbb{R}^{(2\beta+\sigma) \times c}$ and $D  \in \mathbb{R}^{3 \times c}$ are specified such that constrained nodes coincide at all times with $c$ as the number of added constraints. The resulting dynamics with added Lagrange constraints can be written as:
\begin{equation}
\ddot{N}M_s + NK_s = W_T+\Omega P\T,
\label{eq:classk_dyn}
\end{equation}
where  $\Omega \in \mathbb{R}^{3 \times c}$ is a matrix of Lagrange multipliers satisfying the dynamics constraints at all time-steps. Please refer to \cite{Goyal_2020_GYRO} for a detailed derivation of this dynamics model.

\section{Shape control for gyroscopic tensegrity systems}\label{append:control}
This section provides the control law to control the shape for gyroscopic tensegrity systems, which can also be easily used for standard tensegrity structures by keeping the inertia and angular velocity of the gyroscopic wheels to zero. The benefit of using the gyroscopic tensegrity system over standard tensegrity is that it allows us to control the shape of the structure in a plane where even strings are not present or activated. In other words, this new system increases the reachability of the structure and allows for an increased actuation redundancy.

\subsection{Linear relation between force density for bars and strings}
One of the important steps in writing the control of this nonlinear dynamic system into a linear programming problem is to be able to write the force densities in the bar $\lambda = [\lambda_1~\lambda_2~\cdots~\lambda_\beta]\T$ in terms of the linear function of force densities in the strings $\gamma=[\gamma_1~\gamma_2~\cdots~\gamma_\alpha]\T$. This is required in order to write down control only as a function of force densities in the strings. Let us start by extracting the $i^{th}$ diagonal element of $\lambda$ by multiplying Eq.~(\ref{lin_lambda}) from the left by $e_i\T$ and from the right by $e_i$ as [\cite{Wang_2020_RAL}]:
\begin{multline}
    \lambda_i = -J_il_i^{-2} e_i\T\lfloor\dot{B}\T\dot{B} \rfloor e_i \\ - \frac{1}{2}l_i^{-2}e_i\T \lfloor B\T (W+\Omega P\T-S\hat{\gamma}C_s)C_{nb}\T C_b\T \rfloor e_i,
\end{multline}
which again after using the definition of the operator $\lfloor \circ \rfloor$ can be written as:
\begin{multline}
    \lambda_i = -J_il_i^{-2} ||\dot{b_i}||^2 +\frac{1}{2}l_i^{-2}b_i\T S\hat{\gamma}C_sC_{nb}\T C_b\T  e_i \\  - \frac{1}{2}l_i^{-2}b_i\T (W+\Omega P\T)C_{nb}\T C_b\T e_i . 
\end{multline}

Using the identity $\hat{x}y=\hat{y}x$ on the last term for x and y being the column vectors, we get:
\begin{multline}  
\lambda_i = -J_il_i^{-2} ||\dot{b_i}||^2 +\frac{1}{2}l_i^{-2}b_i\T S\reallywidehat{(C_sC_{nb}\T C_b\T  e_i)}\gamma \\ - \frac{1}{2}l_i^{-2}b_i\T (W+\Omega P\T)C_{nb}\T C_b\T e_i .
\end{multline}

Now, stacking all these scalars into a column gives:
\begin{align}
    \lambda = \Lambda \gamma + \tau, \label{eq:lambda_gamma}
\end{align}
where $\Lambda = \begin{bmatrix}
    \Lambda_1\T & \Lambda_2\T & \cdots & \Lambda_{\beta}\T  \end{bmatrix}\T$,  \\
$\tau = \begin{bmatrix} \tau_1\T & \tau_2\T & \cdots & \tau_{\beta}\T  \end{bmatrix}\T$, $\Lambda_i = \frac{1}{2}l_i^{-2}b_i\T S\reallywidehat{(C_sC_{nb}\T C_b\T  e_i)}$, and 
$\tau_i = -J_il_i^{-2} ||\dot{b_i}||^2  - \frac{1}{2}l_i^{-2}b_i\T (W+\Omega P\T)C_{nb}\T C_b\T e_i$, for $i = 1,2 ~ \cdots ~ \beta$.

This provides a linear relation between force density in the strings and force density of the bars.

\subsection{Stability analysis of the model-based controller for gyroscopic tensegrity structures}
This subsection provides the control algorithm to regulate the node position of specified nodes to achieve the desired shape. We start by defining the node positions for the nodes we want to control as $Y = LNR$, where $L \in \mathbb{R}^{n_l\times3}$ is a matrix that allows choosing of the $x,y$ or $z$ coordinates of the node and $R\in \mathbb{R}^{(2\beta+\sigma) \times n_r}$ matrix defines which nodes to be controlled. The matrix $\bar{Y}$ defines the final desired location of the nodes that we want to move from $Y$ to $\Bar{Y}$. The error in the positions at any time can be written as:
\begin{align}
    E = Y-\Bar{Y}, \hspace{.5pc}    E = LNR-\Bar{Y},
\end{align}
and its first and second derivatives with respect to time can be written as:
\begin{align}
    \dot{E} = L\dot{N}R, \hspace{.5pc} 
    \ddot{E} = L\ddot{N}R.
\end{align}

Now, let us write the dynamics of tensegrity systems in a general nonlinear matrix form as:
\begin{align}
    \dot{x} = f(x) + g(x) u \rightarrow \ddot{N} = f(N,\dot{N}) + g(N)\hat{(u)}h(N,\dot{N}), 
\end{align}
where $\hat{(u)}$ is the diagonal matrix made from the control input and the aim is to derive the states to some position $Y(t) \rightarrow \bar{Y}(t)$. 
Let us define the lyapunov function $V(N,\dot{N})$ as:
\begin{align}
    V = \frac{1}{2} \Tr(E\T \Theta E + \dot{E}\T \dot{E}) > 0 ~~ \forall ~ [E,~\dot{E}] \neq 0,
\end{align}
where $\Tr(\cdot)$ is the trace and the matrix $\Theta > 0$ is positive definite, which can be used to change the weights between error and error velocity. Let us write the first derivative w.r.t. time as:
\begin{align}
    \dot{V} = \Tr(\dot{E}\T \Theta {E} + \dot{E}\T \ddot{E}),
\end{align}
where the goal is to get the time derivative to be:
\begin{align}
    \dot{V} = -\Tr(\dot{E}\T \Psi \dot{E} )< 0, ~~ \Psi >0.
\end{align}

Substituting the above equation to previous equation, we get:
\begin{align}
    \Tr(\dot{E}\T \ddot{E} + \dot{E}\T \Psi \dot{E} + \dot{E}\T \Theta {E} ) = 0,
\end{align}
which after using the properties of trace operator, gives the final equation as:
\begin{align}
    \ddot{E}+\dot{E} \Psi+ E \Theta = 0. \label{eq:error_diff}
\end{align}

Notice that this gives a second-order differential equation in the error dynamics to derive the error to zero. The idea is to move the nodes from the current position to the desired position by aptly choosing the control gain parameters matrices $\Psi$ and $\Theta$. Now, we derive the final equations for the lyapunov controller mentioned earlier to generate the solution for the control as the linear programming problem. Let us start by substituting for $E, \dot{E}$ and $\ddot{E}$ in Eq.~(\ref{eq:error_diff}) to obtain:
\begin{align}
    L\ddot{N}R+ L\dot{N}R \Psi +  (LNR-\Bar{Y})\Theta=0.
\end{align}
Further substituting for $\ddot{N}$ from Eq.~(\ref{eq:gyro_dyn_full}), we get:
\begin{align}
    L(W_T+\Omega P\T- NK_s)M_s^{-1}R+ L\dot{N}R \Psi =  (\Bar{Y}-LNR)\Theta, 
\end{align}
\begin{align}    
    LNK_sM_s^{-1}R =L(W_T+\Omega P\T) M_s^{-1}R+ L\dot{N}R\Psi +  (LNR-\Bar{Y})\Theta.
\end{align}
Taking the $i^{th}$ column of the matrices from the above equation gives:
\begin{multline}
    LNK_sM_s^{-1}Re_i = L\dot{N}R \Psi e_i+  (LNR-\Bar{Y})\Theta e_i \\+ L \left(W + \begin{bmatrix}
(\tilde{B} \hat{\dot{B}}\hat{J_a}\hat{l}^{-1}\hat{\omega}_w) C_b & 0
\end{bmatrix}+\Omega P\T \right) M_s^{-1}R e_i.  \label{eq:control_first_gyro}
\end{multline}
Taking the left hand side and substituting for $K_s = C_s\T \hat{\gamma}C_s-C_{nb}\T C_b\T \hat{\lambda} C_bC_{nb}$, we get:
\begin{multline}
    LNK_sM_s^{-1}Re_i = LNC_s\T \hat{\gamma}C_sM_s^{-1}Re_i \\ - LNC_{nb}\T C_b\T \hat{\lambda} C_bC_{nb}M_s^{-1}Re_i,
\end{multline}
and using the identity $\hat{x}y=\hat{y}x$ for the right-hand side terms gives:
\begin{multline}
    LNK_sM_s^{-1}Re_i = LNC_s\T \reallywidehat{(C_sM_s^{-1}Re_i)}\gamma \\ -LNC_{nb}\T C_b\T \reallywidehat{(C_bC_{nb}M_s^{-1}Re_i)}\lambda. \label{eq:control_mid_gyro}
\end{multline}

Substituting for $\lambda$ in terms of $\gamma$ from Eq.~(\ref{eq:lambda_gamma}) gives:
\begin{multline}
    LNK_sM_s^{-1}Re_i = LNC_s\T \reallywidehat{(C_sM_s^{-1}Re_i)}\gamma \\ -LNC_{nb}\T C_b\T \reallywidehat{(C_bC_{nb}M_s^{-1}Re_i)}(\Lambda \gamma + \tau),  
\end{multline}    
which again can be written by combining the terms together as:
\begin{multline}
   \nonumber LNK_sM_s^{-1}Re_i = \Big(LNC_s\T \reallywidehat{(C_sM_s^{-1}Re_i)} \\ - LNC_{nb}\T C_b\T \reallywidehat{(C_bC_{nb}M_s^{-1}Re_i)}\Lambda\Big)\gamma - LNC_{nb}\T C_b\T \reallywidehat{(C_bC_{nb}M_s^{-1}Re_i)} \tau.
\end{multline}

Taking the right-hand side of Eq.~(\ref{eq:control_first_gyro}) and rearranging it gives: 
\begin{multline}
     L \left(W + \begin{bmatrix}
(\tilde{B} \hat{\dot{B}}\hat{J_a}\hat{l}^{-1}\hat{\omega}_w) C_b & 0
\end{bmatrix}+\Omega P\T \right) M_s^{-1}R e_i+ L\dot{N}R \Psi e_i \\ +  (LNR-\Bar{Y})\Theta  e_i  = L \begin{bmatrix}
(\tilde{B} \hat{\dot{B}}\hat{J_a}\hat{l}^{-1}\hat{\omega}_w) C_b & 0
\end{bmatrix} M_s^{-1}R e_i \\ + \left(L (W +\Omega P\T ) M_s^{-1}R+ L\dot{N}R \Psi+  (LNR-\Bar{Y})\Theta\right) e_i.  \label{eq:control_first_gyro_right}
\end{multline}

Now, taking out the angular speed of the wheels $\omega_w$ from the first term of the right-hand side:
\begin{multline}
    L \begin{bmatrix}
(\tilde{B} \hat{\dot{B}}\hat{J_a}\hat{l}^{-1}\hat{\omega}_w) C_b & 0
\end{bmatrix} M_s^{-1}R e_i \\
=  L \begin{bmatrix}
\tilde{B} \hat{\dot{B}}\hat{J_a}\hat{l}^{-1}\hat{\omega}_w C_b & 0
\end{bmatrix} \begin{bmatrix} M_{sb}^{-1} \\ M_{ss}^{-1} \end{bmatrix} R e_i \\
 = L \tilde{B} \hat{\dot{B}}\hat{J_a}\hat{l}^{-1}\hat{\omega}_w C_b M_{sb}^{-1} R e_i,
\end{multline}
and using the identity $\hat{x}y=\hat{y}x$ for the first term gives:
\begin{multline}
    L \tilde{B} \hat{\dot{B}}\hat{J_a}\hat{l}^{-1}\hat{\omega}_w C_b M_{sb}^{-1} R e_i
= L 
\tilde{B} \hat{\dot{B}}\hat{J_a}\hat{l}^{-1} \reallywidehat{(C_b M_{sb}^{-1} R e_i)} \omega_w. 
\end{multline}

Substituting all the terms back in Eq.~(\ref{eq:control_first_gyro}), we get:
\begin{multline}
    \Big(LNC_s\T \reallywidehat{(C_sM_s^{-1}Re_i)}-LNC_{nb}\T C_b\T \reallywidehat{(C_bC_{nb}M_s^{-1}Re_i)}\Lambda\Big)\gamma   \\ = \left(L (W +\Omega P\T ) M_s^{-1}R+ L\dot{N}R \Psi+  (LNR-\Bar{Y})\Theta\right) e_i \\ + LNC_{nb}\T C_b\T \reallywidehat{(C_bC_{nb}M_s^{-1}Re_i)} \tau +  L 
\tilde{B} \hat{\dot{B}}\hat{J_a}\hat{l}^{-1} \reallywidehat{(C_b M_{sb}^{-1} R e_i)} \omega_w . 
\end{multline}

Finally, stacking all the matrices on the left and vectors on the right, we get:
\begin{align}
    \begin{bmatrix}
    \Gamma_{p1} \\ \Gamma_{p2} \\ \vdots \\ \Gamma_{p_{n_r}}
    \end{bmatrix} \gamma = \begin{bmatrix}
    \mu_{p1} \\ \mu_{p2} \\ \vdots \\ \mu_{p_{n_r}}
    \end{bmatrix} + \begin{bmatrix}
    \Upsilon_{p1} \\ \Upsilon_{p2} \\ \vdots \\ \Upsilon_{p_{n_r}}
    \end{bmatrix} \omega_w, \hspace{1 pc} \gamma \geq 0, 
    \label{eq:final_control_pos}
\end{align}
where
\begin{align}
    \Gamma_{pi} &= LNC_s\T \reallywidehat{(C_sM_s^{-1}Re_i)}-LNC_{nb}\T C_b\T \reallywidehat{(C_bC_{nb}M_s^{-1}Re_i)}\Lambda,  \\
   \nonumber \mu_{pi} &= \Big(L(W+\Omega P\T) M_s^{-1}R+ L\dot{N}R \Psi+  (LNR-\Bar{Y}) \Theta \Big)e_i \\ &~~~~~~~~~~~~~~~~~~~~ + LNC_{nb}\T C_b\T \reallywidehat{(C_bC_{nb}M_s^{-1}Re_i)} \tau, \\
    \Upsilon_{pi} & =  L \tilde{B} \hat{\dot{B}}\hat{J_a}\hat{l}^{-1} \reallywidehat{(C_b M_{sb}^{-1} R e_i)}, \hspace{1pc}
    i = 1,2 ~ ... ~ n_r.
\end{align}

Equation~(\ref{eq:final_control_pos}) has $n_l \times n_r$ number of rows and $\sigma$ number of unknowns (control variables), assuming all the strings are activated, which can be solved as a linear programming problem with the linear constraint of $\gamma \geq 0$ due to strings inability to take compressive forces. 

\subsection{Controlling the velocity and acceleration}
The previous subsection detailed the control for node positions by writing the error in position as $E_p = L_p N R_p - \bar{Y}_p$ where subscript $p$ is used for the position. A second-order differential equation was then used to derive the error to zero, which resulted in a linear equation for the force densities in the string (refer Eq.~(\ref{eq:final_control_pos})). The same equation can again be written in a compact form as:
\begin{align}
    \Gamma_p \gamma = \mu_p + \Upsilon_p \omega_w, \hspace{1 pc} \gamma \geq 0.
    \label{eq:final_control_pos_compact}
\end{align}

To control the velocity of certain nodes, the error in the velocity of certain nodes can be written as:
\begin{align}
    E_v = L_v \dot{N} R_v - \bar{\dot{Y}}_v,
\end{align}
with a first-order differential equation to derive the error in velocity to zero:
\begin{align}
        \dot{E}_v+ E_v \Psi_v= 0,
\end{align}
where only first derivative of error $E_v$ is
used to find the linear relation for the control variable. The same approach discussed in the previous subsection is used to write the linear equation to control the nodal velocities as:
\begin{align}
    \begin{bmatrix}
    \Gamma_{v1} \\ \Gamma_{v2} \\ \vdots \\ \Gamma_{v_{n_r}}
    \end{bmatrix} \gamma = \begin{bmatrix}
    \mu_{v1} \\ \mu_{v2} \\ \vdots \\ \mu_{v_{n_r}}
    \end{bmatrix} + \begin{bmatrix}
    \Upsilon_{v1} \\ \Upsilon_{v2} \\ \vdots \\ \Upsilon_{v_{n_r}}
    \end{bmatrix} \omega_w, \hspace{1 pc} \gamma \geq 0, \label{eq:final_control_vel}
\end{align}
where 
\begin{align}
     \nonumber &\Gamma_{vi} = L_v NC_s\T \reallywidehat{(C_s M_s^{-1} R_v e_i)} \\ &~~~~~~~~~~~~~~~~ -L_v NC_{nb}\T C_b\T \reallywidehat{(C_bC_{nb} M_s^{-1} R_v e_i)}\Lambda, \\   \nonumber &\mu_{vi} = \Big(L_v(W+\Omega P\T) M_s^{-1}R_v+ (L_v\dot{N}R_v- \bar{\dot{Y}}_v) \Psi \Big)e_i \\ &~~~~~~ + L_v NC_{nb}\T C_b\T \reallywidehat{(C_bC_{nb}M_s^{-1}R_v e_i)} \tau ,
   \\  &~ \Upsilon_{v_i}  =  L_v \tilde{B} \hat{\dot{B}}\hat{J_a}\hat{l}^{-1} \reallywidehat{(C_b M_{sb}^{-1} R_v e_i)}, \hspace{1pc}
    i = 1,2 ~ ... ~ n_r,
\end{align}
which again can be written in the compact form as:
\begin{align}
    \Gamma_v \gamma = \mu_v + \Upsilon_v \omega_w, \hspace{1 pc} \gamma \geq 0.
        \label{eq:final_control_vel_compact}
\end{align}

Now, the error in the acceleration of the nodes can be written as:
\begin{align}
    E_a = L_a \ddot{N} R_a - \bar{\ddot{Y}}_a,
\end{align}
which can be directly converted to a linear equation in control variable by equating it to zero as $E_a = 0$. Using the same procedure, the linear programming problem to solve for the control variable $\gamma$ is written as:
\begin{align}
    \begin{bmatrix}
    \Gamma_{a1} \\ \Gamma_{a2} \\ \vdots \\ \Gamma_{a_{n_r}}
    \end{bmatrix} \gamma = \begin{bmatrix}
    \mu_{a1} \\ \mu_{a2} \\ \vdots \\ \mu_{a_{n_r}}
    \end{bmatrix}+ \begin{bmatrix}
    \Upsilon_{a1} \\ \Upsilon_{a2} \\ \vdots \\ \Upsilon_{a_{n_r}}
    \end{bmatrix} \omega_w, \hspace{1 pc} \gamma \geq 0, \label{eq:final_control_acc}
\end{align}
where 
\begin{align}
    \nonumber &\Gamma_{ai} = L_a NC_s\T \reallywidehat{(C_s M_s^{-1} R_a e_i)} \\ &~~~~~~~~~~~~~~~~  -L_a NC_{nb}\T C_b\T \reallywidehat{(C_bC_{nb} M_s^{-1} R_a e_i)}\Lambda, \\
   \nonumber &\mu_{ai} = \Big(L_a (W+\Omega P\T) M_s^{-1}R_a - \bar{\ddot{Y}}_a \Big)e_i \\ &~~~~~~~~~~~~~~~~  + L_a NC_{nb}\T C_b\T \reallywidehat{(C_bC_{nb}M_s^{-1}R_a e_i)} \tau ,
   \\ &   \Upsilon_{a_i}  =  L_a \tilde{B} \hat{\dot{B}}\hat{J_a}\hat{l}^{-1} \reallywidehat{(C_b M_{sb}^{-1} R_a e_i)}, \hspace{1pc}
    i = 1,2 ~ ... ~ n_r,
\end{align}
which again can be written in the compact form as:
\begin{align}
    \Gamma_a \gamma = \mu_a + \Upsilon_a \omega_w, \hspace{1 pc} \gamma \geq 0.
        \label{eq:final_control_acc_compact}
\end{align}

Finally, combining Eqs.~(\ref{eq:final_control_pos_compact}, \ref{eq:final_control_vel_compact}, and \ref{eq:final_control_acc_compact}) allows for the simultaneous control of the position, velocity and acceleration of different nodes in the structure as:
\begin{align}
    \begin{bmatrix}
    \Gamma_p \\ \Gamma_v  \\ \Gamma_a
    \end{bmatrix} \gamma = \begin{bmatrix}
    \mu_p \\ \mu_v \\ \mu_a
    \end{bmatrix} + \begin{bmatrix}
    \Upsilon_p \\ \Upsilon_v \\ \Upsilon_a
    \end{bmatrix}\omega_w, \hspace{1 pc} \gamma \geq 0.
    \label{eq:control_pva_compact}
\end{align}

\section{Vectorized equations for tensegrity dynamics with disturbance}
Let us write the second-order matrix differential equation describing dynamics of any tensegrity structure in the presence of some disturbance $W_{d}$ as:
\begin{align}
    \ddot{N}M_s + N K_s = W_T + W_{d},\label{eq:Dynamics_dstrb}
\end{align}
which can also be written using the kronecker product ($\otimes$) in the second-order vector differential equation as:
\begin{align}
    \mathcal{M} \ddot{n} + \mathcal{K} n = \mathcal{W} + w_d, \label{eq:vec_Nddot}
\end{align}
where 
$n = [n_{1x}, n_{2x},\cdots,n_{1y},n_{2y}, \cdots, n_{1z},n_{2z},\cdots]\T \in\mathbb{R}^{3(2\beta+\sigma)
 \times 1}$, $\mathcal{M} = (I_3 \otimes M_s)$, $\mathcal{K} = (I_3 \otimes K_s)$, $\mathcal{W} = [e_1\T W_T~e_2\T W_T~e_3\T W_T]\T$, and $w_d = [e_1\T W_d~e_2\T W_d~e_3\T W_d]\T $.

In class-$k$ tensegrity systems, multiple bars are connected (using ball joints) at a node such that there is no torque/moment transfer from one bar member to any other bar member. 
The dynamics of class-$k$ tensegrity structures (for $k > 1$) can be easily extended from the formulation developed in the previous section.
Each class-$k$ joint can be handled by creating $k-1$ virtual nodes which are constrained to coincide at all times with the use of Lagrange constraint forces. These constraints are written as:
\begin{align}
    A n = d,
    \label{eq:lin_constraint}
\end{align}
where $A\in\mathbb{R}^{N_c \times 3(2\beta+\sigma)
}$ and $N_c$ represents the number of constraints. The linear constraints will restrict the motion in certain directions and will add some constraint forces in the dynamics. Let us define the constraints forces as $f_c$ which satisfies $f_c\T \delta n = 0$ because of no virtual work condition and $A \delta n = 0$ due to motion constraints for any arbitrary displacement $\delta n$ as:
\begin{align}
    \begin{bmatrix}
    f_c\T \\ A \end{bmatrix} \delta n = 0,
\end{align}
The above equation should be satisfied for arbitrary displacement $\delta n$, which will be true if and only if the coefficient matrix will have rank deficiency. Hence, we can write:
\begin{align}
    f_c = A \T \omega,
\end{align}
where  $\omega \in\mathbb{R}^{N_c \times 1}$ represents the Lagrange multipliers vector. Adding this constraint force to the dynamics equation, we obtain:
\begin{align}
     \mathcal{M}\ddot{n}+\mathcal{K}n &= \mathcal{W} + w_d + A \T \omega \label{eq:lin_classk_dyn}.
\end{align}
Dividing $A$ as $A = [A_1~A_2~A_3]$, where $A_1,A_2,A_3 \in\mathbb{R}^{N_c \times (2\beta+\sigma)}$, we can write $\lambda$ as:
\begin{multline}
    \hat{\lambda} = -\hat{J}\hat{l}^{-2} \lfloor\dot{B}\T\dot{B} \rfloor \\ - \frac{1}{2}\hat{l}^{-2}\lfloor B\T (W+[A_1\T \omega~A_2\T \omega~A_3\T \omega]\T-S\hat{\gamma}C_s)C_{nb}\T C_b\T \rfloor.
    \label{lin_lambda}
\end{multline}
\subsection{Reduced-order dynamics}
Adding the linear constraints into the dynamics will restrict the motion in certain dimensions, thus reducing the order of the dynamics to a span a smaller space. To this end, we use the singular value decomposition (SVD) of matrix $A$ as:
\begin{align}
A=U\Sigma V\T=U \begin{bmatrix}
    \Sigma_1 & 0
    \end{bmatrix}\begin{bmatrix}
    V_1\T \\ V_2\T 
    \end{bmatrix} ,
\label{lin_PSVD}
\end{align}
where $U\in\mathbb{R}^{N_c\times N_c}$ and $V\in\mathbb{R}^{3(2\beta+\sigma)\times 3(2\beta+\sigma)}$ are both unitary matrices, $V_1\in\mathbb{R}^{3(2\beta+\sigma)\times N_c}$ and $V_2\in\mathbb{R}^{3(2\beta+\sigma)\times(3(2\beta+\sigma)-N_c)}$ are submatrices of $U$, and $\Sigma_1\in\mathbb{R}^{N_c\times N_c}$ is a diagonal matrix  of positive singular values. By defining:
\begin{align}
\eta=\begin{bmatrix} \eta_1 \\ \eta_2 \end{bmatrix} \triangleq V\T n = \begin{bmatrix} V_1\T n \\ V_2\T n \end{bmatrix}, 
\end{align}
the constraint equation (Eq.~(\ref{eq:lin_constraint})) can be modified as:
\begin{align}
A n =U\Sigma V\T n = U \begin{bmatrix}
    \Sigma_1 & 0 
    \end{bmatrix}     \begin{bmatrix} \eta_1 \\ \eta_2 \end{bmatrix}=d,
\end{align}
which implies:
\begin{align}
\eta_1=\Sigma_1^{-1} U\T d, \;\;\dot{\eta}_1=0,\;\;\ddot{\eta}_1=0.\label{lin_eta_1}
\end{align}
Here, $\eta_1$ represents the no-motion space in transformed coordinates. Moreover, $\eta_2$ will evolve according to the constrained dynamics in new coordinate system. Using Eqs.~(\ref{lin_PSVD}-\ref{lin_eta_1}), the dynamics equation (Eq.~(\ref{eq:lin_classk_dyn})) can be rewritten as:
\begin{align}
\mathcal{M}V V\T \ddot{n}+\mathcal{K} V V\T n = \mathcal{W} + w_d + A \T \omega, \\
\mathcal{M}V_2 \ddot{\eta_2}+\mathcal{K} V_1 \eta_1+\mathcal{K} V_2 \eta_2 = \mathcal{W} + w_d + V_1 \Sigma_1 U\T \omega.
\end{align}
Pre-multiplying the above equation by a non-singular matrix $[V_2~~\mathcal{M}^{-1} V_1 ]\T$ will  yield two parts, where first part gives the second order differential equation for the reduced dynamics:
\begin{align}
V_2\T \mathcal{M}V_2 \ddot{\eta_2}+V_2\T \mathcal{K} V_2 \eta_2 &= V_2\T \mathcal{W} + V_2\T w_d -V_2\T \mathcal{K} V_1 \eta_1,\\
\Rightarrow \mathcal{M}_2\ddot{\eta}_2+\mathcal{K}_2\eta_2&=\widetilde{\mathcal{W}} + \widetilde{w}_d.
\label{lin_RedDynU1}
\end{align}
with $\mathcal{M}_2 = V_2\T \mathcal{M}V_2$ and $\mathcal{K}_2 = V_2\T \mathcal{K}V_2$, and the second part gives an algebraic equation that is used to solve for the Lagrange multiplier:
\begin{multline}
 V_1\T \mathcal{M}^{-1}\mathcal{M}V_2 \ddot{\eta_2}+V_1\T \mathcal{M}^{-1}\mathcal{K} V_1 \eta_1+V_1\T \mathcal{M}^{-1}\mathcal{K} V_2 \eta_2 \\ = V_1\T \mathcal{M}^{-1} (\mathcal{W} + w_d) + V_1\T \mathcal{M}^{-1}V_1 \Sigma_1 U\T \omega ,
\end{multline}
\begin{multline}
V_1\T \mathcal{M}^{-1}\mathcal{K} V_1 \eta_1+V_1\T \mathcal{M}^{-1}\mathcal{K} V_2 \eta_2- V_1\T \mathcal{M}^{-1}V_1 \Sigma_1 U\T \omega  \\= V_1\T \mathcal{M}^{-1} (\mathcal{W} + w_d) ,
\end{multline}
\begin{align}
\Rightarrow V_1\T \mathcal{M}^{-1}\mathcal{K}n - V_1\T \mathcal{M}^{-1}A\T \omega & = V_1\T \mathcal{M}^{-1} (\mathcal{W} + w_d).
\label{lin_RedDynU2}
\end{align}
Notice that $\mathcal{K}$ is also a function of $\omega$, making it a linear algebra problem, the solution for which can be solved following the same approach as discussed in the appendix of \cite{Goyal_Dynamics_2019}.

\subsection{Controller for reduced-order dynamics model with disturbance}
The error in the position of the nodes which is desired to be zero is defined as:
\begin{align}
    e = \mathcal{L}n-\bar{n} = \mathcal{L}(V_1 \eta_1+V_2 \eta_2)-\bar{n}. \label{eq:red_edef}
\end{align}

The second-order output equation can be written as:
\begin{align}
    \ddot{e}+\Psi \dot{e}+ \Theta e = B_1 w_d, \label{eq:red_eddot}
\end{align}    

which can be written in state-space form as:
\begin{align}
    \begin{bmatrix} \dot{e} \\ \ddot{e} \end{bmatrix}& = \underbrace{\begin{bmatrix} 0 & I \\ -\Theta & -\Psi \end{bmatrix}}_{A_{cl}} \begin{bmatrix} e \\ \dot{e} \end{bmatrix} + \underbrace{\begin{bmatrix} 0 \\ B_1 \end{bmatrix}}_{B_{cl}} w_d. 
\end{align}

We need to find the controller gain parameters $\Psi$ and $\Theta$, which we write in controller gain matrix $G$ as:
\begin{align}
    G = \begin{bmatrix}  -\Theta & -\Psi \end{bmatrix}, \label{eq:Gain_close}
\end{align}
and decompose the matrix $A_{cl}$ as:
\begin{align}
    \begin{bmatrix} 0 & I \\ -\Theta & -\Psi \end{bmatrix} = \underbrace{\begin{bmatrix} 0 & I \\ 0 & 0 \end{bmatrix}}_{A_p} + \underbrace{\begin{bmatrix} 0 \\ I \end{bmatrix}}_{B_p} \begin{bmatrix}  -\Theta & -\Psi \end{bmatrix},
\end{align}
which allows us to write the closed-loop dynamics as:
\begin{align}
\dot{x} = (A_p + B_p G) x + B_{cl} w_d, \hspace{0.5pc} y = Cx. \label{eq:sys_Gain_close}
\end{align}

Now, Eq.~(\ref{eq:red_eddot}) after substitution from Eq.~(\ref{eq:red_edef}) can be written as:
\begin{align}
    \mathcal{L} V_2 \ddot{\eta_2}+\Psi \mathcal{L} V_2 \dot{\eta_2}+ \Theta (\mathcal{L} V_1 \eta_1+ \mathcal{L} V_2 \eta_2 - \bar{n}) = B_1 w_d, 
\end{align}    
where $\dot{\eta_1} = \ddot{\eta_1} = 0$ was used from the dynamics model formulation and which after multiplying Eq.~(\ref{lin_RedDynU1}) from left hand side by $\mathcal{M}_2^{-1} $ ($\mathcal{M}_2^{-1} \times$ Eq.~(\ref{lin_RedDynU1})) and substitution gives:
\begin{multline}
   B_1 w_d = \mathcal{L}V_2 \mathcal{M}_2^{-1}  (\widetilde{\mathcal{W}}  +\widetilde{w}_{vec} - \mathcal{K}_2 \eta_2) \\ +\Psi \mathcal{L} V_2 \dot{\eta_2} + \Theta (\mathcal{L} V_1 \eta_1+ \mathcal{L} V_2 \eta_2 - \bar{n}).
\end{multline}

Let us choose $B_1 = \mathcal{L}V_2 \mathcal{M}_2^{-1} V_2\T$ (without loss of generality) to write:
\begin{multline}
  \mathcal{L}V_2 \mathcal{M}_2^{-1}  (\widetilde{\mathcal{W}} - \mathcal{K}_2 \eta_2)+\Psi \mathcal{L} V_2 \dot{\eta_2} \\=  \Theta (\bar{n} - \mathcal{L} V_1 \eta_1 - \mathcal{L} V_2 \eta_2 ) , 
\end{multline}
\begin{multline}  
  \mathcal{L}V_2 \mathcal{M}_2^{-1}  (V_2\T \mathcal{W}  -V_2\T \mathcal{K} V_1 \eta_1 - \mathcal{K}_2 \eta_2)+\Psi \mathcal{L} V_2 \dot{\eta_2} \\ =  \Theta (\bar{n} - \mathcal{L} V_1 \eta_1 - \mathcal{L} V_2 \eta_2 ) .
\end{multline}

Let us write $\mathcal{M}_{sn} = V_2\mathcal{M}_2^{-1} V_2\T$ and substitute for $\mathcal{K}_2$ from Eq.~(\ref{lin_RedDynU1}) to obtain:
\begin{multline}
    \mathcal{L} \mathcal{M}_{sn} \left[\left(I_3 \otimes (C_s\T \hat{\gamma}C_s) \right) -\left(I_3 \otimes (C_{nb}\T C_b\T \hat{\lambda}C_b C_{nb}) \right)  \right]n \\ =   \underbrace{\mathcal{L} \mathcal{M}_{sn} \mathcal{W} +\Psi \mathcal{L} V_2 \dot{\eta_2} + \Theta (\mathcal{L} V_1 \eta_1+ \mathcal{L} V_2 \eta_2 - \bar{n})}_{\mathds{C}},
\end{multline}
which after using the properties of kronecker product, can further be written as:
\begin{multline}
    \mathcal{L} \mathcal{M}_{sn} (I_3 \otimes C_s\T) (I_3 \otimes \hat{\gamma}) (I_3 \otimes C_s) n \\ = \mathcal{L} \mathcal{M}_{sn} (I_3 \otimes C_{nb}\T C_b\T) (I_3 \otimes \hat{\lambda}) (I_3 \otimes C_b C_{nb})  n + \mathds{C}.
\end{multline}

Now, recognizing that the term $(I_3 \otimes C_s) n$ is a vector and $(I_3 \otimes \hat{\gamma})$ is a diagonal matrix, we use the property $\hat{x}y=x\hat{y}$ to get:
\begin{multline}
    \mathcal{L} \mathcal{M}_{sn} (I_3 \otimes C_s\T)  \reallywidehat {\left( (I_3 \otimes C_s) n \right)} (\mathds{1} \otimes \hat{\gamma}) \\ = \mathcal{L} \mathcal{M}_{sn} (I_3 \otimes C_{nb}\T C_b\T) \reallywidehat {\left( (I_3 \otimes C_b C_{nb}) n \right)} (\mathds{1} \otimes \hat{\lambda}) + \mathds{C},
\end{multline}
where $\mathds{1}  \triangleq [1~1~1]\T$. The above equation can again be written as:
\begin{multline}
    \underbrace{\left[ \mathcal{L} \mathcal{M}_{sn} (I_3 \otimes C_s\T)  \reallywidehat {\left( (I_3 \otimes C_s) n \right)} (\mathds{1} \otimes I_{\alpha}) \right]}_{\mathds{A}} \gamma \\= \underbrace{\left[ \mathcal{L} \mathcal{M}_{sn} (I_3 \otimes C_{nb}\T C_b\T) \reallywidehat {\left( (I_3 \otimes C_b C_{nb}) n \right)} (\mathds{1} \otimes I_{\beta}) \right]}_{\mathds{B}} \lambda + \mathds{C}.
\end{multline}

Let us use Eq.~(\ref{eq:lambda_gamma}) to write the above mentioned equation as:
\begin{align}
  \mathds{A} \gamma = \mathds{B} (\Lambda \gamma + \tau) + \mathds{C},
\end{align}
which can be re-written after combining terms for force density $\gamma$ to generate a linear equation as:
\begin{align}
  (\mathds{A} - \mathds{B} \Lambda) \gamma = \mathds{B} \tau + \mathds{C}.
\end{align}
The above equation along with the positive force density constraint $(\gamma \geq 0)$ can be solved as a \textit{linear programming problem} to reject the disturbance for some choice of gain matrix $G$. 
Note that while other ways to manage redundancy exist, the linear program is especially useful in tensegrity since it maintains the positivity of the force density in tension members.

\section{Different error bounds for various disturbances}
In the previous sections, the gains for the second-order differential equations were chosen to stabilize the output differential equation i.e., to derive the errors to zero. However, no performance criteria was discussed in selecting the gains $G = \begin{bmatrix}  -\Theta & -\Psi \end{bmatrix}$ for the system $\dot{x} = (A_p + B_p G) x + B_{cl} w_d, \hspace{0.5pc} y = Cx$. This section solves this issue by formulating five different problems for bounding errors using the LMI framework. The gains for these five different problems are calculated using the semi-definite convex programming problem. 

\subsection{Bound on $\mathcal{L}_\infty$ norm of error or generalized $\mathcal{H}_2$ problem}
The peak value of a variable in the time domain is defined as $\mathcal{L}_\infty$ norm of the variable, i.e. $\|{y}\|_{\mathcal{L}_\infty}^2 = \sup [{ y}(t)\T y(t)]$. The following result provides a bound on peak value such that $\|{ y}\|_{\mathcal{L}_\infty}<\epsilon $, meaning that the peak value of
$[{ y}(t)\T{ y}(t)]$ is less than $\epsilon^2$ in the presence of finite energy disturbance. This problem can be solved as a ``energy to peak gain - $\Gamma_{ep}$" [\cite{Skelton_LMI_1998}] or generalized $\mathcal{H}_2$ problem [\cite{Scherer_LMI_1997}].
\begin{align}
    \Gamma_{ep} &\triangleq \sup_{\|w\|_{\mathcal{L}_2} \leq 1} \|y\|_{\mathcal{L}_\infty},\\
    \Gamma_{ep} = \inf_Q \| CQC\T \|^{1/2}:& A_{cl}Q+QA_{cl}\T + B_{cl}B_{cl}\T < 0, ~ Q>0. \label{eq:Linf_gain}
\end{align}

\begin{lemma}\label{Lem1}
The controller gain matrix $G$ (c.f. Eq.~(\ref{eq:Gain_close})) for system given in Eq.~(\ref{eq:sys_Gain_close}), which provides a $\mathcal{L}_\infty$ bound ($\Gamma_{ep} < \epsilon$) on the error in the desired position can be solved as:
\begin{align}
    &\min \epsilon, \begin{bmatrix} \epsilon I & CQ \\ QC\T & Q \end{bmatrix} > 0, \\ 
    &\begin{bmatrix} sym(A_p Q + B_p R) & B_{cl} \\ B_{cl}\T & -I \end{bmatrix} < 0, 
\end{align}
where $G = R Q^{-1}$.
\end{lemma}

\noindent \textit{Proof} is given in appendix. 

\subsection{Bounded $\Gamma_{ee}$ or $\mathcal{H}_\infty$ problem}
This subsection provides the result to bound the peak value of the frequency response of the transfer function $T(s) \triangleq  C(sI-A_{cl})^{-1}B_{cl}$. The $\mathcal{H}_\infty$ Problem is defined as [\cite{IWASAKI_Hinf_1994, Gahinet_Apkarian_1994}]:
\begin{align}
     \|T\|_{\mathcal{H}_\infty} \triangleq \sup_w \|T(jw)\|< \epsilon
\end{align}
which can also be understood in time domain analysis as the energy-to-energy gain problem [\cite{Skelton_LMI_1998}]:
\begin{align}
    \Gamma_{ee} \triangleq \sup_{\|w\|_{\mathcal{L}_2} \leq 1} \|y\|_{\mathcal{L}_2}< \epsilon .
\end{align}

\begin{lemma}\label{Lem2}
The controller gain matrix $G$ (c.f. Eq.~(\ref{eq:Gain_close})) for system given in Eq.~(\ref{eq:sys_Gain_close}), which provides a $\mathcal{H}_\infty$ bound ($\Gamma_{ee} < \epsilon$) on the error in the desired position can be solved as:
\begin{multline}
   \begin{bmatrix} sym(A_pY+B_p L) & B_{cl} & YC\T \\ B_{cl}\T & -R & 0 \\ CY & 0 & -I \end{bmatrix} < 0,\\  Y>0, ~  R = \epsilon^2 I, ~ G = L Y^{-1}.
\end{multline}
\end{lemma} 

\noindent \textit{Proof} is given in appendix.

\subsection{Bounded $\Gamma_{ie}$ or LQR problem}
We define the linear quadratic regulator (LQR) problem to provide a performance bound $\epsilon > 0$ on the integral squared output such that $\|{y}\|_{\mathcal{L}_2} < \epsilon $ for any vector
${w}_0$ such that ${w}_0\T {w}_0 \leq 1$, and ${x}_0=0$. The disturbance $w$ is the impulsive disturbance ${w}(t) = { w}_0{\delta}(t)$. This can also be defined as the peak disturbance to energy gain ($\|{y}\|_{\mathcal{L}_2}$) for the system [\cite{Skelton_LMI_1998}].
\begin{align}
\Gamma_{ie} &\triangleq \sup_{{w}_0{\delta}(t) \leq 1} \|y\|_{\mathcal{L}_2},\\
    \Gamma_{ie} = \inf_P \| B_{cl}\T P B_{cl} \|^{1/2}:& PA_{cl}+A_{cl}\T P + C\T C < 0, ~ P>0. \label{eq:IE_gain}
\end{align}

\begin{lemma}\label{Lem3}
The controller gain matrix $G$ (c.f. Eq.~(\ref{eq:Gain_close})) for system given in Eq.~(\ref{eq:sys_Gain_close}), which provides a bound on the error $\Gamma_{ie} < \epsilon$ from the desired position can be solved as:
\begin{align}
    &\min \epsilon,  \hspace{0.5pc} \begin{bmatrix} \epsilon I & B_{cl}\T \\ B_{cl} & Y \end{bmatrix} > 0, 
    \\
    & \begin{bmatrix} sym(A_p Y + B_p R) & Y C\T \\ CY & -I \end{bmatrix} < 0,
\end{align}
where $G = R Y^{-1}$.
\end{lemma}

\noindent \textit{Proof} is given in appendix. 

\subsection{Bound on covariance in position error}
It is impossible to derive the error to precise zero in the presence of process noise; however, one can control the statistics of the error given the statistics of the noise [\cite{Skelton_LMI_1998,Scherer_LMI_1997}]. This subsection provides the required controller gain matrix to bound the covariance of the error in the position or velocity of the nodes.
\begin{align}
     \dot{x} = (A_p + B_p G) x + B_{cl} w, \hspace{0.5pc} y = Cx, \\
     \mathcal{E}[y y \T] = Y = CXC\T < \bar{Y}.
\end{align}

\begin{lemma}\label{Lem4}
The covariance bound on output ($Y = CXC\T < \bar{Y}$) can be achieved with the following choices of controller gain matrix $G$ (c.f. Eq.~(\ref{eq:Gain_close})) for system given in Eq.~(\ref{eq:sys_Gain_close}) for the zero-mean white noise of intensity $\mathcal{E}[w w \T] = \mathbb{W}$ for $X>0$:
\begin{align}
    &~~~~~~~~~~\begin{bmatrix} \bar{Y} & CX \\ XC\T & X \end{bmatrix} > 0,  \hspace{.5pc} 
    \\ &\begin{bmatrix} sym(A_p X + B_p R) & B_{cl} \\ B_{cl}\T & -\mathbb{W}^{-1} \end{bmatrix} < 0, 
\end{align}
where $G = R X^{-1}$.
\end{lemma}

\noindent \textit{Proof} is given in appendix.

\subsection{Stabilizing control}
A simple requirement to effectively control the shape of the tensegrity structure is to stabilize the second-order output feedback differential equation (Eq.~(\ref{eq:red_eddot})). This can be achieved by making the matrix $(A_p + B_p G) < 0 $ Hurwitz or $ A_{cl}X+XA_{cl}\T  < 0,~ X>0,$ for the following equation:
\begin{align}
     \dot{x} = (A_p + B_p G) x + B_{cl} w, \hspace{0.5pc} y = Cx.
\end{align}

\begin{lemma}\label{Lem5}
The controller gain matrix $G$ (c.f. Eq.~(\ref{eq:Gain_close})) for system given in Eq.~(\ref{eq:sys_Gain_close}), that will yield a stable controller can be solved as:
\begin{align}
    (A_p X + B_p R) +(A_p X + B_p R )\T < 0,
\end{align}
where $G = R X^{-1}$.
\end{lemma}


\subsection{Bounds on the bar length error}
The idea here is to use extra information provided by the bar length constraints to calculate the control gains for shape control. The formulation provided here allows us to add some convex constraints to choose the control gains to bound the $\mathcal{L}_\infty$ norm of the error in the length of the bars. The LMI constraints can be added with other convex constraints discussed in the previous section to reject different kinds of disturbances in the control of the tensegrity structure.
Let us start by writing a bar vector for the $i$\textsuperscript{th} bar as:
\begin{align}
    b_i = (C_{b_i} \otimes I)n = (C_{b_i} \otimes I) (e+\bar{n})  = \mathcal{C}_{b_i} (e+\bar{n}).
\end{align}

Now, the desired performance to bound the error in bar length $l_i$ for a particular compressive bar member ($\bar{b}_i = \mathcal{C}_{b_i} \bar{n}$) can be written as:
\begin{align}
    z_i = \| \mathcal{C}_{b_i} (e+\bar{n}) \| - l_i &= \| \mathcal{C}_{b_i} e+\bar{b}_i \| - l_i,  \\
     z_i = e\T \mathcal{C}_{b_i}\T \mathcal{C}_{b_i} e &+ 2 \bar{b}_i \T \mathcal{C}_{b_i} e. 
\end{align}

Let us define $y_i \triangleq \mathcal{C}_{b_i} e$ and let us write the peak value of the error in the bar-length can be bound as:
\begin{align}
    \|z_ i\|_{\mathcal{L}_\infty}< \epsilon_{b_i}, \\ \|y_i \T y_i + 2 \bar{b}_i \T y_i \|_{\mathcal{L}_\infty}< \epsilon_{b_i},
\end{align}
where $ \epsilon_{b_i}$ is the maximum allowed value of the bar length error for the bar $b_i$. 
It is important to notice that the actual problem at hand is to bound the ($y_i \T y_i + 2 \bar{b}_i \T y_i $), which can be done by bounding the peak value of ($y_i\T y_i$) using the formulation discussed further.
Now, the aim here is to bound the maximum value of the $y_i$ for all time as this is an easier problem to tackle, known as generalized $\mathcal{H}_2$ problem or energy to peak gain $\Gamma_{ep}$ problem.

\begin{lemma}
The S-Procedure or S-Lemma [\cite{boyd2004convex,S-Lemma_2007}]: 
Let $A_1$ and $A_2$ be symmetric matrices, $b_1$ and $b_2$ be vectors and $c_1$ and $c_2$ be real numbers. Assume that there is some $x_0$ such that the strict inequality  ${\displaystyle x_{0}^{T}A_{1}x_{0}+2b_{1}^{T}x_{0}+c_{1}<0}$ holds. Then the implication
\begin{align}
x^{T}A_{1}x+2b_{1}^{T}x+c_{1}\leq 0\Longrightarrow x^{T}A_{2}x+2b_{2}^{T}x+c_{2}\leq 0,
\end{align}
holds if and only if there exists some nonnegative number $\lambda$ such that
\begin{align}
 \lambda {\begin{bmatrix}A_{1}&b_{1}\\b_{1}^{T}&c_{1}\end{bmatrix}}-{\begin{bmatrix}A_{2}&b_{2}\\b_{2}^{T}&c_{2}\end{bmatrix}} > 0.
 \end{align}
\end{lemma}~\\

\noindent Using the standard S-Procedure or S-Lemma [\cite{S-Lemma_2007}] given above, we can write the following for our problem:
\begin{align}
    y_i \T y_i < \bar{\epsilon}_{b_i} &\Rightarrow y_i \T y_i + 2 \bar{b}_i \T y_i  < \epsilon_{b_i} \\ &\Updownarrow
     \\
     \lambda \geq 0, ~ \lambda \begin{bmatrix} I & 0 \\ 0 & -\bar{\epsilon}_{b_i} \end{bmatrix}& - \begin{bmatrix} I & \bar{b}_i \\ \bar{b}_i \T & -{\epsilon}_{b_i} \end{bmatrix} \geq 0, 
\end{align}
which can again be written as the following to minimize the maximum value of error in the bar length:
\begin{align}
    \min ~\epsilon_{b_i},~ \lambda \geq 0,~  \begin{bmatrix} (\lambda-1)I &  -\bar{b}_i \\ -\bar{b}_i \T & {\epsilon}_{b_i} - \lambda \bar{\epsilon}_{b_i} \end{bmatrix} \geq 0.
\end{align}
The above mentioned LMI for variables $\epsilon_{b_i}$ and $\lambda$ and given value of $\bar{\epsilon}_{b_i}$ and $\bar{b}_i$ can be solved along with the LMIs mentioned earlier for bounding different kinds of errors. Another way to write this equation is by defining $\kappa = 1/\lambda$ and then writing the LMI as:
\begin{align}
    \max ~\bar{\epsilon}_{b_i},~ \kappa \geq 0,~  \begin{bmatrix} (1-\kappa)I &  -\kappa\bar{b}_i \\ -\kappa\bar{b}_i \T & \kappa {\epsilon}_{b_i} -  \bar{\epsilon}_{b_i} \end{bmatrix} \geq 0.
\end{align}
Notice that this LMI allows us to calculate the \textit{maximum value} of bound to be put on ($y_i \T y_i < \bar{\epsilon}_{b_i} $) by defining the desired bound on the bar length error ($y_i \T y_i + 2 \bar{b}_i \T y_i < {\epsilon}_{b_i}$).

\section{Results for the tensegrity robotic arm}
\begin{figure}[ht!]
    \centering
    \includegraphics[width=9cm,height=7cm,keepaspectratio]{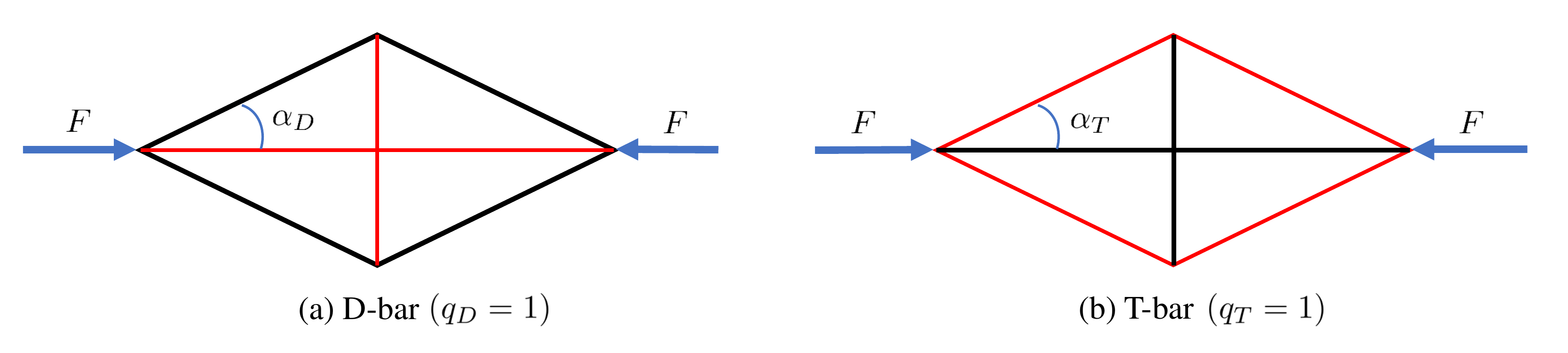}
    \caption{2D representation of D-bar and T-bar structure with bars shown in black and strings shown in red.}
    \label{TandD_bar}
\end{figure}

A traditional tensegrity D-bar structure of complexity ($q_D=1$) consists of 4 compressive members (bars) and 2 tensile members (strings) as shown in Figure~\ref{TandD_bar}(a) where all the compressive members are connected through ball joints [\cite{Skelton_2009_Tensegrity_Book}]. The vertical string in the structure supports the compressive loading and the horizontal string is needed to provide a self-equilibrium state and deployability [\cite{Goyal_2019_Buckling}]. The structure got his name from the ``Diamond (D)'' shape its compressive members form. 
A tensegrity T-bar structure of complexity ($q_T = 1$) contains 4 compressive members (bars) and 4 tensile members (strings). The four compressive members are constrained at the center with a spherical joint to form a ``T" shape structure, as shown in Fig.~\ref{TandD_bar}(b) [\cite{Skelton_2009_Tensegrity_Book}]. Strings are connected along the periphery to avoid global buckling in the structure by carefully optimizing the prestress in the strings and T-bar angle $\alpha_T$ [\cite{Goyal_2020_MRC}]. 

The concept of self-similar iterations is used to provide better mass efficiency as every compressive member in the structure can be replaced with the structure itself. This works for both D-bar and T-bar structure yielding higher complexities to minimize the mass of the structure to take a given load. 
Tensegrity T-bar structure has been shown to provide better mass efficiency than a D-bar structure [\cite{Goyal_2020_MRC}]. 
Please refer to \cite{Skelton_2009_Tensegrity_Book} for more details. However, D-bar provides a special feature of deployability making it more suitable for robotic applications.
In this study, we combine the mass efficiency of the T-bar and with deployability of the D-bar to make a dexterous tensegrity robotic arm. 

\begin{figure}[ht!]
    \centering
    \includegraphics[width=8cm,height=5cm,keepaspectratio]{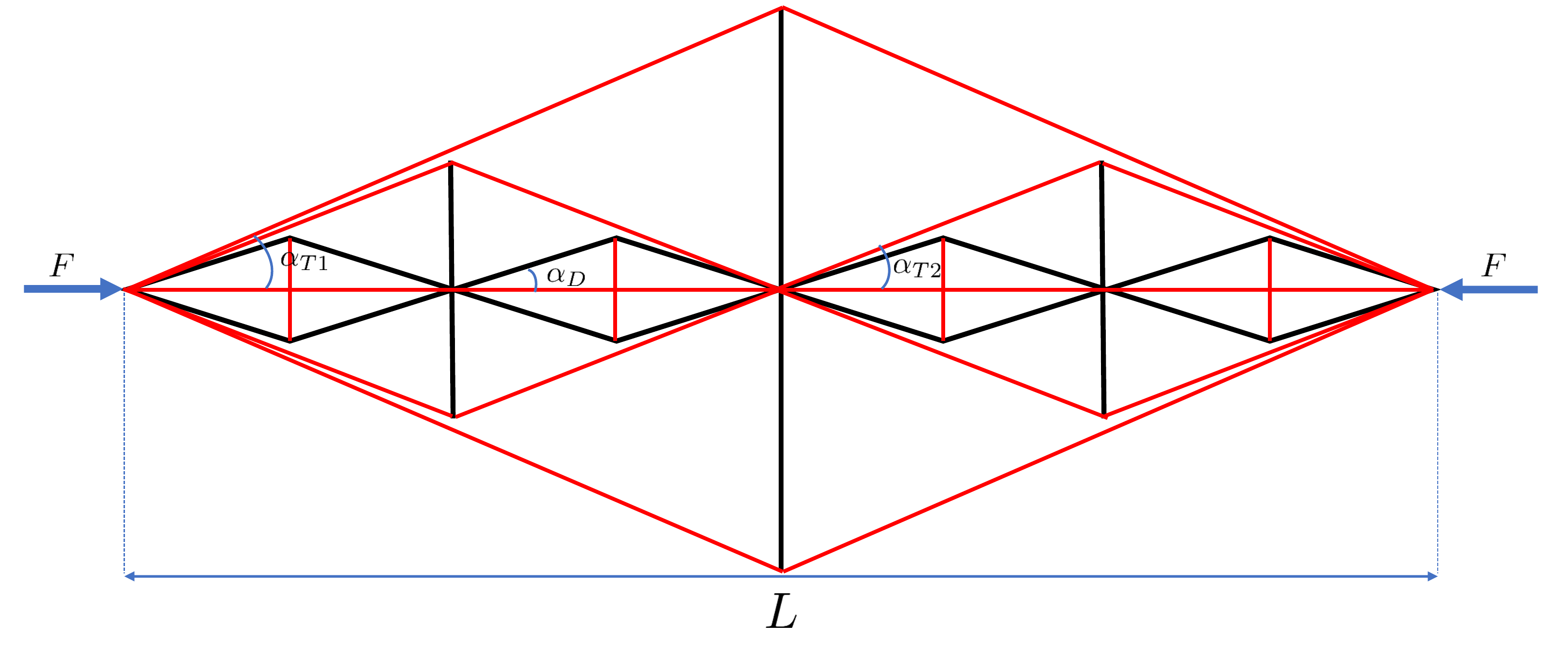}  
    \caption{2D representation of $T_2D_1$ tensegrity structure.}
    \label{TD_2D}
\end{figure}

To define our design of the tensegrity robotic arm, we use the concept of fractals or self-similar iterations. The structure is made from a T-bar of complexity $q_T = n$ where at the last stage horizontal compressive members are replaced with a D-bar structure of complexity $q_D = 1$ to yield a $T_nD_1$ tensegrity structure. Figure~\ref{TD_2D} shows the 2-dimensional representation of a $T_2D_1$ tensegrity structure with T-bar angle $\alpha_T$ and D-bar angle $\alpha_D$. Notice that because of complexity $q_T = 2$ for the T-bar, there would be two T-bar angles, $q_{T1}$ corresponding to the first iteration and $q_{T2}$ corresponding to the second iteration.
\begin{figure}[ht!]
    \centering
    \includegraphics[width=.85\linewidth]{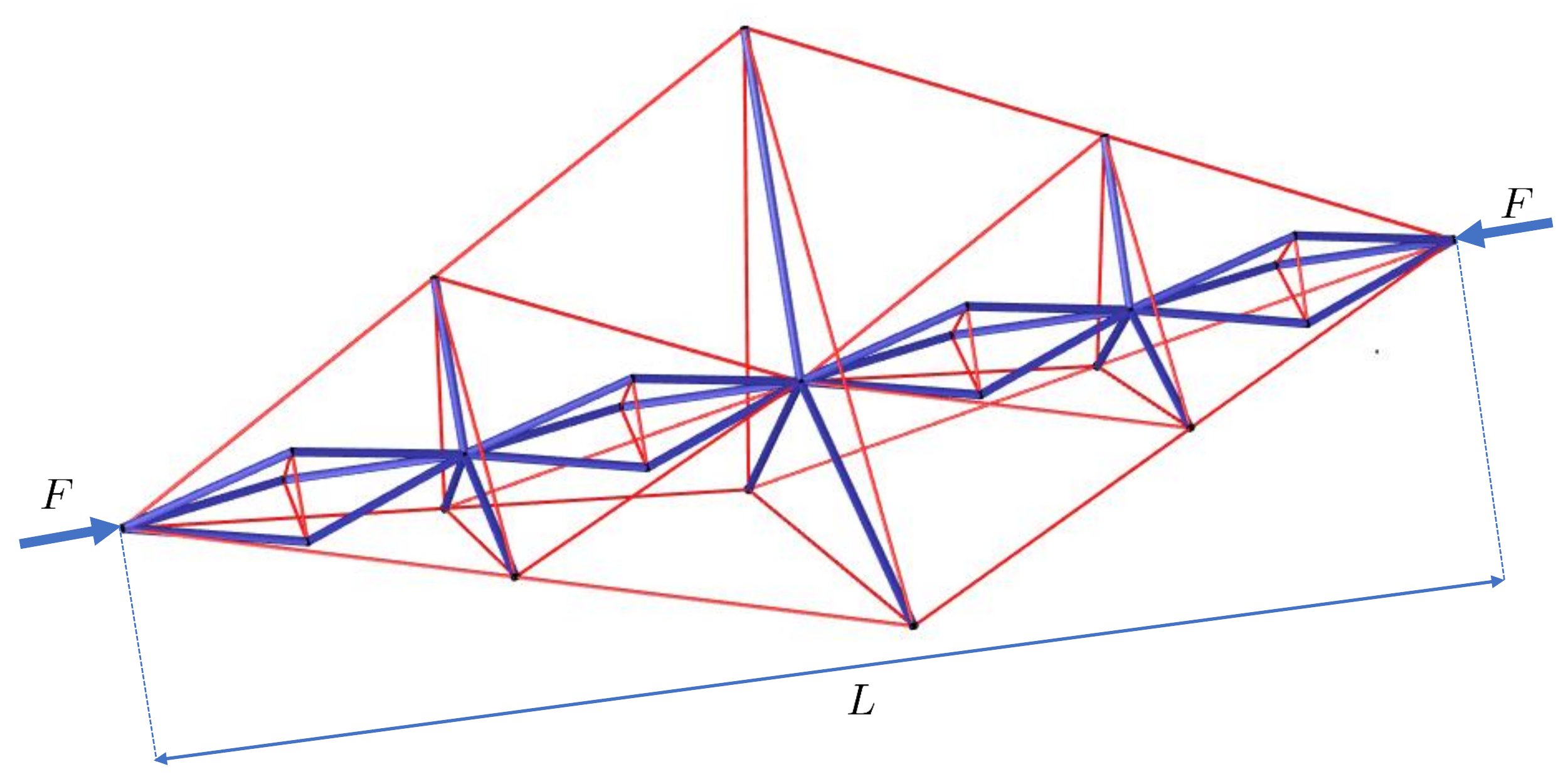}  
    \caption{3D representation of $T_2D_1$ tensegrity structure.}
    \label{TD_3D}
\end{figure}
Figure~\ref{TD_3D} shows the 3-dimensional tensegrity structure. In this section, simulation results are given to show the extension of the robotic arm from a stowed configuration to an extended configuration and its movement (angular motion) to reach any particular point in a given 3-dimensional hemisphere. The motion to rotate the robotic arm about its own axis can not be achieved using the standard tensegrity system. In other words, the last section of the robotic arm which is a D-bar, cannot be rotated about its axis (the line connecting two ends) by simply controlling the rest lengths of the string. However, we can achieve this rotation using gyroscopic tensegrity systems, which is needed in order to reach the desired orientation of the end-effector. Therefore, six gyroscopic wheels are attached to each of the bars in the last D-bar unit to allow for the rotation of the end-effector about its own axis (see Fig.~\ref{f:Error_stow_Final}).
All the dynamic simulations are performed using a {\tt MATLAB}\textregistered~based software {\tt MOTES} developed using the dynamic formulation described earlier [\cite{Goyal_MOTES_2019}].

\subsection{Extension from a stowed configuration}
\begin{figure}[ht!]
\centering
\includegraphics[width=1\linewidth]{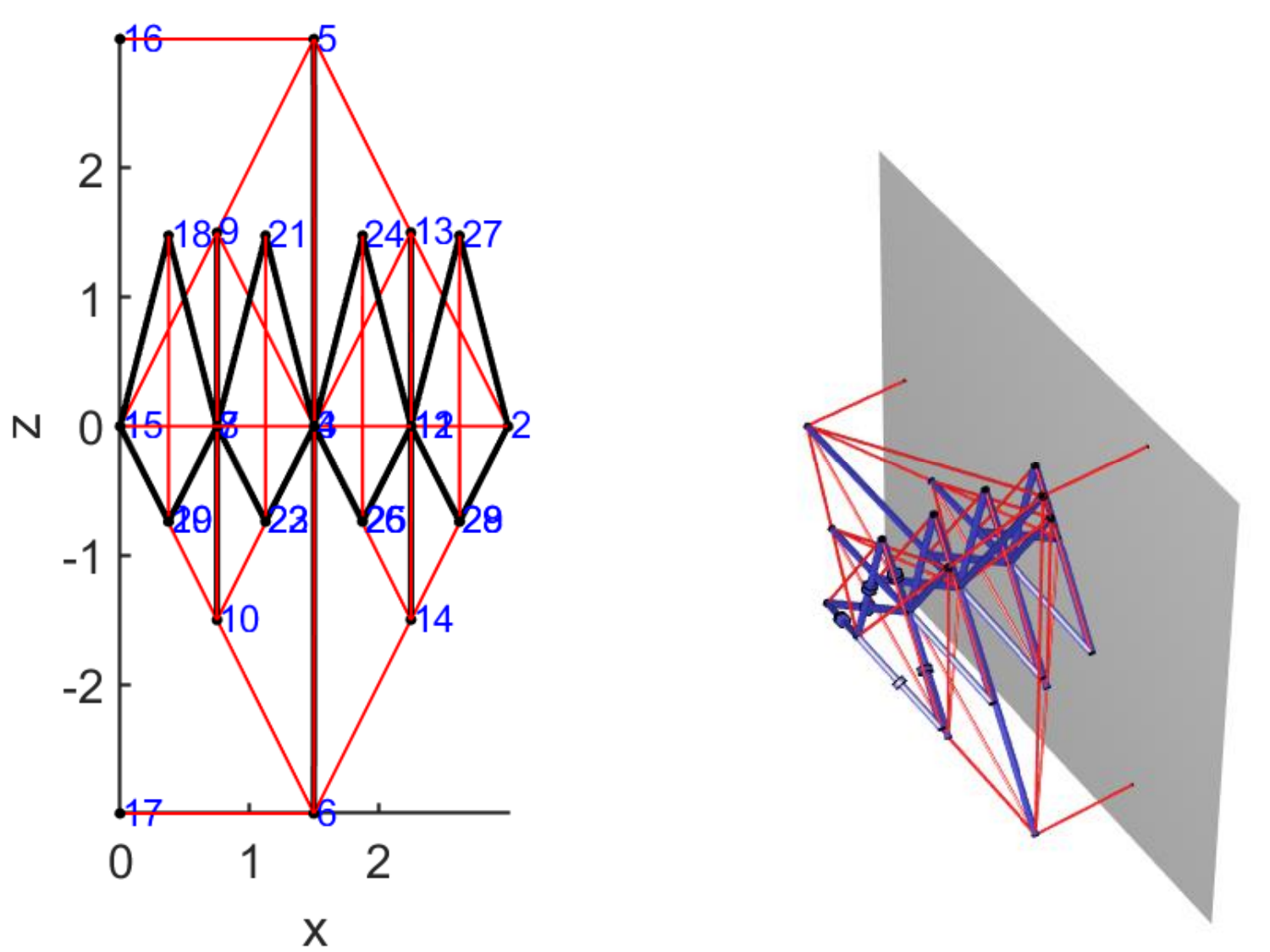}
\caption{Initial configuration of gyroscopic tensegrity $T_2D_1$ robotic arm. The numbers in the blue represent the node numbers.}
\label{f:Error_stow_Initial}
\end{figure}

\begin{figure}[ht!]
\centering
\includegraphics[width=1\linewidth]{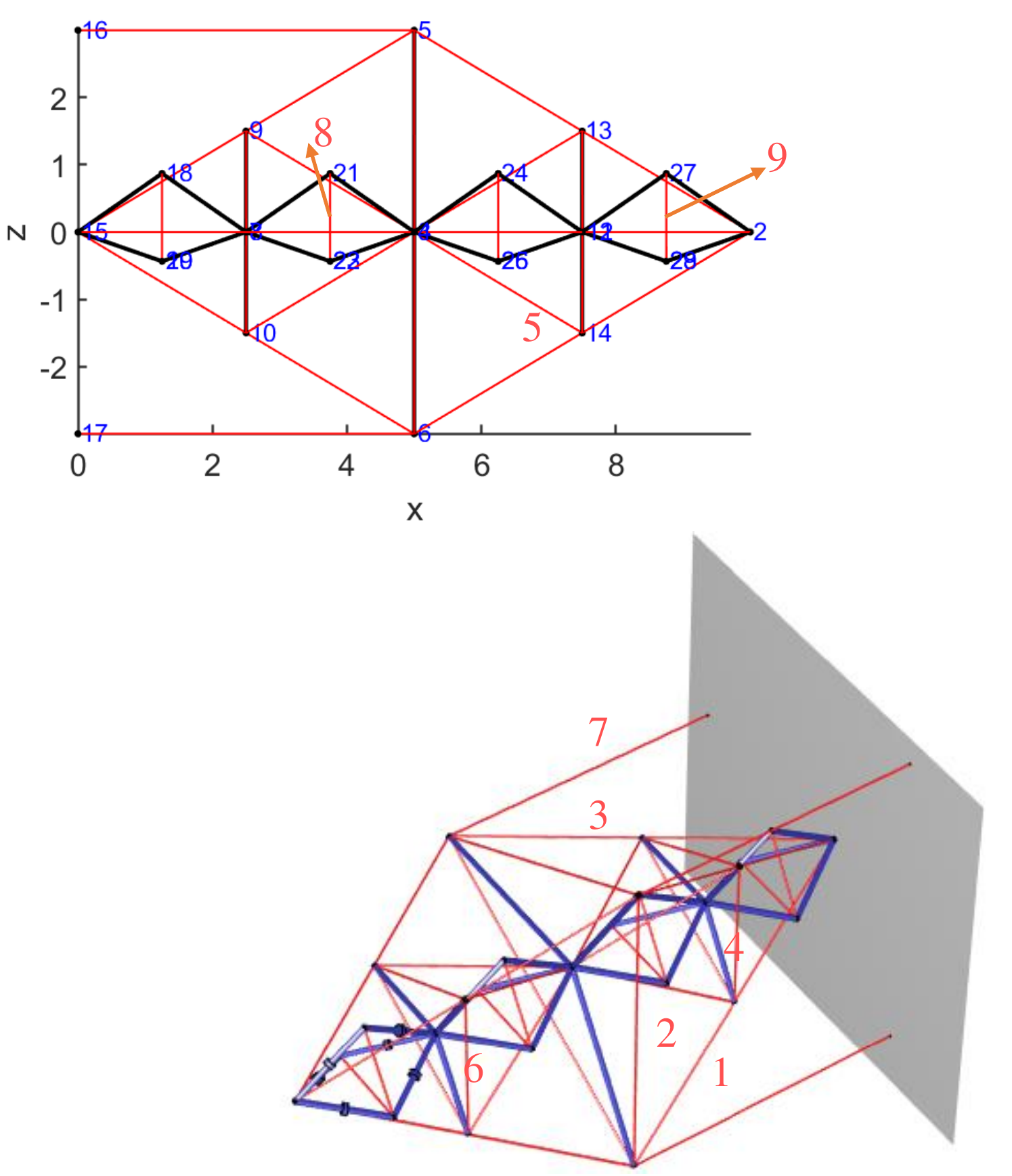}
\caption{Final configuration of gyroscopic tensegrity $T_2D_1$ robotic arm. The numbers in the red represent the string numbers.}
\label{f:Error_stow_Final}
\end{figure}

In this subsection, we extend the $T_2D_1$ tensegrity robotic arm from a stowed configuration to an extended configuration. This can be understood as controlling the shape of the robotic arm from some initial configuration shown in Fig.~\ref{f:Error_stow_Initial}, to a final configuration shown in Fig.~\ref{f:Error_stow_Final}.
%
%
%
%
%
\begin{figure*}[h!]
\begin{multicols}{3}
    \centering
      \subfloat[Error in node-position]{\includegraphics[width=1\linewidth]{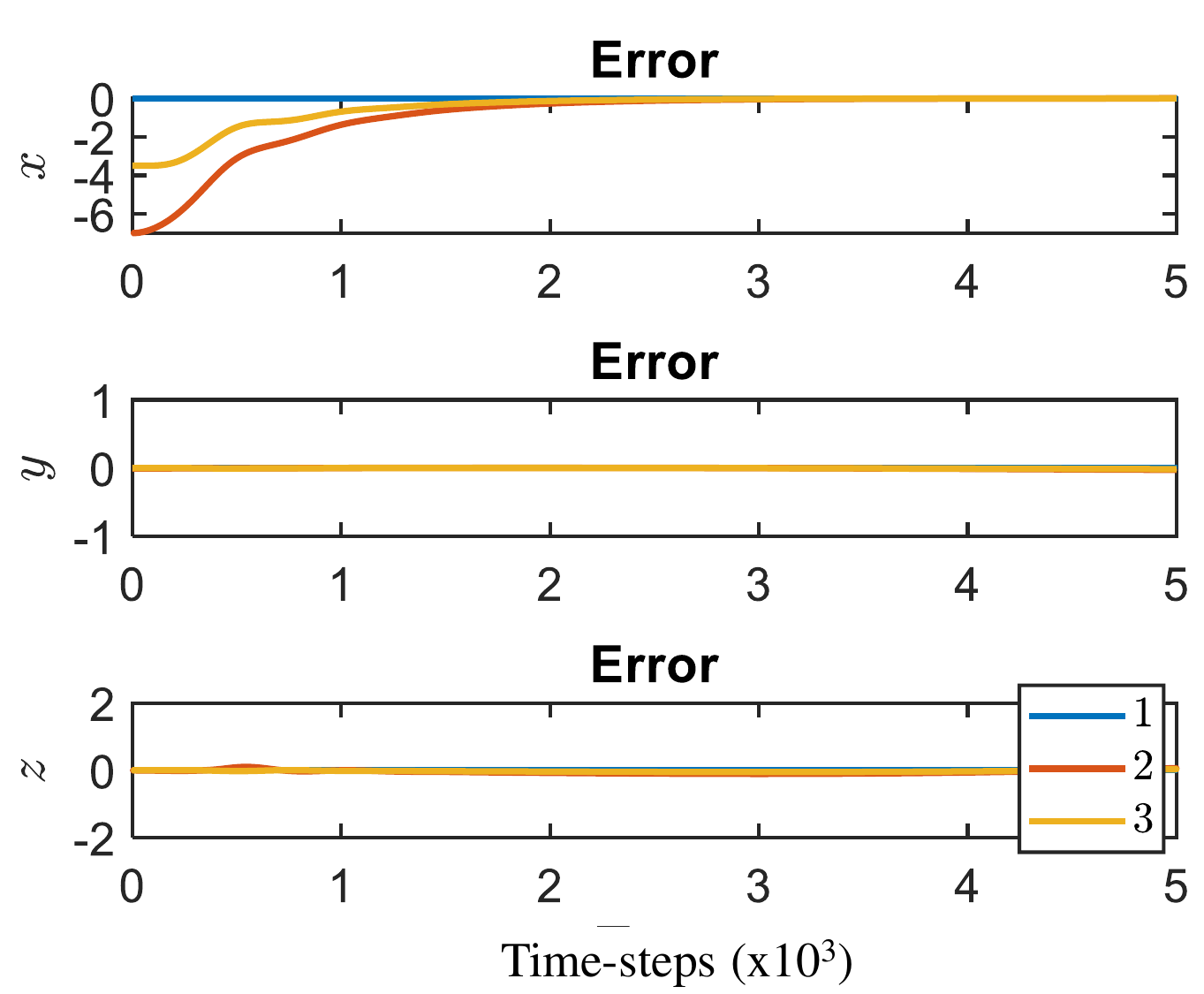}} 
      \subfloat[Position trajectories]{\includegraphics[width=1\linewidth]{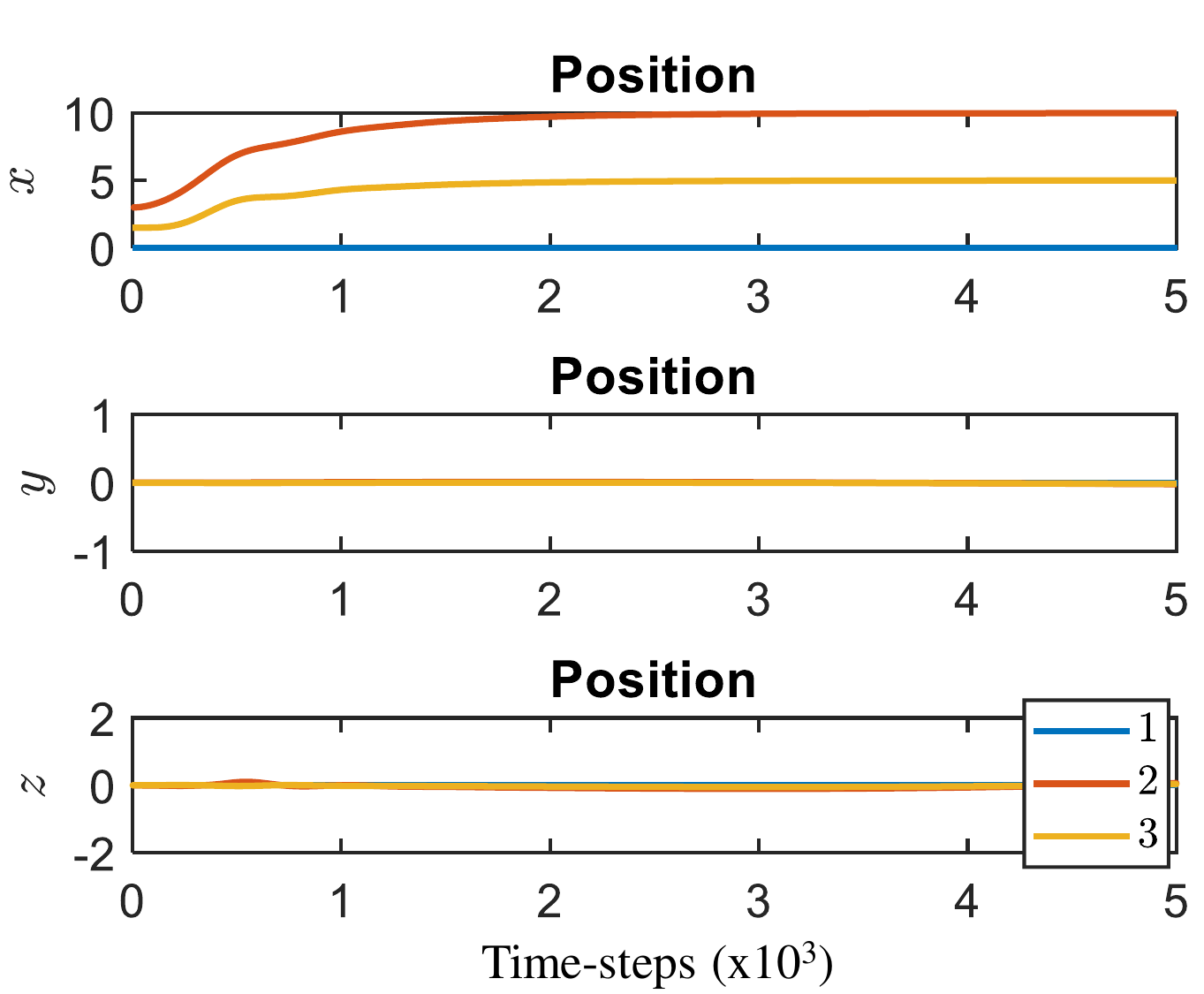}}
      \subfloat[Velocity trajectories]{\includegraphics[width=1\linewidth]{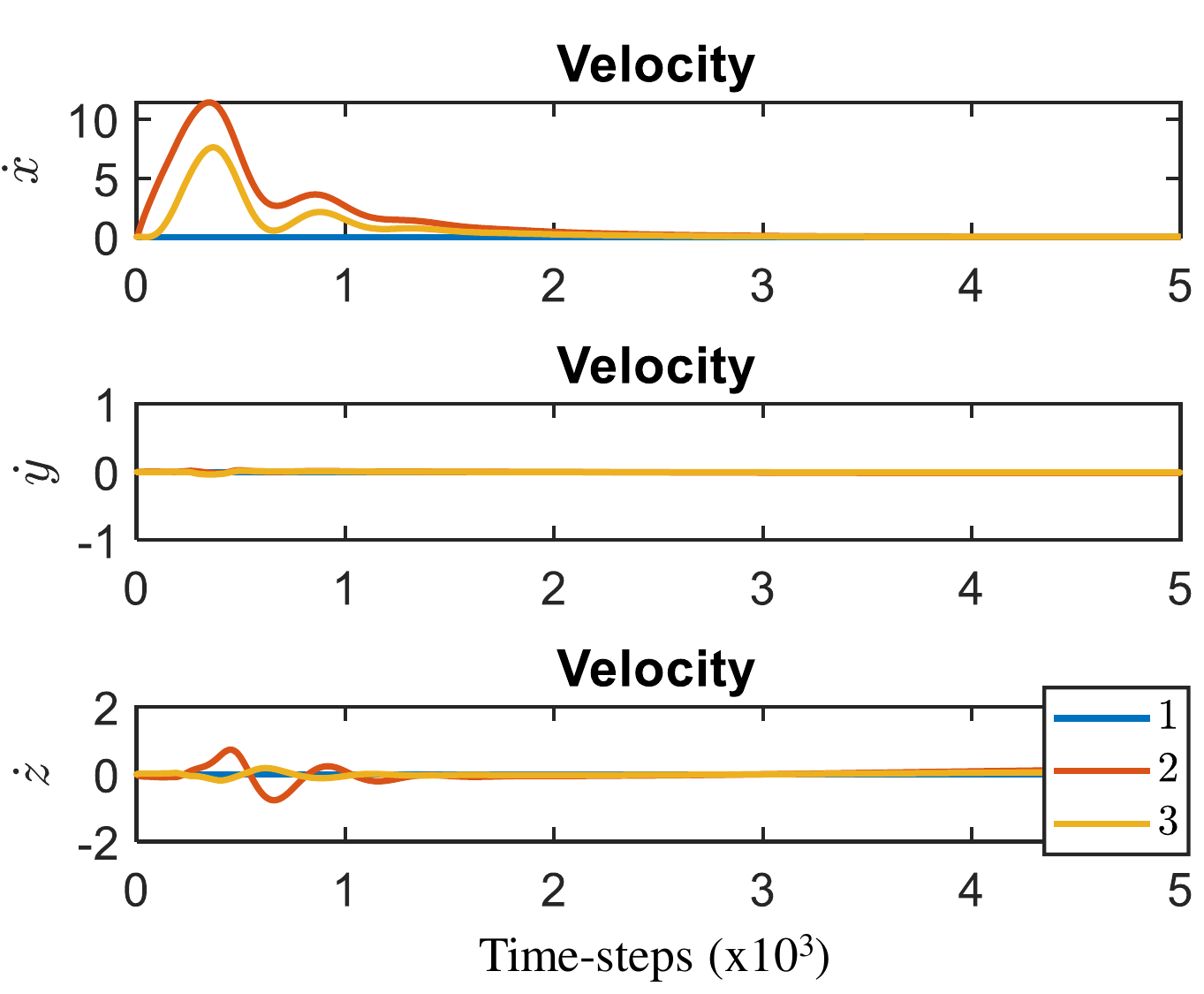}}
\end{multicols}
\caption{Plots for the extension of tensegrity $T_2D_1$ robotic arm.}
\label{f:Error_stow_3}
\end{figure*}

We use the control algorithm developed in the previous section to control the configuration of the robotic arm, which drives the errors to zero by carefully designing a trajectory based on the control gains. The angular speed of all the six wheels in this simulation was kept to be zero. Figure~\ref{f:Error_stow_3}(a) shows the trajectory for the errors in the node position for nodes $n_1, n_2$ and $n_3$. Notice that the error goes to zero and stays there after roughly $2 \times 10^3$ time-steps. Notice the small error in $y$ and $z$ directions as the arm extends along its length which lies along the $x$ axis as shown in Fig.~\ref{f:Error_stow_Final}. Figure~\ref{f:Error_stow_3}(b) shows the node position trajectories for nodes $n_1, n_2$ and $n_3$ during the extension of the tensegrity arm. Notice that as the arm extends, the $x$ coordinates for the nodes increases and settle for the desired value, and it was also observed that as the length of the bars are constant, the $z$ coordinates of the nodes decrease and finally reaches the steady-state position (observed in the trajectories of different nodes which are not given here). This can also be understood as the arm extension in length causes the thickness of the arm to decrease. The velocities of those nodes are shown in Fig.~\ref{f:Error_stow_3}(c).
\begin{figure}[ht!]
\centering
\includegraphics[width=.75\linewidth]{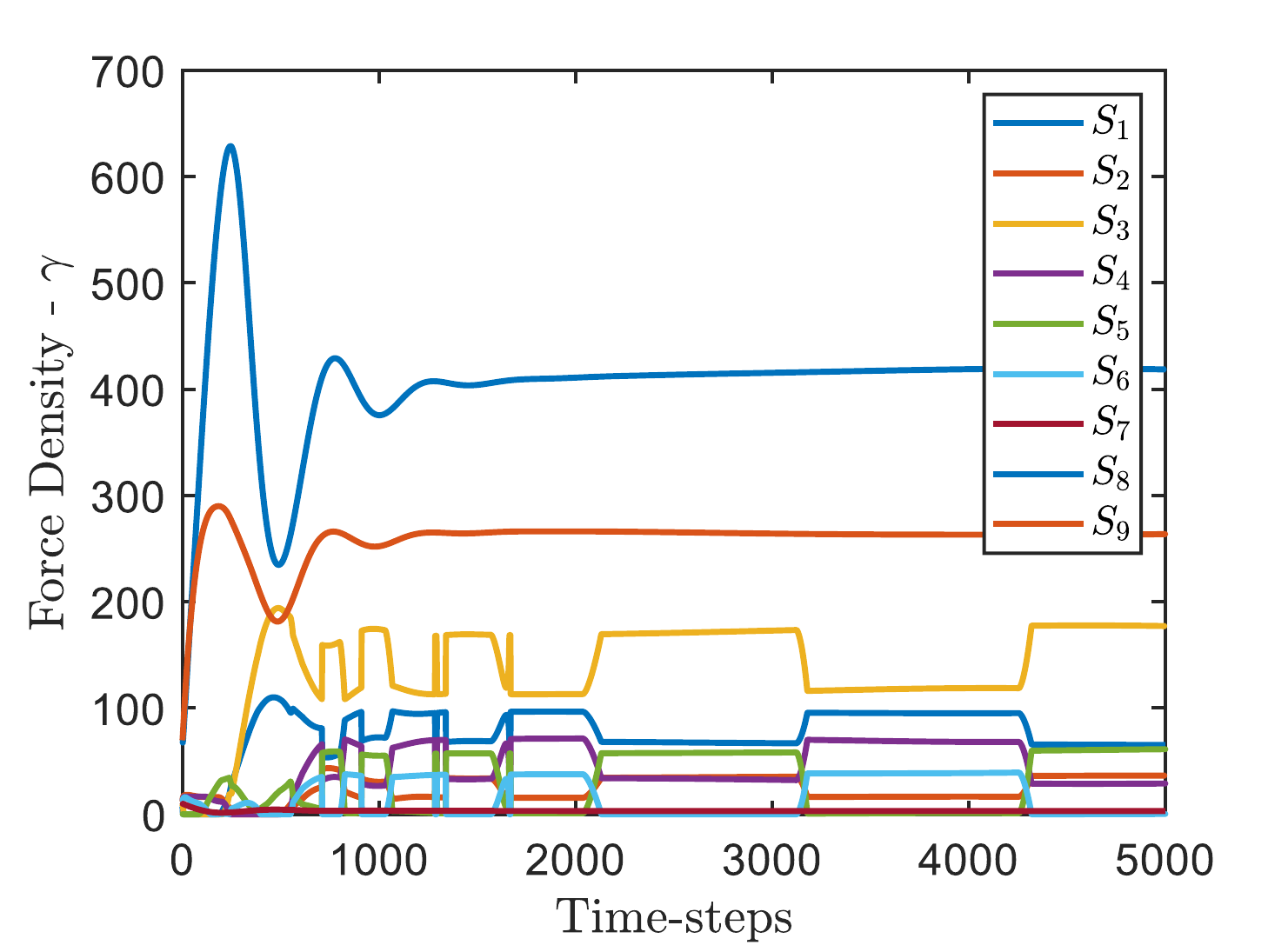}
\caption{Trajectories for force-densities in the strings for the extension of tensegrity $T_2D_1$ robotic arm.}
\label{f:Control_Stow}
\end{figure}

Finally, Fig.~\ref{f:Control_Stow} shows the control inputs required to extend the tensegrity $T_2D_1$ robotic arm. The control inputs in these systems are assumed to be the force density in the strings, which can be uniquely converted to the rest length (real physical control) of the strings. The figure shows trajectories only for nine strings in the structure as due to the symmetry of the structure, other strings will follow the same trend and can easily be recognized. Notice that the control values reach a steady-state value after 2 seconds, but small changes in the strings can still be expected as tensegrity structures have multiple equilibrium solutions for a given configuration of the structure. The step changes in the control values are also due to multiple solutions of force densities switching from one set of strings to the other set of strings.

\subsection{Tip movement in 3-dimensional hemisphere}
The previous subsection extended the arm to the desired location, and in this subsection, we control the shape of the structure (angular motion) for the tip of the arm to reach any particular point in the given 3-dimensional space. The reduced-order dynamic model was used in controlling the shape of the robotic arm. 
\begin{figure}[ht!]
\centering
\includegraphics[width=1\linewidth]{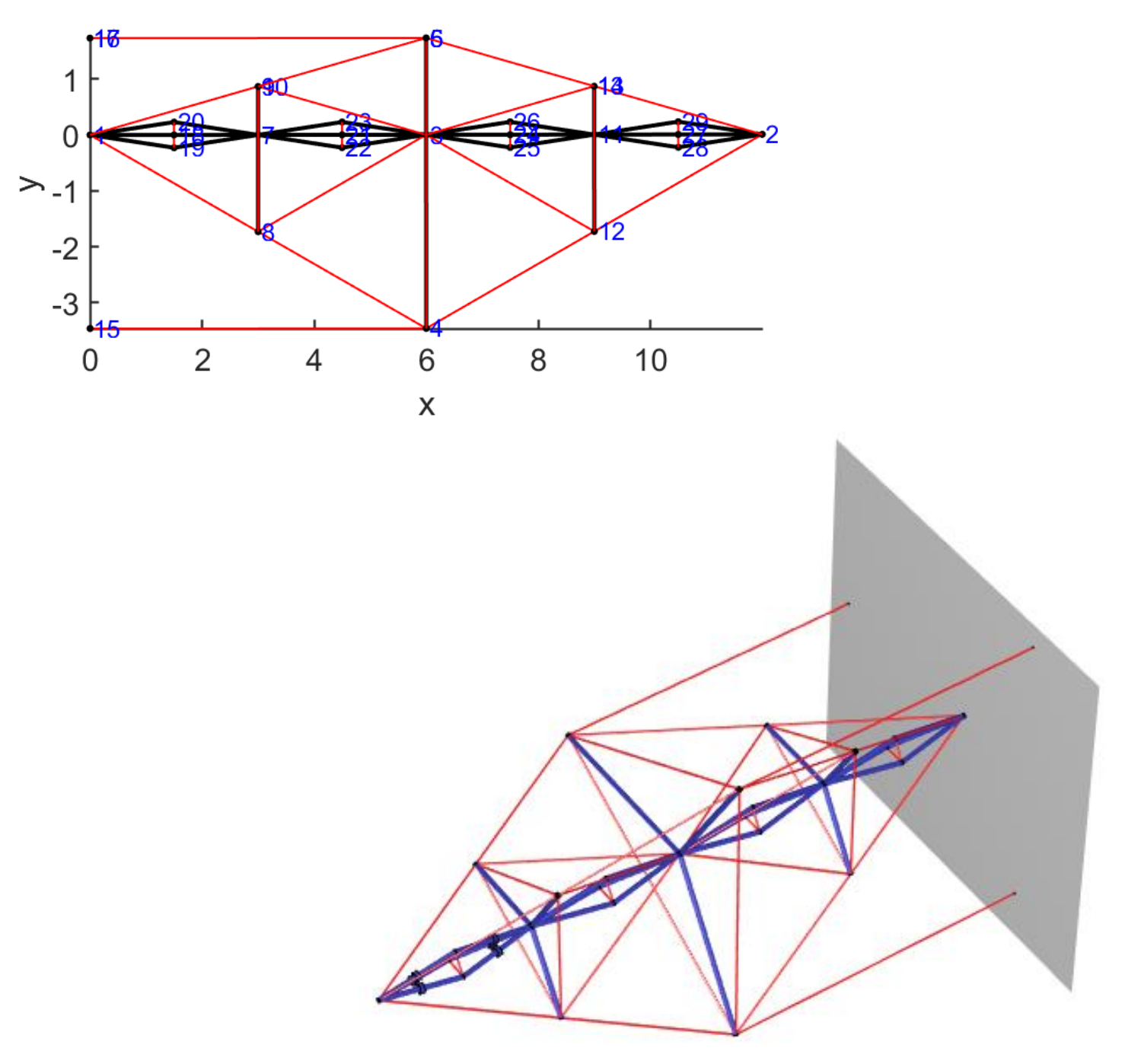}
\caption{Initial configuration of gyroscopic tensegrity $T_2D_1$ robotic arm for angular rotation.}
\label{f:Error_Angle_Initial}
\end{figure}
%
%
%
%
\begin{figure*}[ht!]
\begin{multicols}{3}
    \centering
      \subfloat[Error in node-position]{\includegraphics[width=1\linewidth]{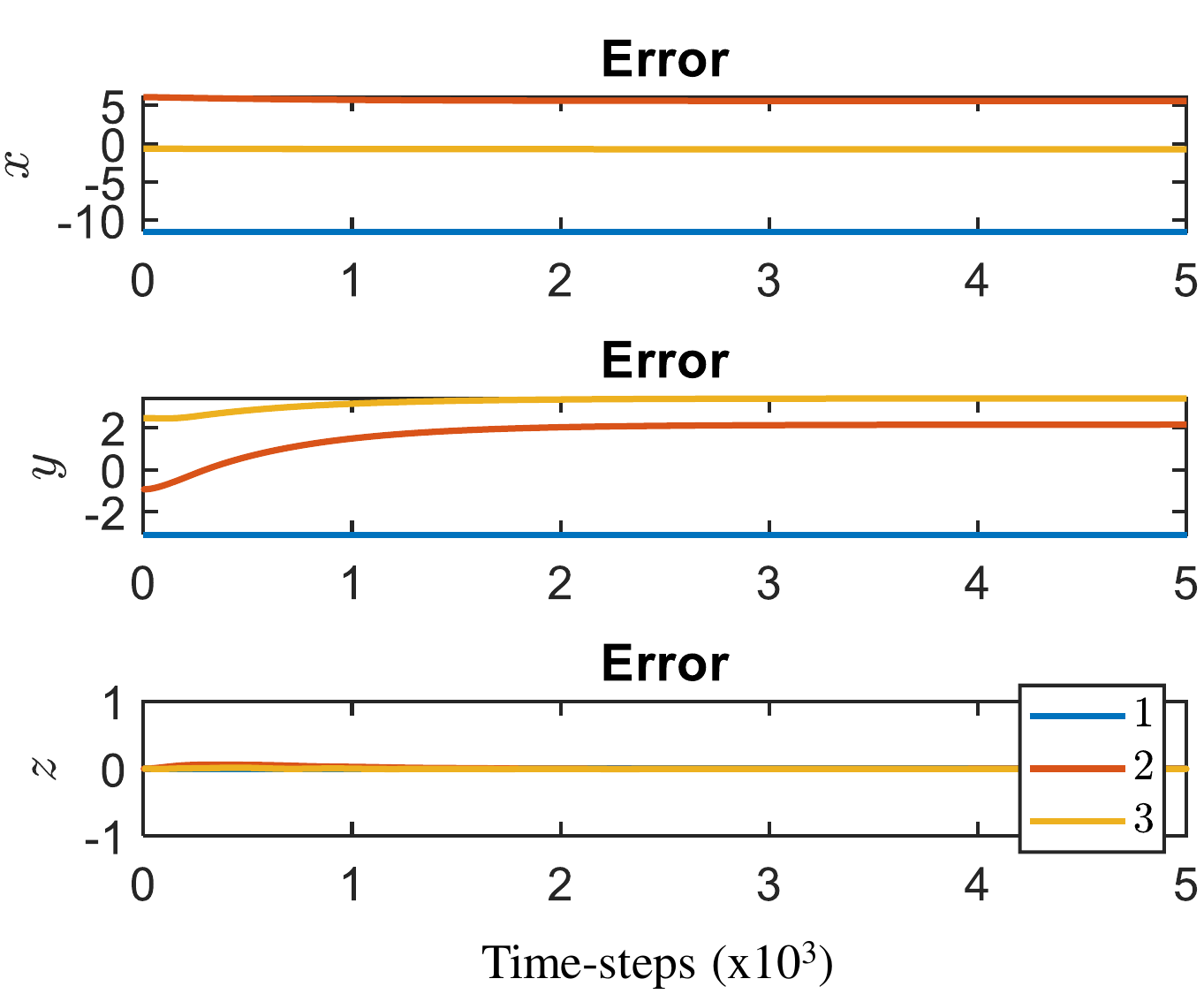}} 
      \subfloat[Position trajectories]{\includegraphics[width=1\linewidth]{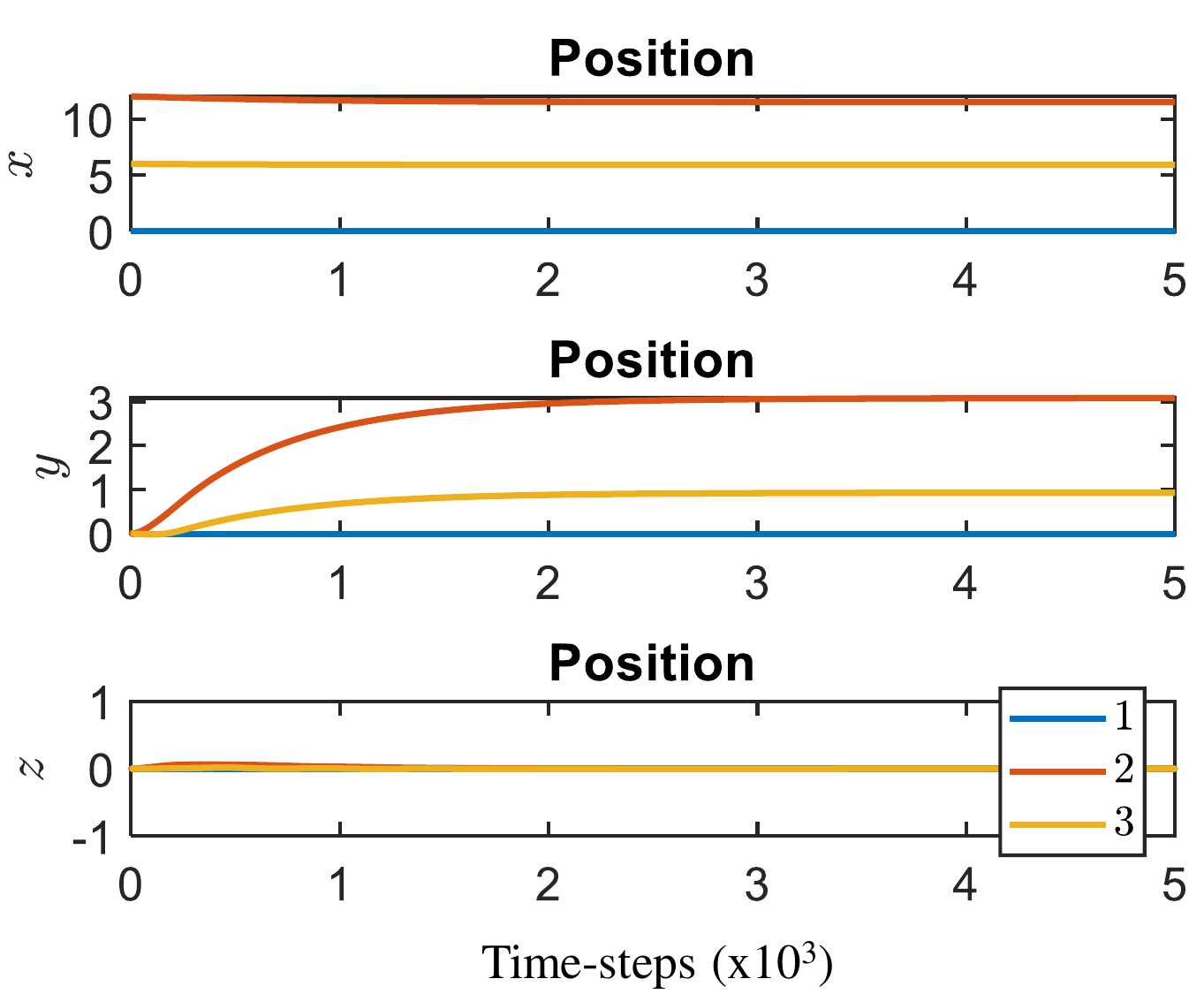}}
      \subfloat[Velocity trajectories]{\includegraphics[width=1\linewidth]{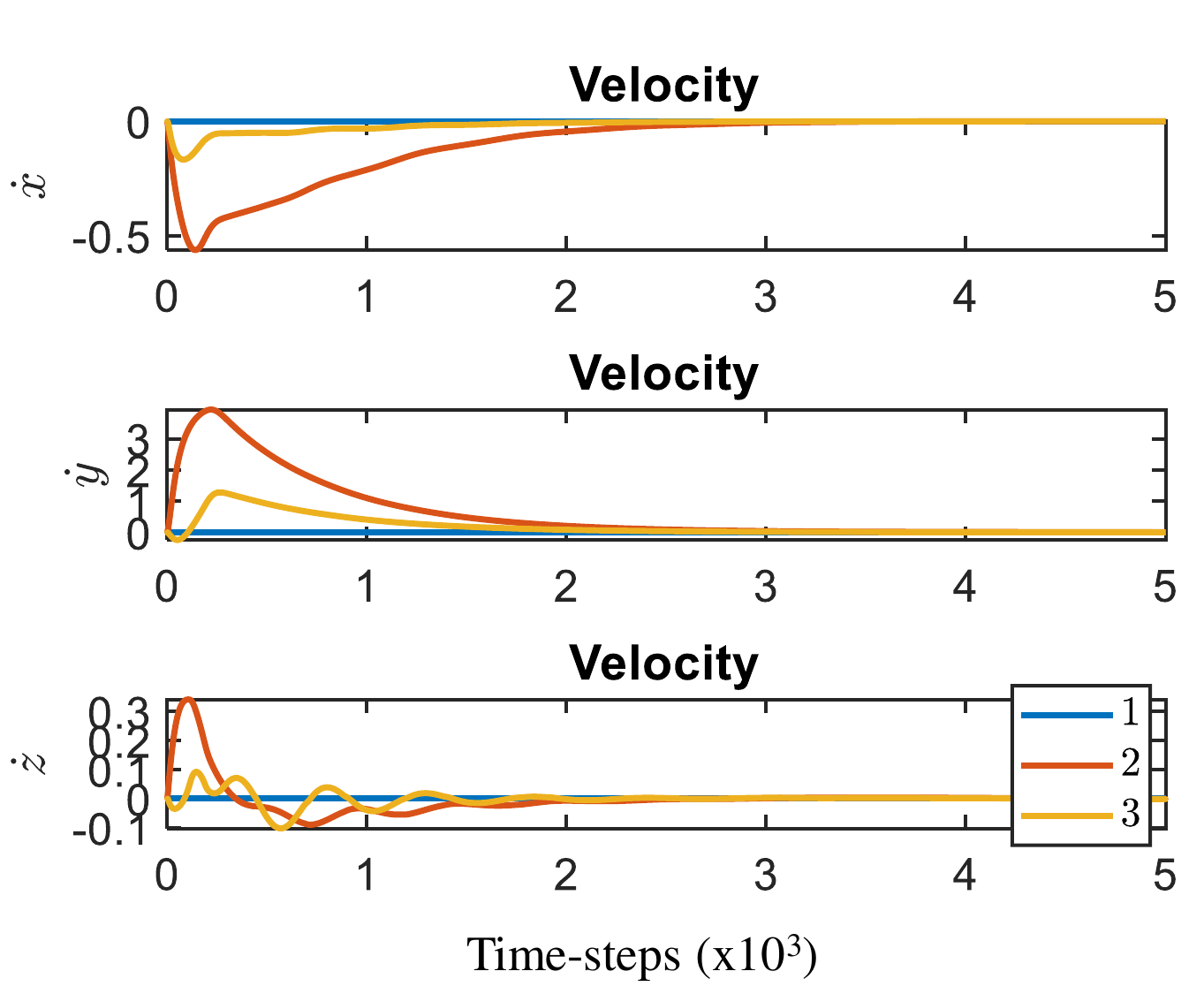}}
\end{multicols}
\caption{Plots for the angular rotation of gyroscopic tensegrity $T_2D_1$ robotic arm.}
\label{f:Error_Angle_2}
\end{figure*}
\begin{figure}[h!]
\centering
\includegraphics[width=1\linewidth]{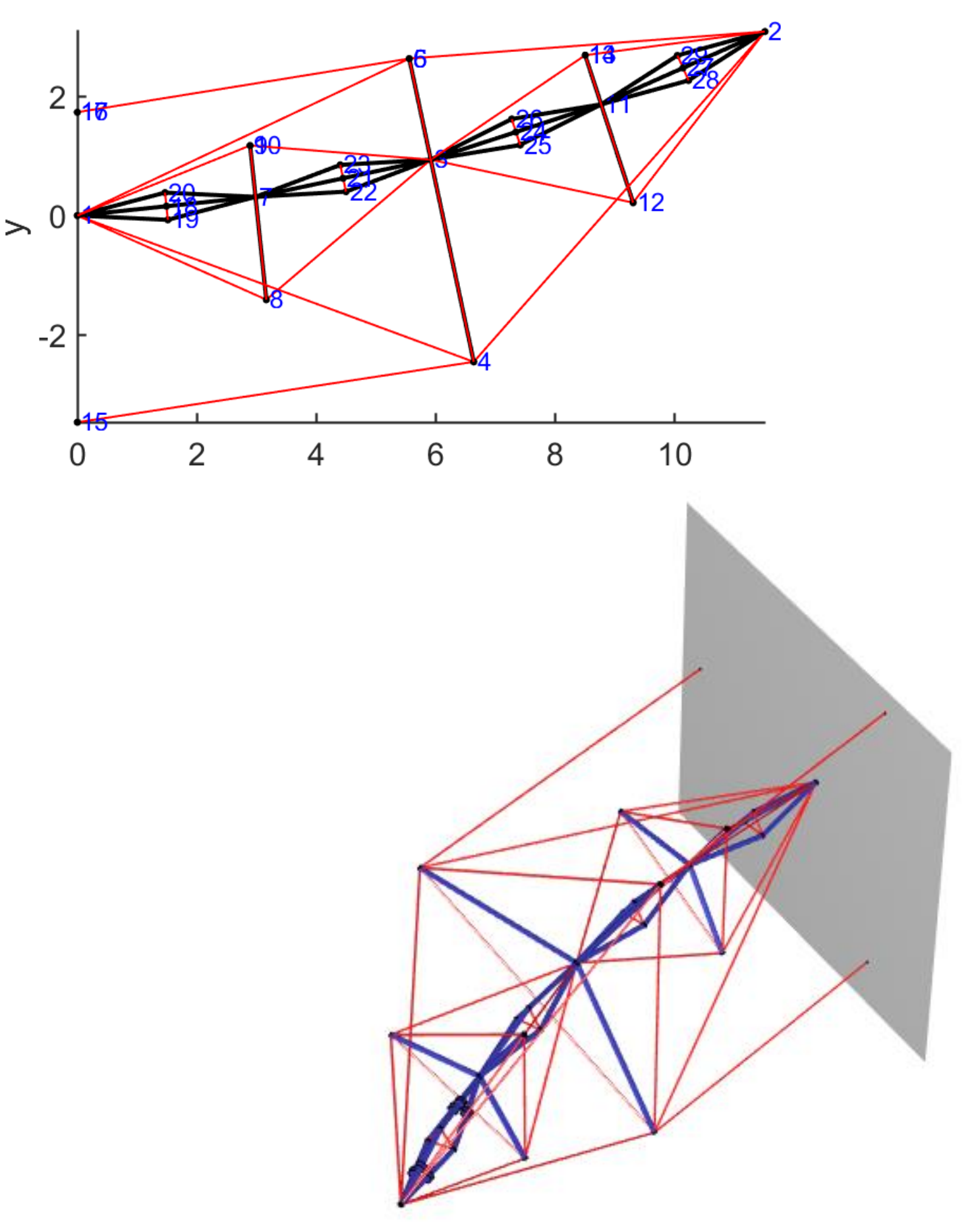}
\caption{Final configuration of gyroscopic tensegrity $T_2D_1$ robotic arm for angular rotation.}
\label{f:Error_Angle_Final}
\end{figure}
Figure~\ref{f:Error_Angle_Initial} shows the initial configuration of the robot with the tip of the arm pointing along the length of the arm. In this example, we aim to move the tip to the arm to the desired location corresponding to the configuration shown in Fig.~\ref{f:Error_Angle_Final}.

\begin{figure}[ht!]
\centering
\includegraphics[width=.75\linewidth]{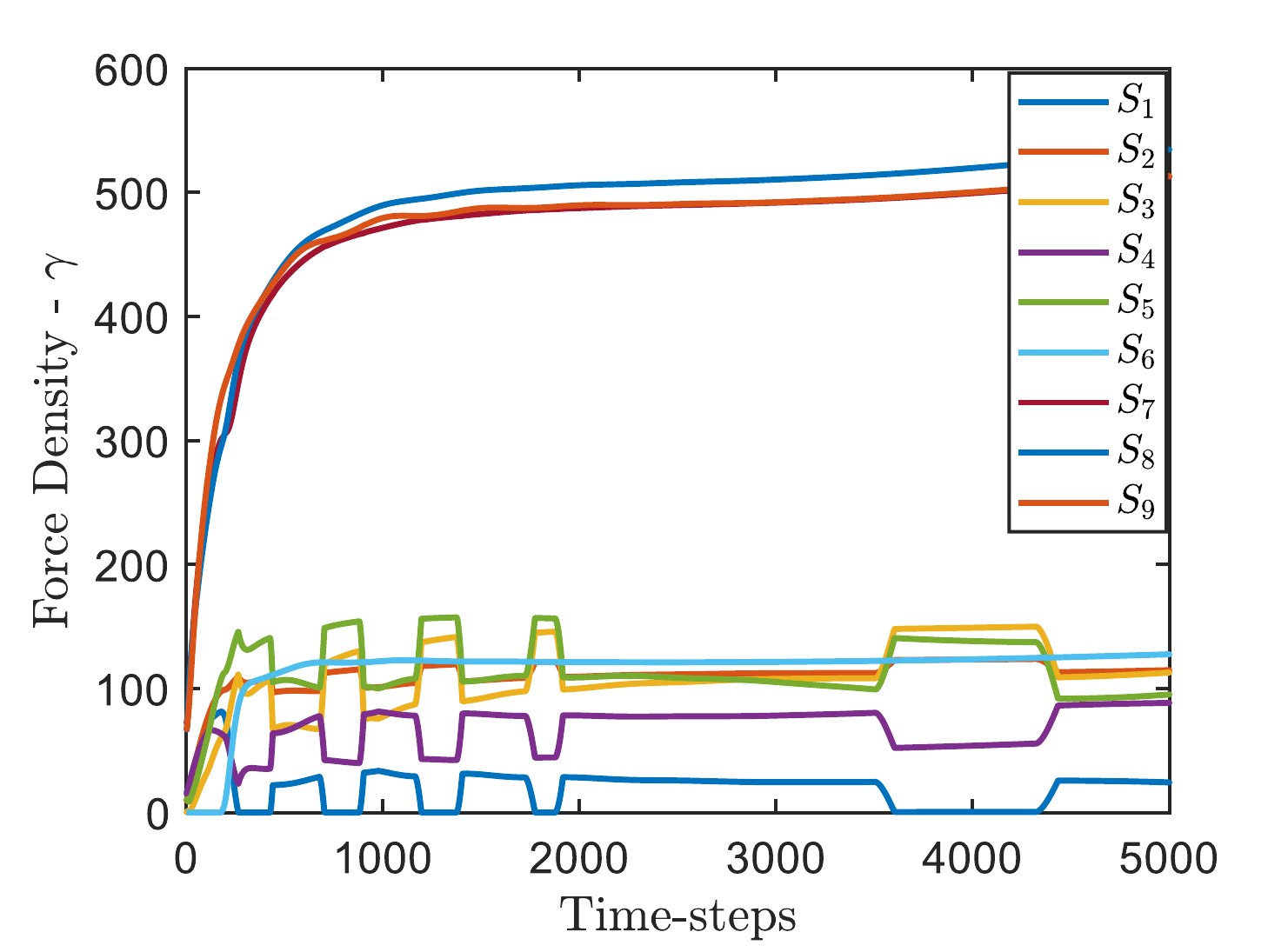}
\caption{Trajectories for force-densities in the strings for the angular rotation of the robotic arm.}
\label{f:Control_Angle}
\end{figure}

It is relatively difficult to reach the desired position for the tip of the arm without specifying the position of the other nodes in the structure. Therefore, to perform this simulation, the node positions of the final configuration was first calculated based on the kinematic analysis.
Figure~\ref{f:Error_Angle_2}(a) shows the plots for the nodes $n_1, n_2$ and $n_3$. Notice that for this simulation, both the $x$ coordinate and $y$ coordinate change (refer Fig.~\ref{f:Error_Angle_Final}) and reach a steady state value in again $\approx 2 \times 10^3$ time-steps. Small disturbances are observed in $z$ direction also as the arm moves to the desired position. The node position trajectories for nodes $n_1, n_2$ and $n_3$ during the movement of the tip of the robotic arm to reach a given desired position in the 3-dimensional space are shown in Fig.~\ref{f:Error_Angle_2}(b) and the corresponding velocities are shown in Fig.~\ref{f:Error_Angle_2}(c). 
The wheels were assumed to be static for this motion of the robotic arm. 
Finally, Fig.~\ref{f:Control_Angle} shows the control inputs (force density in the strings) required to move the tip of the tensegrity robotic $T_2D_1$ arm. 
The figure shows the trajectories for some of the strings in the process as the rest of the strings follow the same trend due to the symmetry of the structure.

\subsection{End-effector rotation about the axis of the arm}
The gyroscopic wheels in this robotic arm are needed only for the last D-bar section of the arm to achieve the desired rotation of the end-effector about its own axis.
\begin{figure*}[h!]
\begin{multicols}{3}
    \centering
      \subfloat[Error in node-position]{\includegraphics[width=1.05\linewidth]{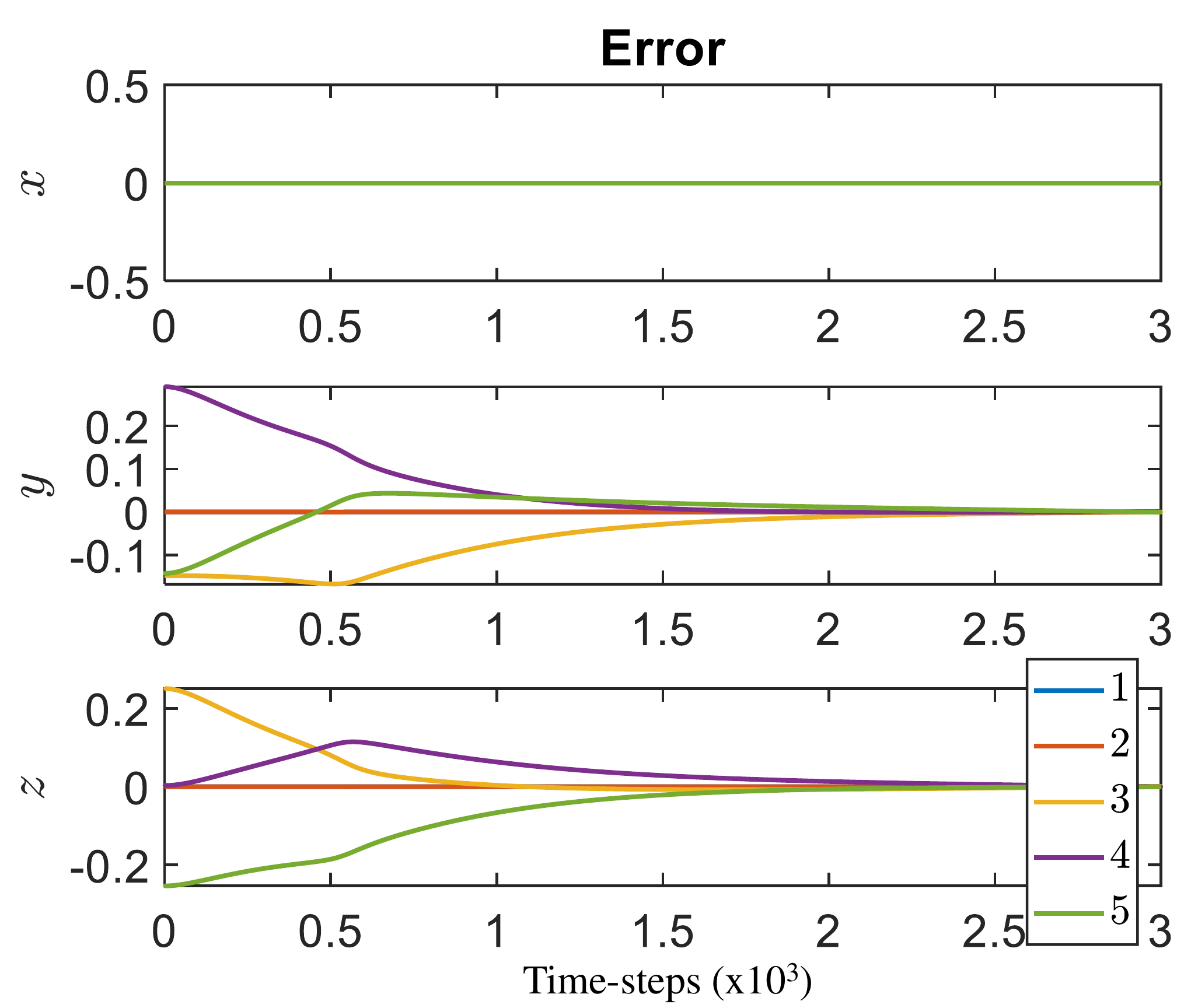}} 
      \subfloat[Position trajectories]{\includegraphics[width=1\linewidth]{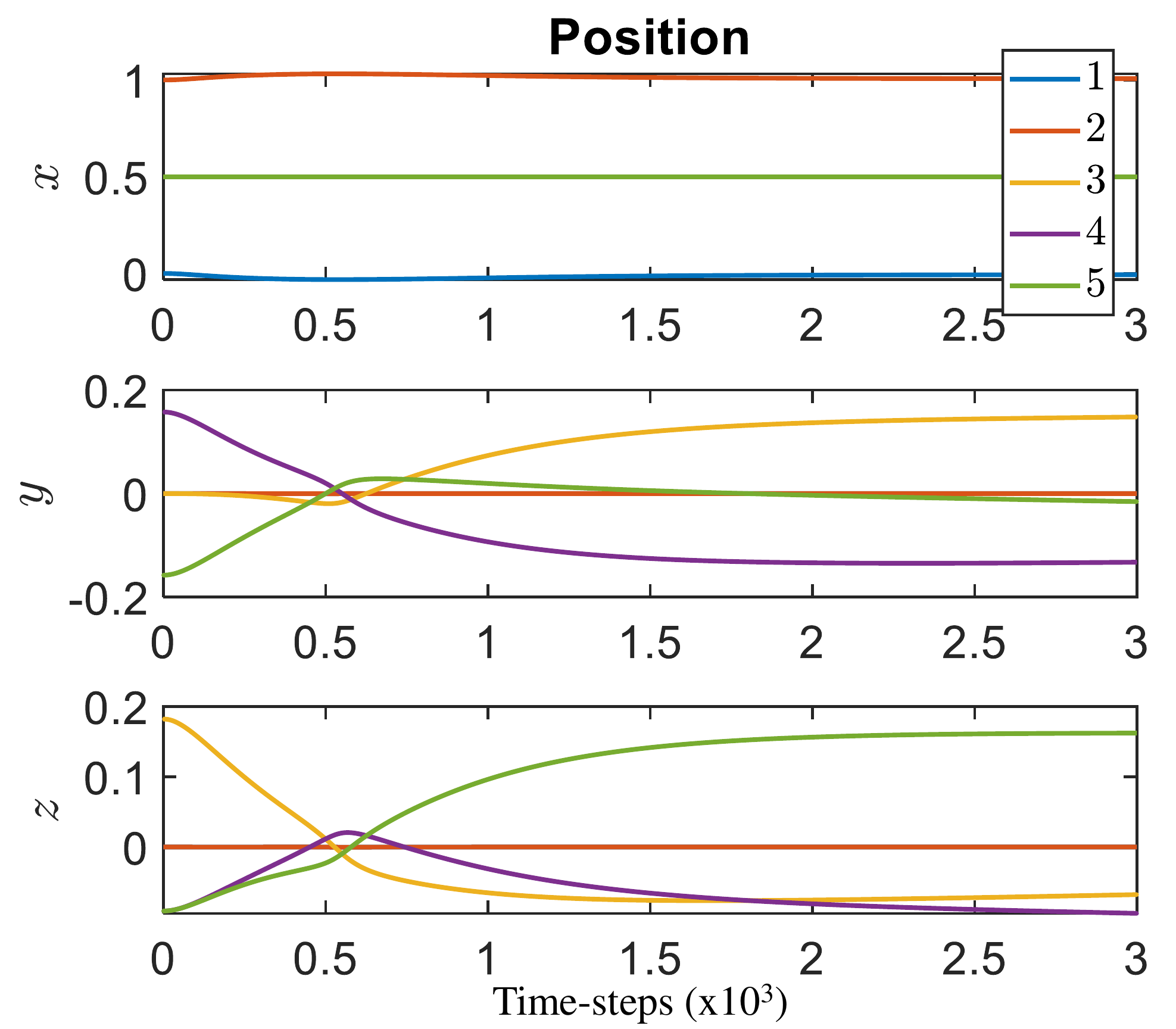}}
      \subfloat[Velocity trajectories]{\includegraphics[width=1\linewidth]{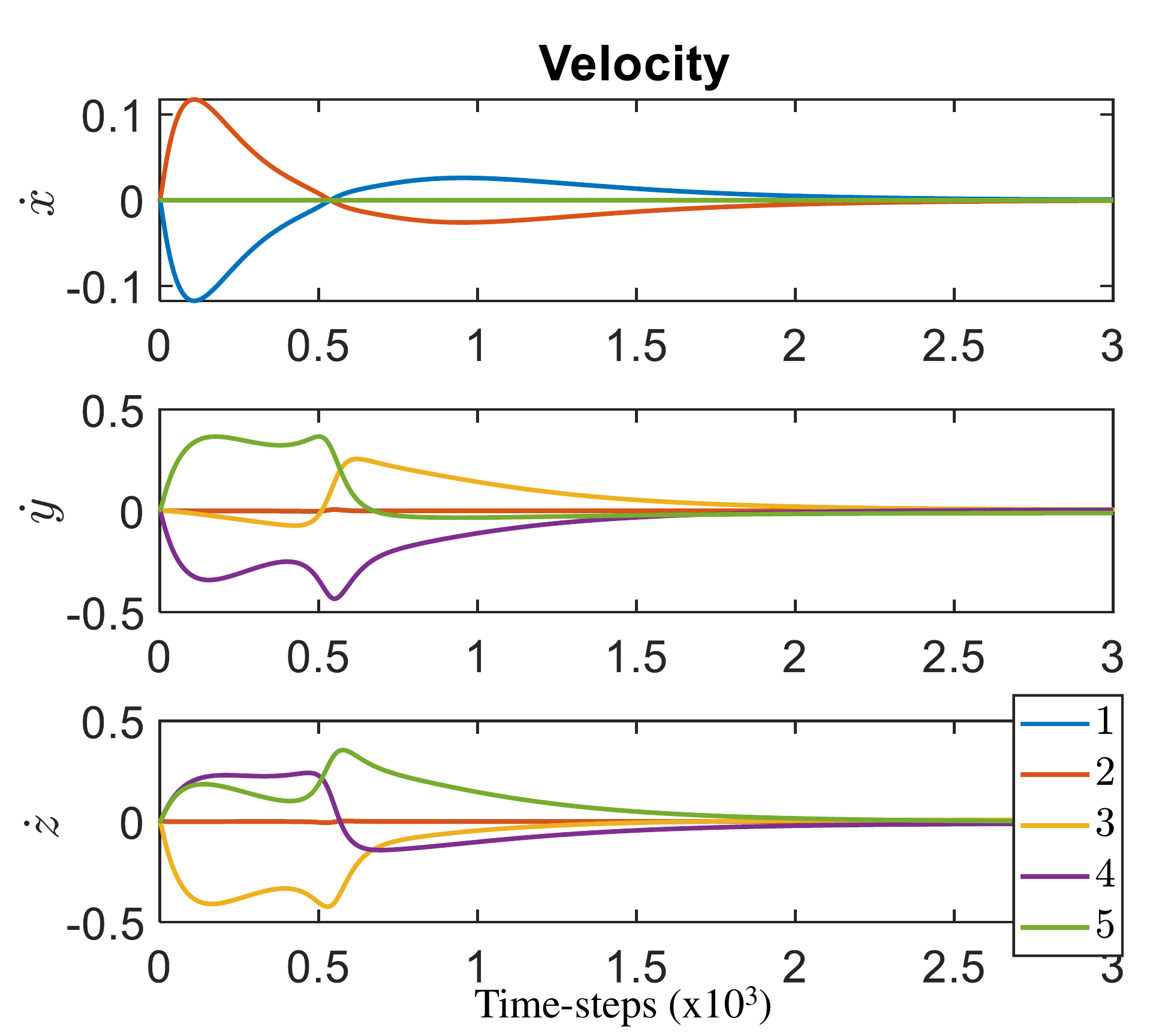}}
\end{multicols}
\caption{Plots for the end-effector rotation of gyroscopic tensegrity robotic arm.}
\label{f:3DDbar_Gyro}
\end{figure*}
\begin{figure}[ht!]
\centering
\includegraphics[width=1\linewidth]{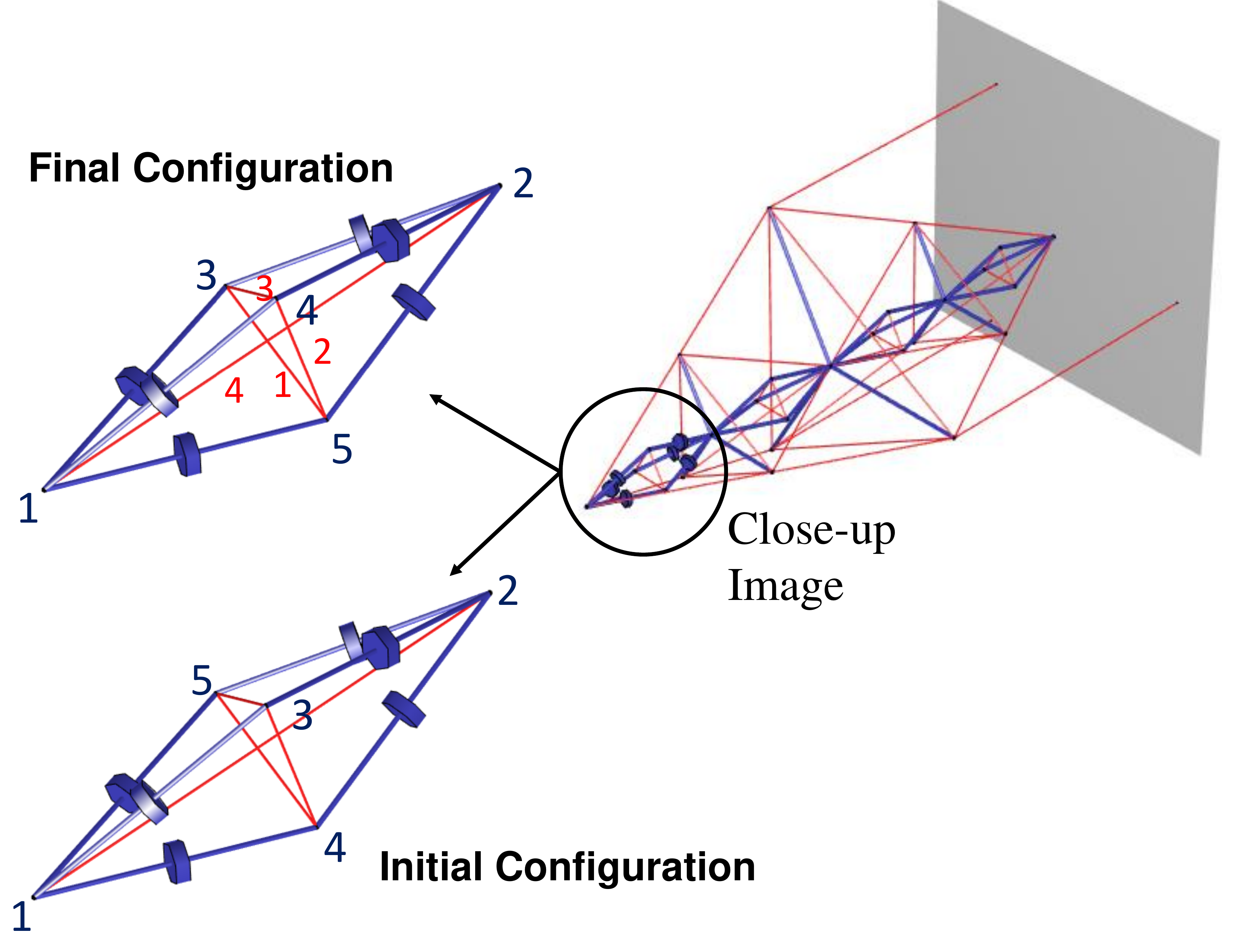}
\caption{Initial and final configuration of the gyroscopic tensegrity $T_2D_1$ robotic arm for end-effector rotation.}
\label{f:Config_3DDbar_Gyro}
\end{figure}
The objective here is to rotate the last part of the robotic arm (the D-bar structure) from the initial configuration to the final configuration, as shown in Fig.~\ref{f:Config_3DDbar_Gyro}. Notice the nodes $n_3, n_4$, and $n_5$ in both configurations to see the different orientation of the D-bar section of the robotic arm.

Figure~\ref{f:3DDbar_Gyro}(a) shows the plots for error in node-position in the orientation control of the gyroscopic tensegrity $T_2D_1$ robotic arm. The error in the figure reaches a steady-state value of zero in around $2.5 \times 10^3$ time-steps for all the three axes. Notice that the $x$ axis is shown to have no initial and final error, it is because the matrix $L$ in the formulation was chosen to control only $y$ and $z$ axis:
\begin{align}
    L = \begin{bmatrix} 0 & 1 & 0\\0 & 0 & 1 \end{bmatrix}.
\end{align}


Figure~\ref{f:3DDbar_Gyro}(b) shows the node position trajectories in controlling the orientation of the end-effector of the gyroscopic tensegrity robotic arm. The plots corresponding to $y$ and $z$ axis show the movement for the rotation as the values change from the initial configuration to the final configuration. The node velocity trajectories in the orientation control of the end-effector are shown in Fig.~\ref{f:3DDbar_Gyro}(c). The velocity for all the nodes reaches the steady-state value of zero depicting that the structure has reached and will maintain the desired final configuration.

\begin{figure}[h!]
\centering
\includegraphics[width=.85\linewidth]{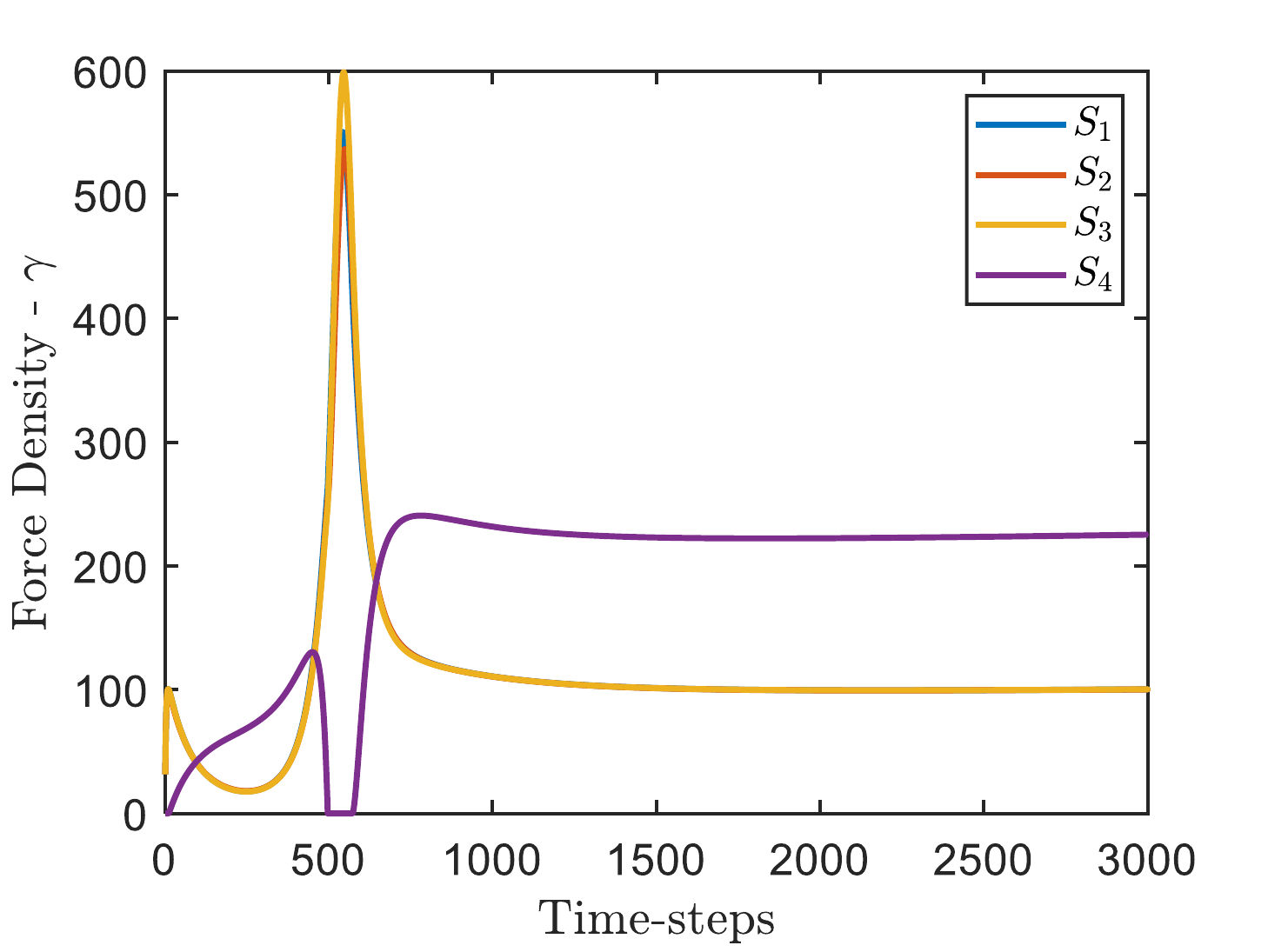}
\caption{Trajectories for force-densities in the strings in the end-effector rotation of gyroscopic tensegrity $T_2D_1$ robotic arm.}
\label{f:Control_3DDbar_Gyro}
\end{figure}
The control inputs (force densities in the string) required to rotate the gyroscopic robotic arm are shown in Fig.~\ref{f:Control_3DDbar_Gyro}. The force densities for the first three strings ($S_1, S_2$, and $S_3$) which are part of the D-bar triangle, follow the same trend due to the symmetry of the structure and reach a steady-state value along with the last string $S_4$. Notice that only these four strings are needed in rotating the end-effector of the tensegrity $T_2D_1$ robotic arm about its axis as the last D-bar is connected to the rest of the arm by a ball joint.

\subsection{Complete motion analysis}
\begin{figure}[h!]
\centering
\includegraphics[width=.8\linewidth]{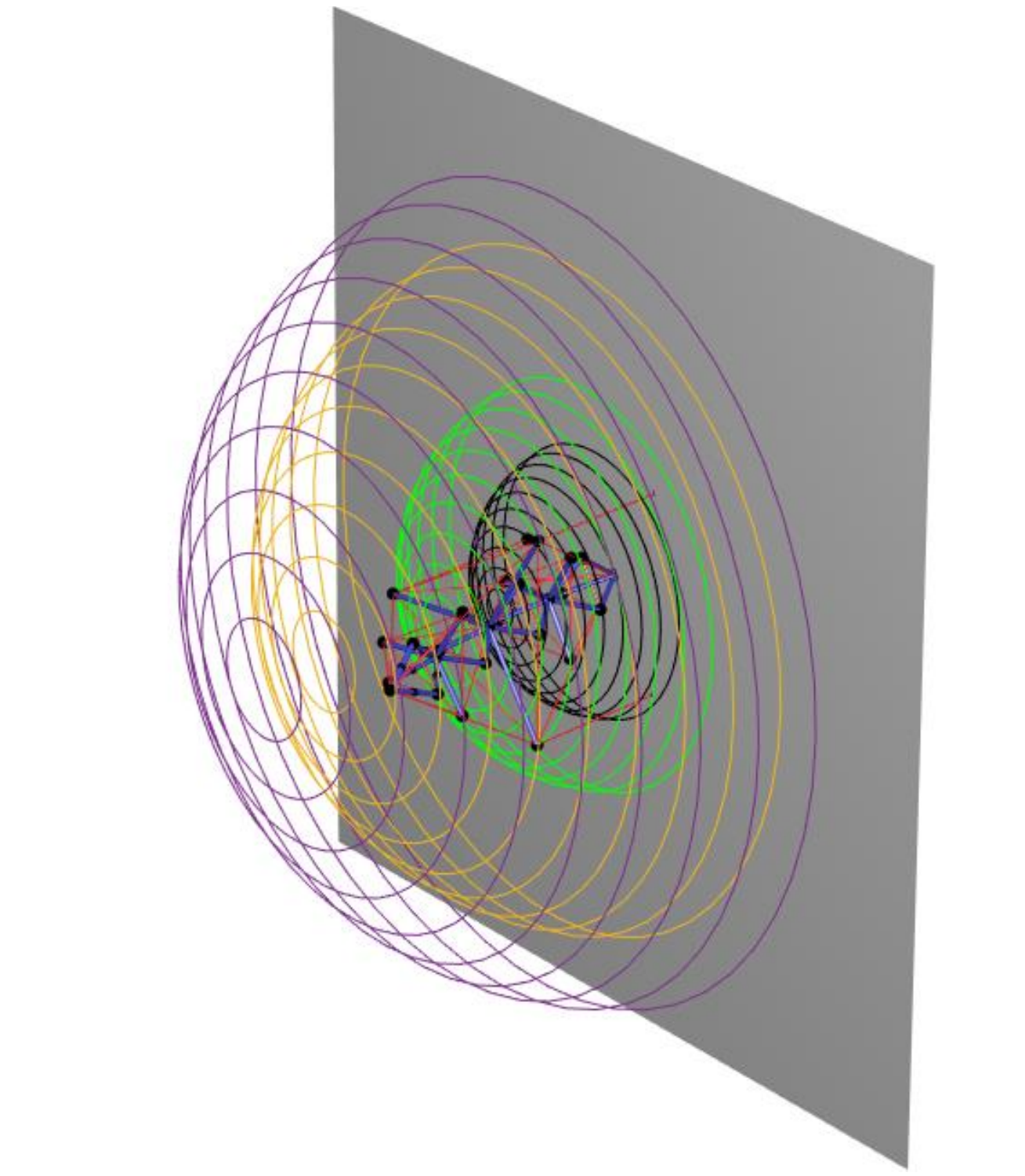}
\caption{Contour circles representing a similar motion to reach the desired position of end-effector in the complete control of the gyroscopic tensegrity robotic arm. The contour circles are shown for four different reachable space hemispheres.}
\label{f:ReachableSpace}
\end{figure}

In traditional robotics, optimal use of redundantly actuated manipulators is carried out to maximize dexterity while staying away from kinematic singularities [\cite{Nakamura_1990}]. Control effort is typically not included as performance measures and kinematic criteria are often employed. The simplicity of vector dynamics approaches, albeit non-minimal, are quite useful in including control effort for optimal use of redundancy. An objective of the complete motion analysis is to provide control effort measures in addition to kinematic approaches for managing actuator redundancy. This is explained using an example. 
The previous subsections show the result for two independent movements of the robotic arm to reach a given point in space with the extension of the arm followed by the angular rotation. In this subsection, the results are shown for a single shape control simulation to reach any given point in space. Figure~\ref{f:ReachableSpace} shows the contour circles for four different 3D hemispheres. 
\begin{figure}[h!]
\centering
\includegraphics[width=1\linewidth]{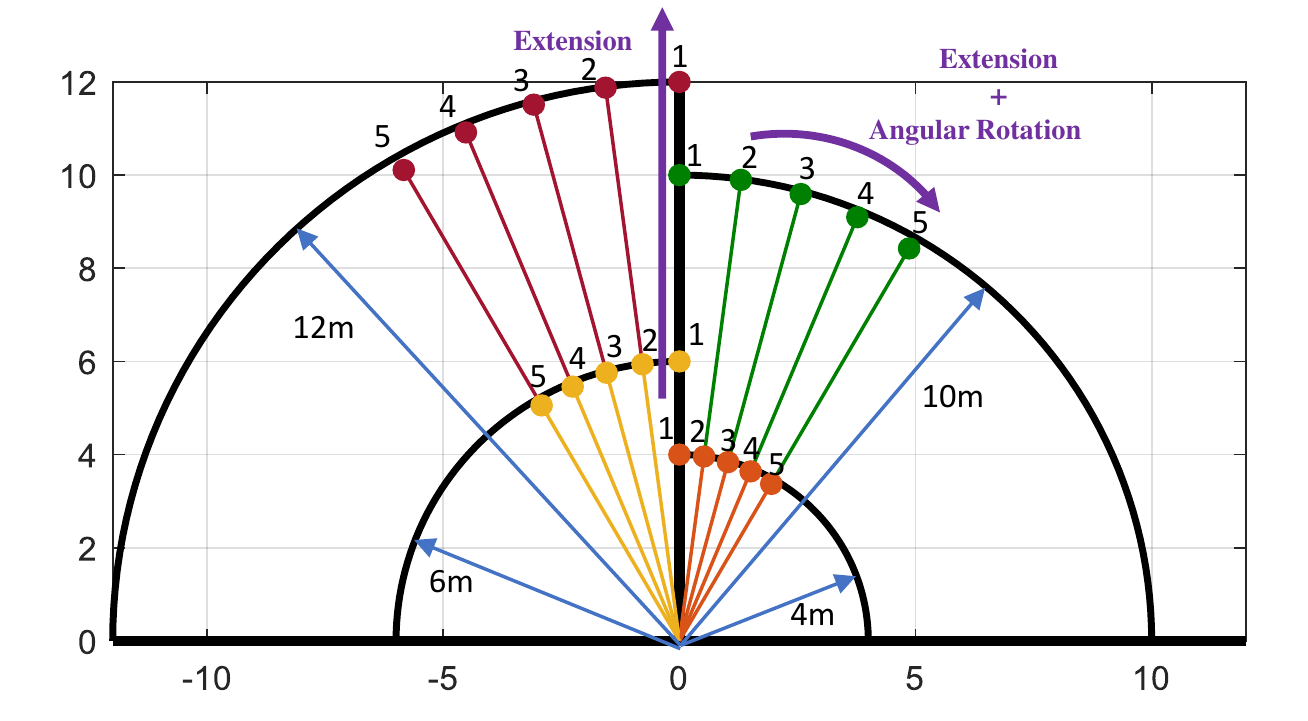}
\caption{The figure shows various reachable points for the end-effector of the robotic arm in a polar coordinate representation.}
\label{f:ReachableSpace2}
\end{figure}
\begin{figure}[h!]
\centering
\includegraphics[width=1\linewidth]{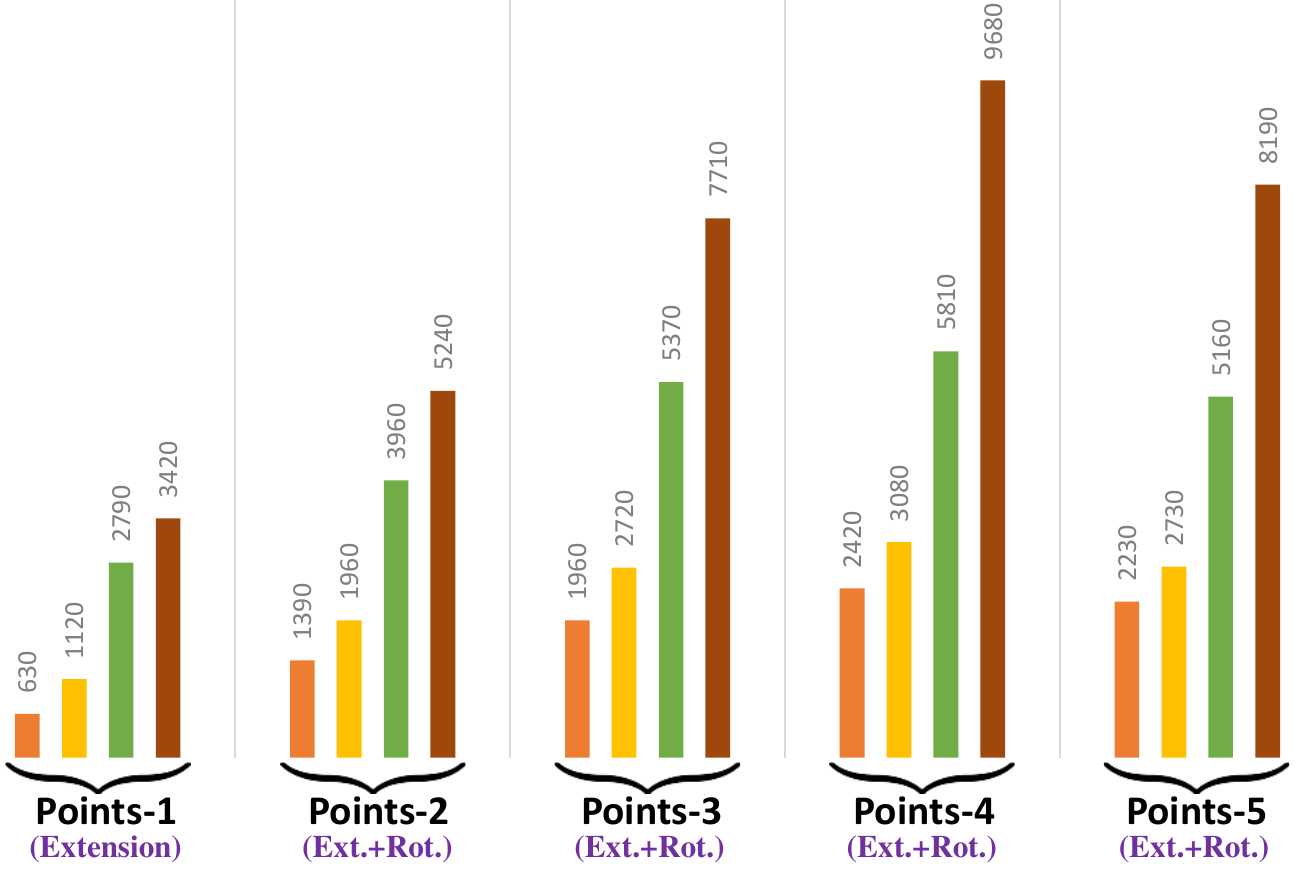}
\caption{The figure shows values of avg.($\gamma$) for the end-effector to reach different points given in Fig.~\ref{f:ReachableSpace2}.}
\label{f:ReachableSpace3}
\end{figure}
The contour circles represent the symmetric space where the end-effector will reach the desired point with similar motion and control actions of the strings due to the rotational symmetry about the x-axis (direction of extension).
The simulation is performed for the final position to be any point on the contour circles of hemispheres with the starting position as the configuration given in Fig.~\ref{f:Error_stow_Initial}.

A more detailed inference of the control required to move the end-effector in 3D space can be drawn from Figs.~\ref{f:ReachableSpace2} and \ref{f:ReachableSpace3}. A spherical coordinate representation of the workspace is used to identify trends in control required corresponding to the traversal point from the initial configuration (see Fig.~\ref{f:Error_stow_Initial}). The points are chosen for four different reaches of 4m, 6m, 10m and 12m with five different angular rotations. The points numbered ``1" are achieved with the only extension of the arm and points numbered ``2, 3, 4 and 5" are achieved with simultaneous extension and angular rotation as shown in Fig.~\ref{f:ReachableSpace2}. Figure~\ref{f:ReachableSpace3} provides the values of avg.($\gamma$) control required to move the end-effector in 3D space with four different colors representing the four different reach of 4m (red), 6m (yellow), 10m (green) and 12m(maroon) corresponding to Fig.~\ref{f:ReachableSpace2}. 
The avg.($\gamma$) is the average value of the force density calculated as the sum of force density for all the strings averaged with the total number of time-steps. 
Notice the systematic increase in control required to move the arm from the initial configuration to achieve the increasing reach from 4m to 12m. The second observation is the increase in control effort to achieve a large angular rotation going from points numbered ``1" to points numbered ``2", ``3" and ``4" for the chosen control gains of $\Omega = 30I$ and $\Psi = 20I$ with a total of 5000 time-steps. For points numbered ``5", smaller control gains had to be used to achieve the desired end point location.
The decrease in control effort to reach points numbered ``5" in Fig.~\ref{f:ReachableSpace2} can be attributed to thin gain change. In conclusion, these path-dependent and systematic costs can be used to augment traditional measures of dexterity to control the shape change of complex tensegrity systems. These preliminary demonstrations provide optimism to utilize redundancy to ensure control economy stability.

\subsection{Disturbance rejection with different bounds on error}
\begin{figure}[ht!]
\centering
\includegraphics[width=.9\linewidth]{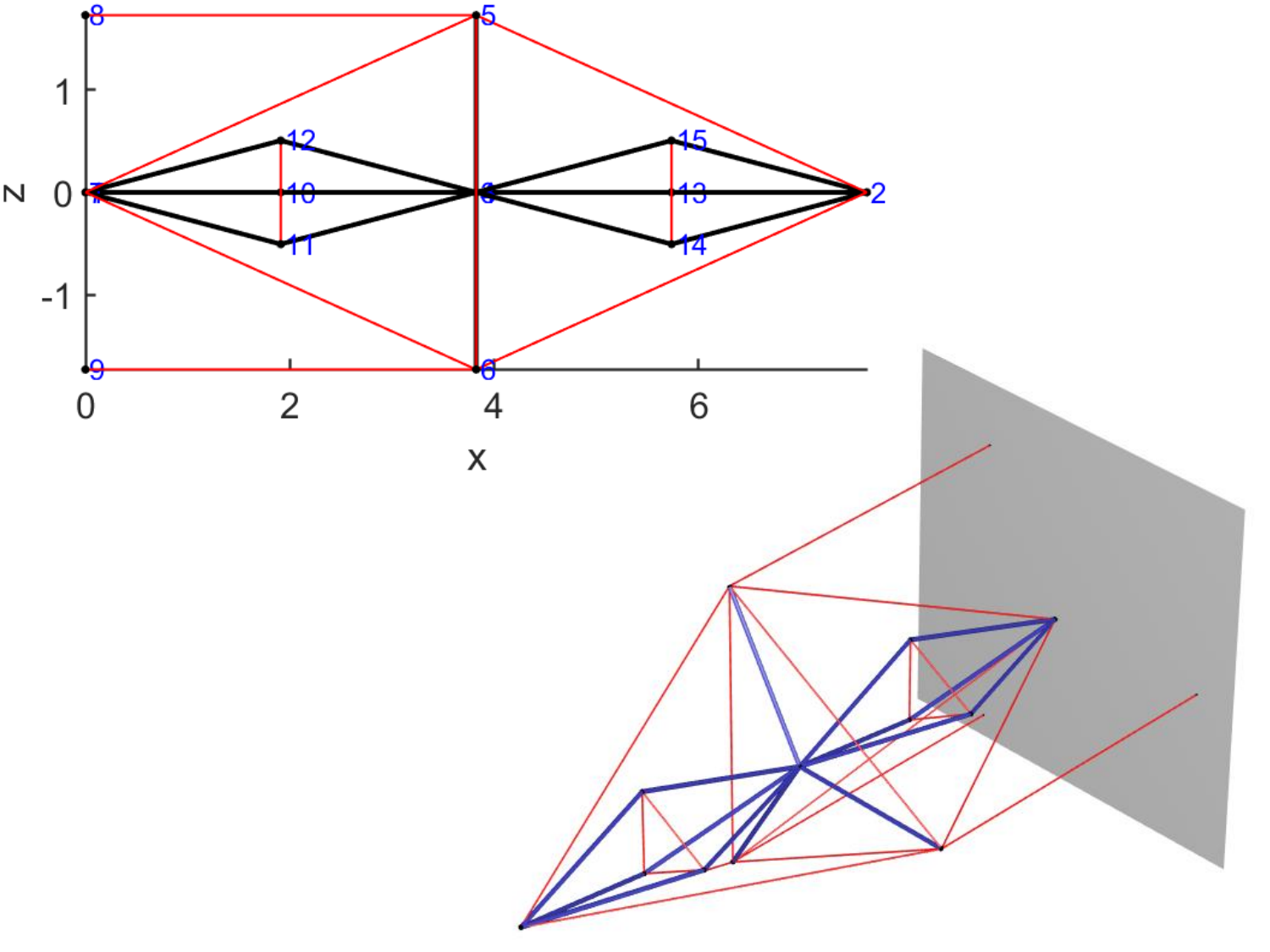}
\caption{Initial configuration of the tensegrity $T_1D_1$ robotic arm.}
\label{f:T_1D_1_config}
\end{figure}

The tensegrity $T_1D_1$ robotic arm with the initial configuration shown in Fig.~\ref{f:T_1D_1_config} is used to demonstrate the efficacy of the disturbance rejection formulation discussed in Section 5. 
A {\tt MATLAB}\textregistered~based CVX toolbox is used for numerical implementation [\cite{cvx}].

\begin{figure*}[h!]
\begin{multicols}{1}
    \centering
      \subfloat{\includegraphics[width=2\linewidth]{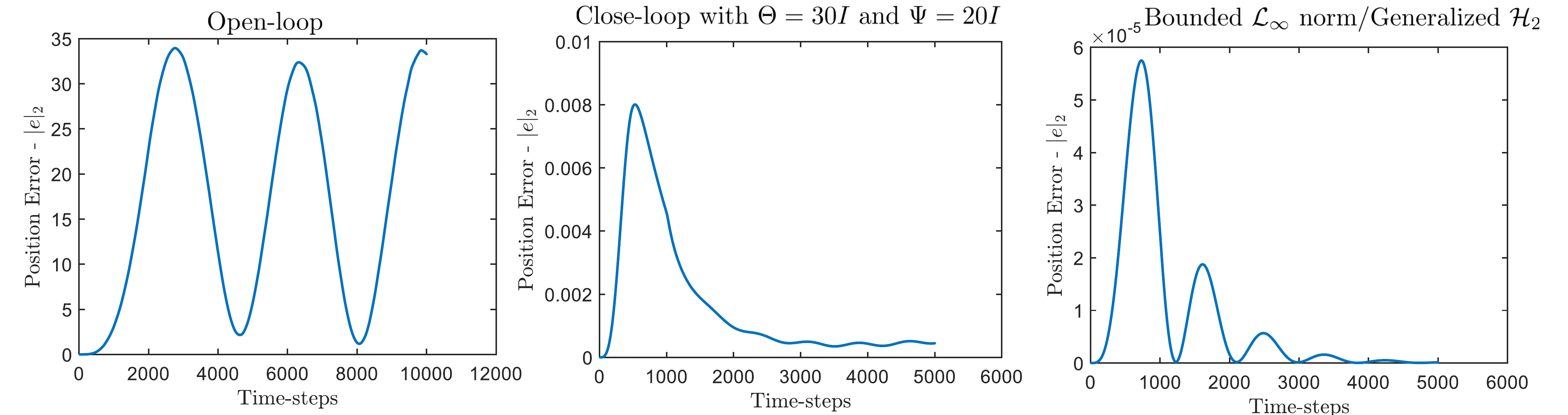}}  
\end{multicols}
\caption{Plots of norm in position error for open-loop, closed loop with $\Theta = 30I$ and $\Psi = 20I$, and gains calculated using the bounded $\mathcal{L}_\infty$ norm or Generalized $\mathcal{H}_2$ Problem for the tensegrity robotic arm.}
\label{f:Linf_LMI2}
\end{figure*}
\begin{figure*}[h!]
\begin{multicols}{1}
    \centering
      \subfloat{\includegraphics[width=2\linewidth]{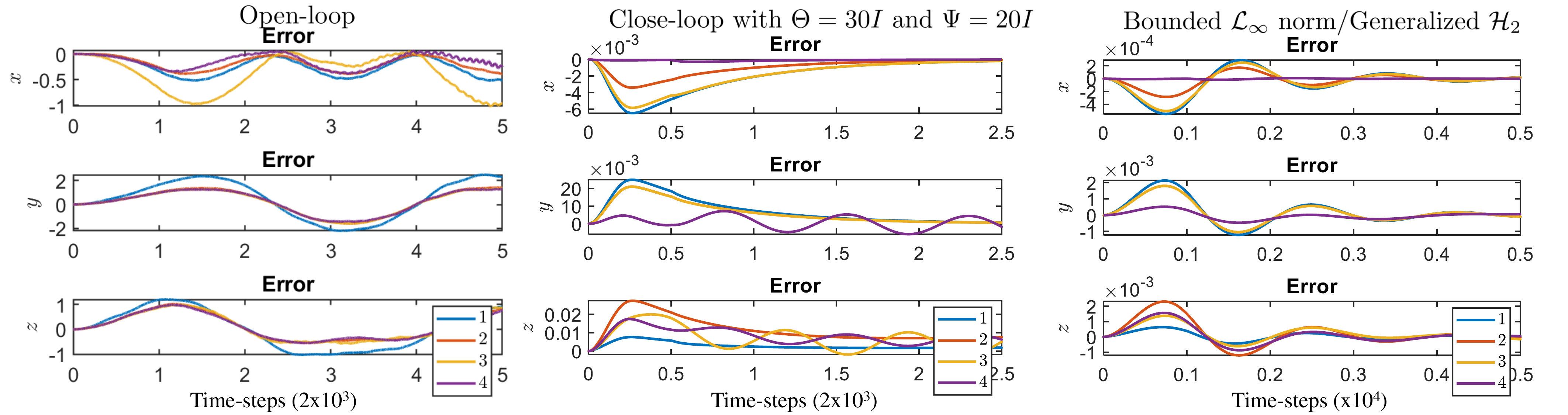}}  
\end{multicols}
\caption{Trajectories of node positions ($n_1$, $n_2$, $n_3$ and $n_4$) for open-loop, closed loop with $\Theta = 30I$ and $\Psi = 20I$, and gains calculated using the bounded $\mathcal{L}_\infty$ norm or Generalized $\mathcal{H}_2$ Problem for the tensegrity robotic arm.}
\label{f:Linf_LMI2_Err}
\end{figure*}

The first plot from Fig.~\ref{f:Linf_LMI2} shows the norm of position error for the open-loop simulation for a finite energy force disturbance given at all node locations. The plot shows a periodic motion for the position error from the initial nominal configuration. The second plot in Fig.~\ref{f:Linf_LMI2} shows the norm in position error for the closed-loop system with gains value $\Theta = 30I$ and $\Psi = 20I$. The chosen stable gains derive the system to zero but with a large value of peak value of error. The last plot in Fig.~\ref{f:Linf_LMI2} shows the closed-loop performance with the gains calculated using the bounded $\mathcal{L}_\infty$ norm or generalized $\mathcal{H}_2$ problem for the depicted robotic arm. Notice that the peak value of the error has been brought down considerably using the LMI formulation described earlier. 
\begin{figure}[ht!]
\centering
\includegraphics[width=.8\linewidth]{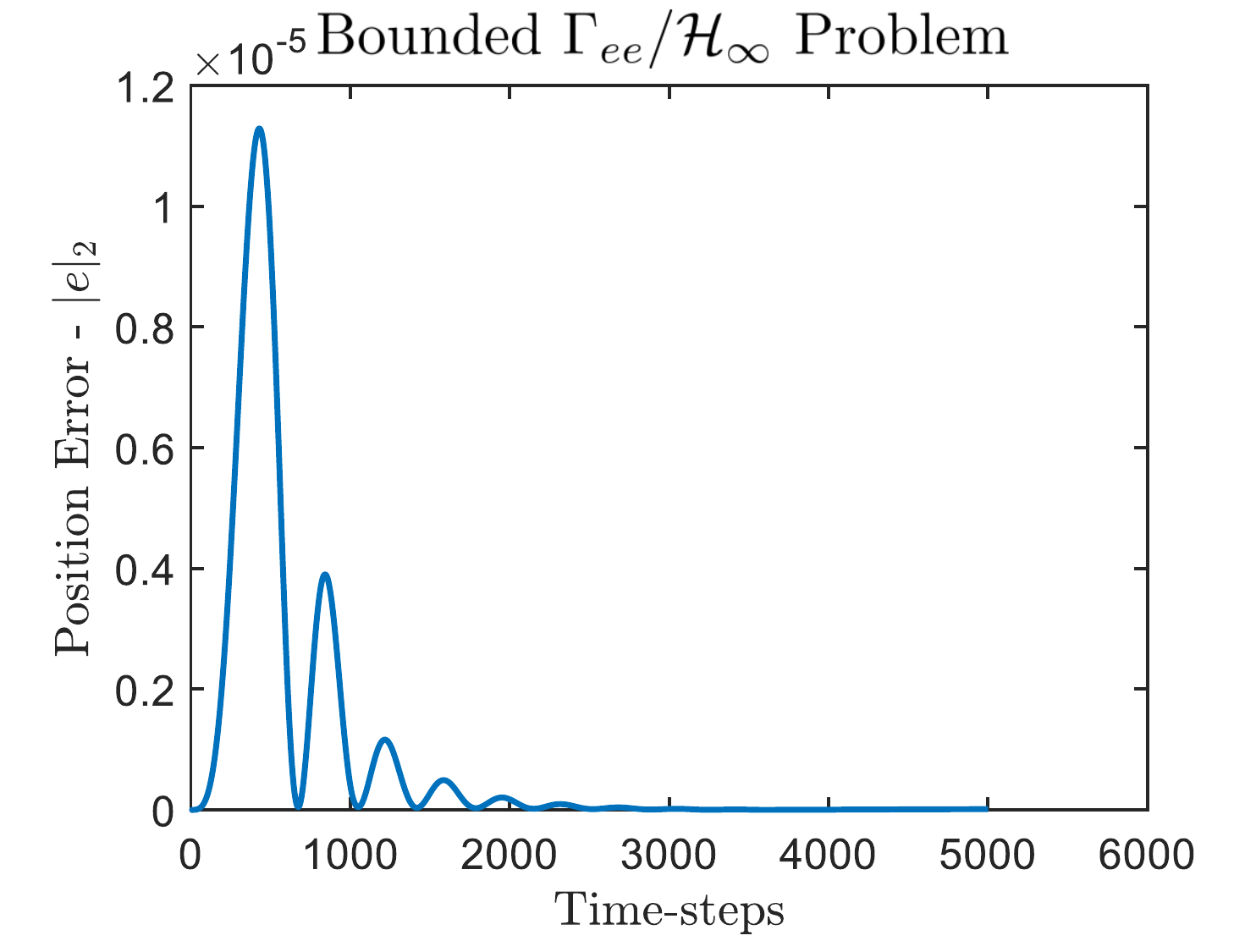}
\caption{Plots of norm in position error for gains calculated using the bounded $\Gamma_{ee}$ or $\mathcal{H}_\infty$ problem for the $T_1D_1$ robotic arm.}
\label{f:Hinf_LMI5}
\end{figure}
The theoretical value of the gain by solving the LMIs in Lemma~\ref{Lem1} was calculated to be $\Gamma_{ep} = 1.00 \times 10^{-4}$ and the gain calculated using the nonlinear simulation result was observed to be $\Gamma_{ep} = 6.82 \times 10^{-7}$, which satisfies the requirement. Moreover, the value from the simulation results was considerably smaller as the disturbance values of the output will only match the theoretical results for the worst-case disturbance.
Figure~\ref{f:Linf_LMI2_Err} shows the trajectories for error in node positions for nodes ($n_1$, $n_2$, $n_3$ and $n_4$) for all the three cases mentioned earlier.




\begin{figure}[ht!]
\centering
\includegraphics[width=.8\linewidth]{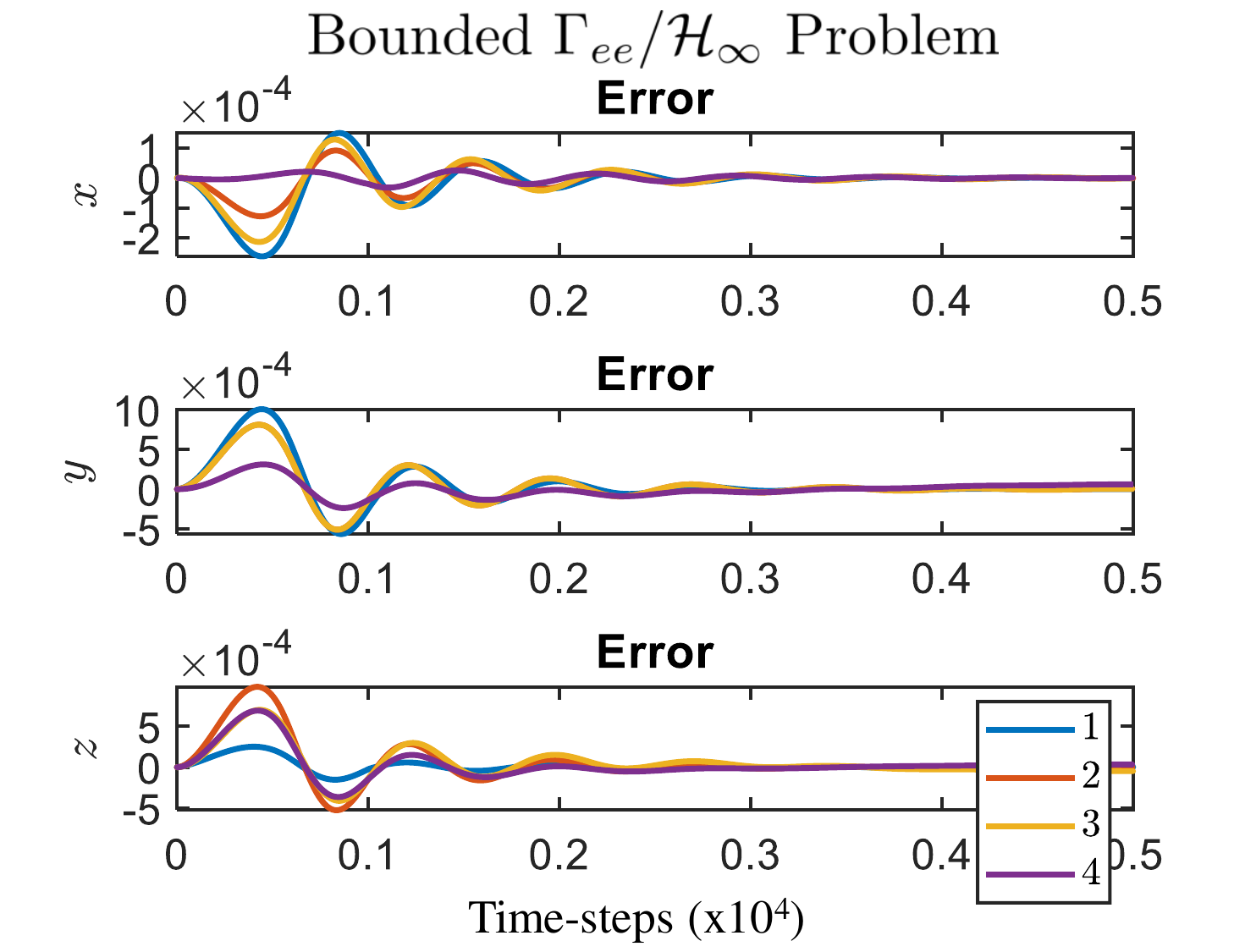}
\caption{Trajectories of node positions ($n_1$, $n_2$, $n_3$ and $n_4$) for the gains calculated using the bounded $\Gamma_{ee}$ or $\mathcal{H}_\infty$ problem for the $T_1D_1$ robotic arm.}
\label{f:Hinf_LMI5_Err}
\end{figure}

The following simulation results are generated using the same energy bounded disturbance used in the analysis of bounded $\mathcal{L}_\infty$ norm. 
The first two plots for $\mathcal{H}_\infty$ results would be the same as Fig.~\ref{f:Linf_LMI2}, showing the norm of position error for the open-loop simulation for the closed-loop system with gains value $\Theta = 30I$ and $\Psi = 20I$. 
\begin{figure*}[ht!]
\begin{multicols}{1}
    \centering
      \subfloat{\includegraphics[width=2\linewidth]{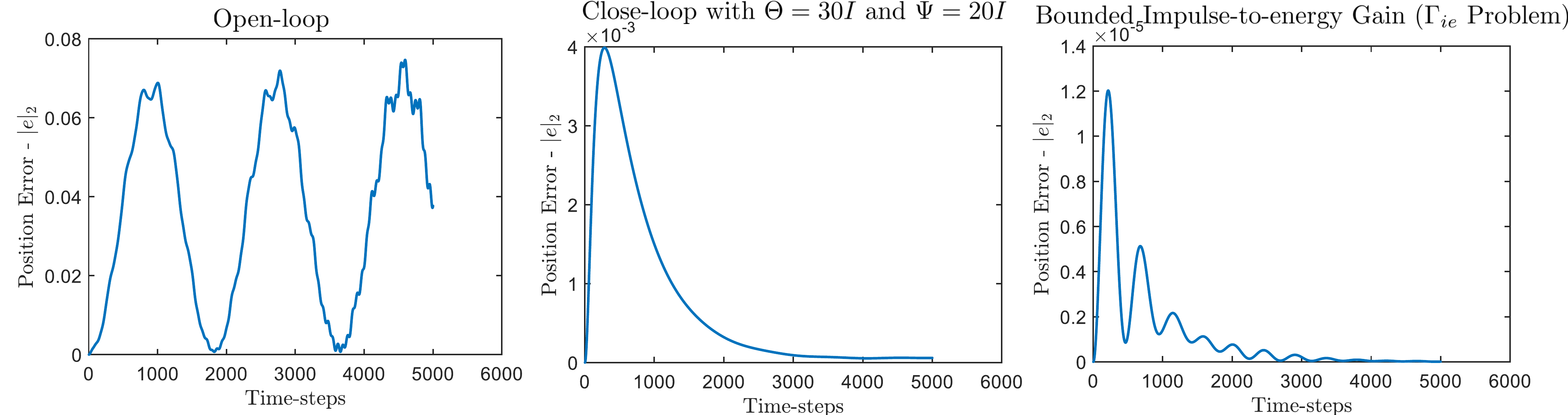}}  
\end{multicols}
\caption{Plots of norm in position error for open-loop, closed loop with $\Theta = 30I$ and $\Psi = 20I$, and gains calculated using the bounded impulse to energy problem ($\Gamma_{ie}$) for the $T_1D_1$ robotic arm.}
\label{f:IE_LMI4}
\end{figure*}
\begin{figure*}[ht!]
\begin{multicols}{1}
    \centering
      \subfloat{\includegraphics[width=2\linewidth]{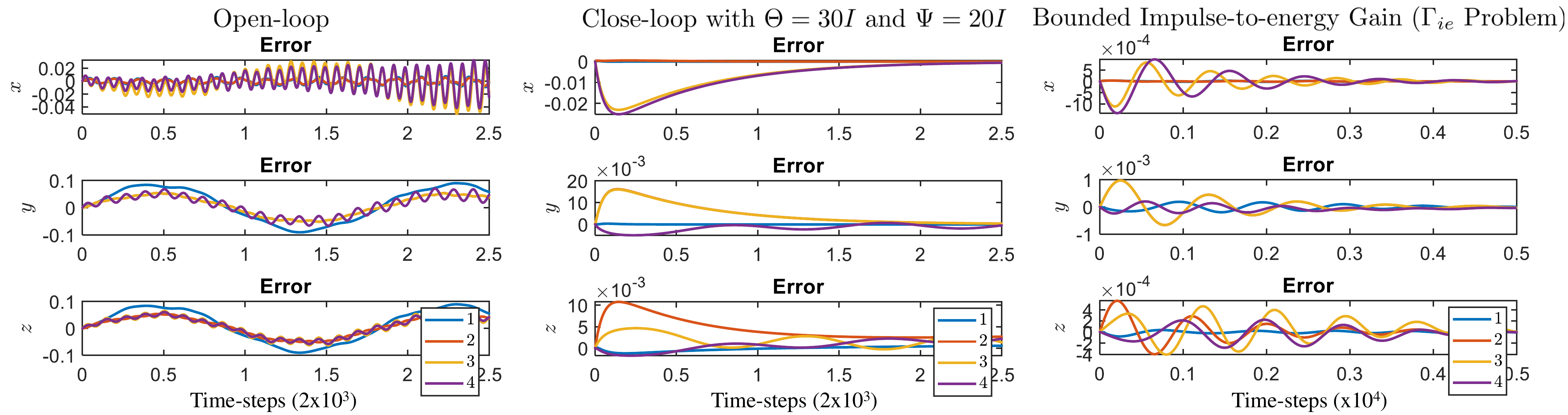}}  
\end{multicols}
\caption{Trajectories of node positions ($n_1$, $n_2$, $n_3$ and $n_4$) for open-loop, closed loop with $\Theta = 30I$ and $\Psi = 20I$, and gains calculated using the bounded impulse to energy problem ($\Gamma_{ie}$) for the $T_1D_1$ robotic arm.}
\label{f:IE_LMI4_Err}
\end{figure*}
The plot in Fig.~\ref{f:Hinf_LMI5} shows the closed-loop performance with the gains calculated using the bounded $\Gamma_{ee}$ or $\mathcal{H}_\infty$ problem. 
The theoretical value of the energy-to-energy gain was obtained to be $\Gamma_{ee} = 1.27 \times 10^{-5}$ by solving the LMIs in Lemma~\ref{Lem2} and the gain in energy from disturbance to error for the nonlinear simulation was observed to be $\Gamma_{ee} = 5.14 \times 10^{-9}$, which satisfies the requirement. Just as in the previous case, there would be some worst-case disturbance for this nonlinear system that would cause the error due to disturbance to meet the theoretical value.
The similar trend can be observed from Fig.~\ref{f:Hinf_LMI5_Err}, which shows the trajectories for error in node positions ($n_1$, $n_2$, $n_3$ and $n_4$) for all the three cases mentioned earlier.




The first plot from Fig.~\ref{f:IE_LMI4} shows the norm of position error for the open-loop simulation for an impulsive disturbance given at all node locations in terms of force. A periodic motion for the position error from the initial nominal configuration was observed for the open-loop system. The second plot in Fig.~\ref{f:IE_LMI4} shows the result for the closed-loop system with gains value $\Theta = 30I$ and $\Psi = 20I$. The chosen stable gains derive the system to zero but with large energy of the error ($\|{ y}\|_{\mathcal{L}_2}$) as compared to the simulation obtained with the calculated gains using the  Bounded $\Gamma_{ie}$ solution. 
Notice the initial spike in error, but the total energy for the entire simulation is substantially less than the other two simulations. 
The theoretical value of the impulse-to-energy gain was obtained to be $\Gamma_{ie} = 1.98 \times 10^{-5}$ by solving the LMIs in Lemma~\ref{Lem3} and the gain from impulsive disturbance to error energy for the simulation was observed to be $\Gamma_{ie} = 1.04 \times 10^{-10}$, which satisfies the requirement. 
Figure~\ref{f:IE_LMI4_Err} shows the trajectories for error in node positions ($n_1$, $n_2$, $n_3$ and $n_4$) for all the three cases mentioned earlier. Notice that the values of the error in the last plot are much smaller than the values corresponding to the first two plots.

\section{Conclusions}
In this paper, a model-based nonlinear control methodology to drive the node-positions/shape of a very high DOF tensegrity robot is provided. 
A shape control algorithm for both standard and gyroscopic tensegrity system is developed, which allows for added orientation control of the structure.
A generalized reduced-order vector form of tensegrity dynamics in the presence of given disturbance is formulated and a second-order output regulator was used to control the shape, velocity, and acceleration of certain nodes.
The solutions for control gains to generate different kinds of performance bounds formulated in the LMI framework are shown to be soluble as convex semi-definite programming problems. 
A tensegrity $T_2D_1$ robotic arm was introduced and controlled to show the efficacy of the formulation by allowing the end effector to reach the desired point and orientation in a 3-dimensional space. The results are further detailed to reject energy bounded disturbance using the formulation for \textit{Generalized} $\mathcal{H}_2$, $\mathcal{H}_\infty$ and \textit{Impulse-to-energy gain problem}.





\bibliographystyle{IEEEtran}
\bibliography{Raman_Tensegrity.bib,Raman_Dynamics.bib,Raman_Control.bib}

\begin{thebibliography}{10}
\providecommand{\url}[1]{#1}
\csname url@samestyle\endcsname
\providecommand{\newblock}{\relax}
\providecommand{\bibinfo}[2]{#2}
\providecommand{\BIBentrySTDinterwordspacing}{\spaceskip=0pt\relax}
\providecommand{\BIBentryALTinterwordstretchfactor}{4}
\providecommand{\BIBentryALTinterwordspacing}{\spaceskip=\fontdimen2\font plus
\BIBentryALTinterwordstretchfactor\fontdimen3\font minus
  \fontdimen4\font\relax}
\providecommand{\BIBforeignlanguage}[2]{{%
\expandafter\ifx\csname l@#1\endcsname\relax
\typeout{** WARNING: IEEEtran.bst: No hyphenation pattern has been}%
\typeout{** loaded for the language `#1'. Using the pattern for}%
\typeout{** the default language instead.}%
\else
\language=\csname l@#1\endcsname
\fi
#2}}
\providecommand{\BIBdecl}{\relax}
\BIBdecl

\bibitem{Snelson_1965}
K.~D. Snelson, ``Continuous tension, discontinuous compression structures,''
  Feb.~16 1965, uS Patent 3,169,611.

\bibitem{Skelton_2009_Tensegrity_Book}
R.~E. Skelton and M.~C. de~Oliveira, \emph{Tensegrity Systems}.\hskip 1em plus
  0.5em minus 0.4em\relax Springer US, 2009.

\bibitem{Rieffel_softRobot}
J.~Rieffel and J.-B. Mouret, ``Adaptive and resilient soft tensegrity robots,''
  \emph{Soft Robotics}, vol.~5, no.~3, p. 318‐329, 2018.

\bibitem{Ingber_1998}
D.~E. Ingber, ``The architecture of life,'' \emph{Scientific America}, vol.
  278, no.~1, pp. 48--57, 1998.

\bibitem{Tibert_Pellegrino_2011_Review}
A.~G. Tibert and S.~Pellegrino, ``Review of form-finding methods for tensegrity
  structures,'' \emph{International Journal of Space Structures}, vol.~26,
  no.~3, pp. 241--255, 2011.

\bibitem{Arsenault_Gosselin_2008_IJRR}
M.~Arsenault and C.~M. Gosselin, ``Kinematic and static analysis of a
  three-degree-of-freedom spatial modular tensegrity mechanism,'' \emph{The
  International Journal of Robotics Research}, vol.~27, no.~8, pp. 951--966,
  2008.

\bibitem{Goyal_2020_MRC}
R.~Goyal, R.~Skelton, and E.~A. Peraza~Hernandez, ``Design of minimal mass
  load-bearing tensegrity lattices,'' \emph{Mechanics Research Communications},
  vol. 103, p. 103477, 2020.

\bibitem{Skelton_2010_Michell}
R.~E. Skelton and M.~C. de~Oliveira, ``Optimal tensegrity structures in
  bending: the discrete {M}ichell truss,'' \emph{Journal of the Franklin
  Institute}, vol. 347, no.~1, pp. 257--283, 2010.

\bibitem{Caluwaerts_2014}
K.~Caluwaerts, J.~Despraz, A.~Işçen, A.~P. Sabelhaus, J.~Bruce, B.~Schrauwen,
  and V.~SunSpiral, ``Design and control of compliant tensegrity robots through
  simulation and hardware validation,'' \emph{Journal of The Royal Society
  Interface}, vol.~11, no.~98, 2014.

\bibitem{Goyal_2019_Buckling}
R.~Goyal, E.~A. Peraza~Hernandez, and R.~Skelton, ``Analytical study of
  tensegrity lattices for mass-efficient mechanical energy absorption,''
  \emph{International Journal of Space Structures}, vol.~34, no. 1-2, pp.
  3--21, 2019.

\bibitem{Tibert_Pellegrino_2003_Mast}
A.~G. Tibert and S.~Pellegrino, ``Deployable tensegrity mast,'' \emph{In: 44th
  AIAA/ASME/ASCE/AHS/ASC, Structures, Structural Dynamics and Materials
  Conference and Exhibit, Norfolk , VA, USA.}, p. 1978, 2003.

\bibitem{yang2019deployment}
S.~Yang and C.~Sultan, ``Deployment of foldable tensegrity-membrane systems via
  transition between tensegrity configurations and tensegrity-membrane
  configurations,'' \emph{International Journal of Solids and Structures}, vol.
  160, pp. 103--119, 2019.

\bibitem{Peng_2018_DynamicDeploy}
Z.~Kan, H.~Peng, B.~Chen, and W.~Zhong, ``A sliding cable element of multibody
  dynamics with application to nonlinear dynamic deployment analysis of
  clustered tensegrity,'' \emph{International Journal of Solids and
  Structures}, vol. 130-131, pp. 61 -- 79, 2018.

\bibitem{Sabelhaus_2017_ACC}
A.~P. {Sabelhaus}, A.~K. {Akella}, Z.~A. {Ahmad}, and V.~{SunSpiral},
  ``Model-predictive control of a flexible spine robot,'' in \emph{2017
  American Control Conference (ACC)}, May 2017, pp. 5051--5057.

\bibitem{SoftRobots2_IJRR}
C.~D. Santina, R.~K. Katzschmann, A.~Bicchi, and D.~Rus, ``Model-based dynamic
  feedback control of a planar soft robot: trajectory tracking and interaction
  with the environment,'' \emph{The International Journal of Robotics
  Research}, vol.~39, no.~4, pp. 490--513, 2020.

\bibitem{Karnan_Goyal_2017_IROS}
H.~Karnan, R.~Goyal, M.~Majji, R.~E. Skelton, and P.~Singla, ``Visual feedback
  control of tensegrity robotic systems,'' \emph{2017 IEEE/RSJ International
  Conference on Intelligent Robots and Systems (IROS)}, pp. 2048--2053, Sept
  2017.

\bibitem{Goyal_2019_SEMC}
R.~Goyal and R.~E. Skelton, ``Integrating structure and control design using
  tensegrity paradigm,'' in \emph{Advances in Engineering Materials, Structures
  and Systems: Innovations, Mechanics and Applications}.\hskip 1em plus 0.5em
  minus 0.4em\relax CRC Press, 2019, pp. 989--994.

\bibitem{Edwin_plate}
S.~Jiang, R.~E. Skelton, and E.~A.~P. Hernandez, ``Analytical equations for the
  connectivity matrices and node positions of minimal and extended tensegrity
  plates,'' \emph{International Journal of Space Structures}, p.
  0956059920902375, 2020.

\bibitem{tibert2002_ReflectorSatellites}
A.~Tibert and S.~Pellegrino, ``Deployable tensegrity reflectors for small
  satellites,'' \emph{Journal of Spacecraft and Rockets}, vol.~39, no.~5, pp.
  701--709, 2002.

\bibitem{moored_MorphWing}
K.~W. Moored and H.~Bart-Smith, ``The analysis of tensegrity structures for the
  design of a morphing wing,'' \emph{Journal of applied mechanics}, vol.~74,
  no.~4, pp. 668--676, 2007.

\bibitem{Murakami_2001a_Dynamics}
H.~Murakami, ``Static and dynamic analyses of tensegrity structures. part 1.
  nonlinear equations of motion,'' \emph{International Journal of Solids and
  Structures}, vol. 38(20), pp. 3599--3613, 2001.

\bibitem{Goyal_Dynamics_2019}
R.~Goyal and R.~Skelton, ``Tensegrity system dynamics with rigid bars and
  massive strings,'' \emph{Multibody System Dynamics}, vol. 46(3), pp.
  203--228, 2019.

\bibitem{Peng_2018_AIAA}
Z.~Kan, H.~Peng, and B.~Chen, ``Complementarity framework for nonlinear
  analysis of tensegrity structures with slack cables,'' \emph{AIAA Journal},
  vol.~56, no.~12, pp. 5013--5027, 2018.

\bibitem{Varol_2019_IEEE_Access}
D.~{Fadeyev}, A.~{Zhakatayev}, A.~{Kuzdeuov}, and H.~A. {Varol}, ``Generalized
  dynamics of stacked tensegrity manipulators,'' \emph{IEEE Access}, vol.~7,
  pp. 63\,472--63\,484, 2019.

\bibitem{Paul_2006}
C.~Paul, F.~J. Valero-Cuevas, and H.~Lipson, ``Design and control of tensegrity
  robots for locomotion,'' \emph{IEEE Transactions on Robotics}, vol.~22,
  no.~5, pp. 944--957, 2006.

\bibitem{Zhang_Levine_2017_ICRA}
M.~{Zhang}, X.~{Geng}, J.~{Bruce}, K.~{Caluwaerts}, M.~{Vespignani},
  V.~{SunSpiral}, P.~{Abbeel}, and S.~{Levine}, ``Deep reinforcement learning
  for tensegrity robot locomotion,'' in \emph{2017 IEEE International
  Conference on Robotics and Automation (ICRA)}, May 2017, pp. 634--641.

\bibitem{IanSmith_RL}
B.~Adam and I.~F. Smith, ``Reinforcement learning for structural control,''
  \emph{Journal of Computing in Civil Engineering}, vol.~22, no.~2, pp.
  133--139, 2008.

\bibitem{Bekris_2019_IJRR2}
D.~Surovik, K.~Wang, M.~Vespignani, J.~Bruce, and K.~E. Bekris, ``Adaptive
  tensegrity locomotion: Controlling a compliant icosahedron with
  symmetry-reduced reinforcement learning,'' \emph{The International Journal of
  Robotics Research}, vol.~0, no.~0, p. 0278364919859443, 2019.

\bibitem{Wang_2020_RAL}
R.~Wang, R.~Goyal, S.~Chakravorty, and R.~E. Skelton, ``Model and data based
  approaches to the control of tensegrity robots,'' \emph{IEEE Robotics and
  Automation Letters}, vol. 5(3), pp. 3846 -- 3853, 2020.

\bibitem{IanSmith_CivilControl}
B.~Adam and I.~F. Smith, ``Active tensegrity: A control framework for an
  adaptive civil-engineering structure,'' \emph{Computers \& Structures},
  vol.~86, no.~23, pp. 2215 -- 2223, 2008.

\bibitem{LandolfBarbarigos_Thesis}
L.-G.-A. Rhode-Barbarigos, ``An active deployable tensegrity structure,'' p.
  188, 2012.

\bibitem{Bekris_2019_IJRR1}
Z.~Littlefield, D.~Surovik, M.~Vespignani, J.~Bruce, W.~Wang, and K.~E. Bekris,
  ``Kinodynamic planning for spherical tensegrity locomotion with effective
  gait primitives,'' \emph{The International Journal of Robotics Research},
  vol.~38, no. 12-13, pp. 1442--1462, 2019.

\bibitem{sabelhaus2015system}
A.~P. Sabelhaus, J.~Bruce, K.~Caluwaerts, P.~Manovi, R.~F. Firoozi, S.~Dobi,
  A.~M. Agogino, and V.~SunSpiral, ``System design and locomotion of
  {SUPER}ball, an untethered tensegrity robot,'' in \emph{IEEE international
  conference on robotics and automation (ICRA)}, 2015, pp. 2867--2873.

\bibitem{Koizumi_2012_ICRA}
Y.~Koizumi, M.~Shibata, and S.~Hirai, ``Rolling tensegrity driven by pneumatic
  soft actuators,'' in \emph{Robotics and Automation (ICRA), 2012 IEEE
  International Conference on}, 2012, Conference Proceedings, pp. 1988--1993.

\bibitem{blissIwasakiCPGexperimental}
T.~Bliss, J.~Werly, T.~Iwasaki, and H.~Bart-Smith, ``Experimental validation of
  robust resonance entrainment for cpg-controlled tensegrity structures,''
  \emph{IEEE Transactions On Control Systems Technology}, vol.~21, no.~3, pp.
  666--678, 2012.

\bibitem{blissIwasakiCPG}
T.~Bliss, T.~Iwasaki, and H.~Bart-Smith, ``Central pattern generator control of
  a tensegrity swimmer,'' \emph{IEEE/ASME Transactions on Mechatronics},
  vol.~18, no.~2, pp. 586--597, 2012.

\bibitem{PENG_2020_MPC}
H.~Peng, F.~Li, and Z.~Kan, ``A novel distributed model predictive control
  method based on a substructuring technique for smart tensegrity structure
  vibrations,'' \emph{Journal of Sound and Vibration}, vol. 471, p. 115171,
  2020.

\bibitem{Yang_Sultan_2017_IJRNC}
S.~{Yang} and C.~{Sultan}, ``Control-oriented modeling and deployment of
  tensegrity–membrane systems,'' \emph{International Journal of Robust and
  Nonlinear Control}, vol.~27, no.~16, pp. 2722--2748, 2017.

\bibitem{Yang_Sultan_2017_LPVControl}
S.~Yang and C.~Sultan, ``Lpv control of a tensegrity-membrane system,''
  \emph{Mechanical Systems and Signal Processing}, vol.~95, pp. 397 -- 424,
  2017.

\bibitem{Varol_Manpltr}
A.~{Kuzdeuov}, M.~{Rubagotti}, and H.~A. {Varol}, ``Neural network augmented
  sensor fusion for pose estimation of tensegrity manipulators,'' \emph{IEEE
  Sensors Journal}, vol.~20, no.~7, pp. 3655--3666, 2020.

\bibitem{Varol_RoboPlan}
B.~{Nurimbetov}, M.~{Issa}, and H.~A. {Varol}, ``Robotic assembly planning of
  tensegrity structures,'' in \emph{2019 IEEE/SICE International Symposium on
  System Integration (SII)}, 2019, pp. 73--78.

\bibitem{Goyal_2020_GYRO}
R.~Goyal, M.~Chen, M.~Majji, and R.~E. Skelton, ``Gyroscopic tensegrity
  robots,'' \emph{IEEE Robotics and Automation Letters}, vol. 5(2), pp.
  1239--1246, 2020.

\bibitem{Skelton_LMI_1998}
R.~E. Skelton, T.~Iwasaki, and K.~Grigoriadis, \emph{A Unified Algebraic
  Approach to Control Design.}\hskip 1em plus 0.5em minus 0.4em\relax Taylor \&
  Francis, London, UK, 1998.

\bibitem{Scherer_LMI_1997}
C.~{Scherer}, P.~{Gahinet}, and M.~{Chilali}, ``Multiobjective output-feedback
  control via lmi optimization,'' \emph{IEEE Transactions on Automatic
  Control}, vol.~42, no.~7, pp. 896--911, 1997.

\bibitem{IWASAKI_Hinf_1994}
T.~Iwasaki and R.~Skelton, ``All controllers for the general
  $\mathcal{H}_\infty$ control problem: Lmi existence conditions and state
  space formulas,'' \emph{Automatica}, vol.~30, no.~8, pp. 1307 -- 1317, 1994.

\bibitem{Gahinet_Apkarian_1994}
P.~Gahinet and P.~Apkarian, ``A linear matrix inequality approach to
  $\mathcal{H}_\infty$ control,'' \emph{International Journal of Robust and
  Nonlinear Control}, vol.~4, no.~4, pp. 421--448, 1994.

\bibitem{boyd2004convex}
S.~Boyd and L.~Vandenberghe, \emph{Convex optimization}.\hskip 1em plus 0.5em
  minus 0.4em\relax Cambridge university press, 2004.

\bibitem{S-Lemma_2007}
I.~Pólik and T.~Terlaky, ``A survey of the s-lemma,'' \emph{SIAM Review},
  vol.~49, no.~3, pp. 371--418, 2007.

\bibitem{Goyal_MOTES_2019}
R.~Goyal, M.~Chen, M.~Majji, and R.~E. Skelton, ``Motes: Modeling of tensegrity
  structures,'' \emph{Journal of Open Source Software}, vol. 4(42), p. 1613,
  2019.

\bibitem{Nakamura_1990}
Y.~Nakamura, \emph{Advanced Robotics: Redundancy and Optimization},
  1st~ed.\hskip 1em plus 0.5em minus 0.4em\relax USA: Addison-Wesley Longman
  Publishing Co., Inc., 1990.

\bibitem{cvx}
M.~Grant and S.~Boyd, ``{CVX}: Matlab software for disciplined convex
  programming, version 2.1,'' \url{http://cvxr.com/cvx}, Mar. 2014.

\end{thebibliography}

\section*{Appendix A : Proofs for lemmas}
\noindent \textit{Proof of Lemma~\ref{Lem1}:} The gain matrix $G$ can be calculated by substituting Eq.~(\ref{eq:sys_Gain_close}) in Eq.~(\ref{eq:Linf_gain}) to give: 
\begin{multline}
    \min \epsilon,  \hspace{0.5 pc} \epsilon I - CQC\T > 0, \\ (A_p + B_p G) Q+Q(A_p + B_p G)\T + B_{cl}B_{cl}\T < 0,
\end{multline} 
which can be written as the following after using the Schur's complement with $Q > 0$ as:
\begin{align}
    \min \epsilon,  \hspace{0.25 pc} \begin{bmatrix} \epsilon I & CQ \\ QC\T & Q \end{bmatrix} > 0, \hspace{0.25 pc}  \begin{bmatrix} sym(A_p Q + B_p G Q) & B_{cl} \\ B_{cl}\T & -I \end{bmatrix} < 0, 
\end{align}
which further can be substituted as $R = G Q$ to obtain:
\begin{align}
    &\begin{bmatrix} sym(A_p Q + B_p R) & B_{cl} \\ B_{cl}\T & -I \end{bmatrix} < 0. 
\end{align}
\hfill \qedsymbol

\noindent \textit{Proof of Lemma~\ref{Lem2}:} The $\mathcal{H}_\infty$ problem with given $\epsilon$ and for a positive definite matrix $P>0$ can be solved with the following matrix inequality [\cite{Scherer_LMI_1997,Gahinet_Apkarian_1994}]:
\begin{multline}
P>0, ~ R = \epsilon^2 I > 0, \\
    PA_{cl}+A_{cl}\T P + PB_{cl}R^{-1}(PB_{cl})\T + C\T C < 0, 
\end{multline}
which can be written as:
\begin{multline}
   Y = P^{-1} >0, ~ R = \epsilon^2 I > 0, \\
   A_{cl}Y +Y A_{cl}\T + B_{cl}R^{-1}B_{cl}\T + Y C\T C Y < 0, ~  
\end{multline}
which can substituted with Eq.~(\ref{eq:sys_Gain_close}) to give:
\begin{align}
    (A_p Y +B_p G Y) + (\bullet)\T + \begin{bmatrix} B_{cl} & Y C\T \end{bmatrix} \begin{bmatrix} R^{-1} & \\ & I \end{bmatrix}\begin{bmatrix} B_{cl}\T \\ CY \end{bmatrix} < 0 , 
\end{align}
and after using the Schur's complement with $R > 0$ and $L = GY$ can be written as:
\begin{align}
    \begin{bmatrix} sym(A_pY+B_p L) & B_{cl} & YC\T \\ B_{cl}\T & -R & 0 \\ CY & 0 & -I \end{bmatrix} < 0,~  Y>0.
\end{align}
\hfill \qedsymbol

\noindent \textit{Proof of Lemma~\ref{Lem3}:} The gain matrix $G$ can be calculated using Eq.~(\ref{eq:IE_gain}) which can be substituted with Eq.~(\ref{eq:sys_Gain_close}) to give: 
\begin{multline}
    \min \epsilon,  \hspace{0.5 pc} \epsilon I - B_{cl}\T P B_{cl} > 0, \\  P(A_p + B_p G) +(A_p + B_p G)\T P+ C\T C < 0,
\end{multline} 
and the last equation on both sides can be multiplied by $P^{-1}>0$ to obtain:
\begin{multline}
    \min \epsilon,  \hspace{0.5 pc} \epsilon I - B_{cl}\T P B_{cl} > 0, \\
    (A_p + B_p G)P^{-1} +P^{-1}(A_p + B_p G)\T + P^{-1}C\T C P^{-1}< 0,
\end{multline} 
which can be written as the following after using the Schur's complement with $Y = P^{-1}$ and $R = G Y$ as:
\begin{align}
    \min \epsilon, \hspace{0.25pc} \begin{bmatrix} \epsilon I & B_{cl}\T \\ B_{cl} & Y \end{bmatrix} > 0, \hspace{0.25pc} \begin{bmatrix} sym(A_p Y + B_p R) & Y C\T \\ CY & -I \end{bmatrix} < 0.
\end{align}
\hfill \qedsymbol

\noindent \textit{Proof of Lemma~\ref{Lem4}:} The covariance matrix $X$ for the linear system for a given system can be written as:
\begin{align}
    A_{cl}X+XA_{cl}\T + B_{cl} W B_{cl}\T < 0,
\end{align}
which can be used to bound the output covariance and after substitution from Eq.~(\ref{eq:sys_Gain_close}) can be written as [\cite{Skelton_LMI_1998}]:
\begin{align}
    CXC\T < \bar{Y}, \hspace{0.25 pc} (A_p + B_p G) X+X(A_p + B_p G)\T + B_{cl} W B_{cl}\T < 0,
\end{align}
which again can be written with $R = G Y$ as:
\begin{align}
    &\begin{bmatrix} \bar{Y} & CX \\ XC\T & X \end{bmatrix} > 0, \hspace{0.5 pc}
    \begin{bmatrix} sym(A_p X + B_p R) & B_{cl} \\ B_{cl}\T & -W^{-1} \end{bmatrix} < 0.
\end{align}
\hfill \qedsymbol

\end{document}